\documentclass{article}
\usepackage{array}
\usepackage{iclr2027_conference,times}

\usepackage{amsmath, amssymb, amsthm}
\usepackage{xcolor}
\usepackage{graphicx}
\usepackage{caption}
\usepackage{subcaption}
\usepackage{wrapfig}
\usepackage{float}
\usepackage{longtable}
\usepackage{enumitem}
\usepackage{hyperref}
\usepackage{url}
\usepackage{comment}
\usepackage[capitalise,noabbrev]{cleveref}   %
\usepackage{placeins}
\usepackage{booktabs}

\newtheorem{theorem}{Theorem}
\newtheorem{lemma}[theorem]{Lemma}
\newtheorem{proposition}[theorem]{Proposition}
\newtheorem{corollary}[theorem]{Corollary}
\theoremstyle{definition}
\newtheorem{definition}[theorem]{Definition}
\newtheorem{remark}[theorem]{Remark}
\newtheorem{assumption}[theorem]{Assumption}
\numberwithin{theorem}{section}

\usepackage{etoolbox}

\AtBeginEnvironment{lemma}{\crefalias{theorem}{lemma}}
\AtBeginEnvironment{proposition}{\crefalias{theorem}{proposition}}
\AtBeginEnvironment{corollary}{\crefalias{theorem}{corollary}}
\AtBeginEnvironment{definition}{\crefalias{theorem}{definition}}
\AtBeginEnvironment{remark}{\crefalias{theorem}{remark}}
\AtBeginEnvironment{assumption}{\crefalias{theorem}{assumption}}

\crefname{lemma}{Lemma}{Lemmas}
\crefname{proposition}{Proposition}{Propositions}
\crefname{corollary}{Corollary}{Corollaries}
\crefname{definition}{Definition}{Definitions}
\crefname{remark}{Remark}{Remarks}
\crefname{assumption}{Assumption}{Assumptions}

\crefname{figure}{fig.}{figs.}
\Crefname{figure}{Fig.}{Figs.}
\crefname{table}{tab.}{tabs.}
\Crefname{table}{Tab.}{Tabs.}
\crefname{section}{sec.}{secs.}
\Crefname{section}{Sec.}{Secs.}
\crefname{equation}{eq.}{eqs.}
\Crefname{equation}{Eq.}{Eqs.}
\crefname{theorem}{thm.}{thms.}
\Crefname{theorem}{Thm.}{Thms.}
\crefname{remark}{rem.}{rems.}
\Crefname{remark}{Rem.}{Rems.}
\crefname{definition}{dfn.}{dfns.}
\Crefname{definition}{Dfn.}{Dfns.}
\crefname{algorithm}{alg.}{algs.}
\Crefname{algorithm}{Alg.}{Algs.}
\crefname{proposition}{prop.}{props.}
\Crefname{proposition}{Prop.}{Props.}
\crefname{appendix}{App.}{Apps.}
\Crefname{appendix}{App.}{Apps.}
\crefname{subappendix}{App.}{Apps.}
\Crefname{subappendix}{App.}{Apps.}
\apptocmd{\appendix}{\crefalias{section}{appendix}\crefalias{subsection}{subappendix}}{}{}
\crefname{assumption}{asmp.}{asmps.}
\Crefname{assumption}{Asmp.}{Asmps.}
\crefname{lemma}{lem.}{lems.}
\Crefname{lemma}{Lem.}{Lems.}

\newcommand{\nud}[1]{\nu^{\mathrm{down}}_{#1}}
\newcommand{\nuu}[1]{\nu^{\mathrm{up}}_{#1}}
\newcommand{\gen}{\boldsymbol{\mathcal{M}}}

\newcommand{\mA}{\mathbf{A}}
\newcommand{\mB}{\mathbf{B}}
\newcommand{\mD}{\mathbf{D}}
\newcommand{\mF}{\mathbf{F}}
\newcommand{\mG}{\mathbf{G}}

\newcommand{\mI}{\mathbf{I}}
\newcommand{\mL}{\mathbf{L}}
\newcommand{\mM}{\mathbf{M}}
\newcommand{\mP}{\mathbf{P}}
\newcommand{\mS}{\mathbf{S}}
\newcommand{\mT}{\mathbf{T}}

\newcommand{\mW}{\mathbf{W}}
\newcommand{\mX}{\mathbf{X}}
\newcommand{\mDelta}{\boldsymbol{\Delta}}
\newcommand{\mOmega}{\boldsymbol{\Omega}}
\newcommand{\mPi}{\boldsymbol{\Pi}}

\newcommand{\mTheta}{\boldsymbol{\Theta}}
\newcommand{\mcA}{\boldsymbol{\mathcal{A}}}
\newcommand{\mcL}{\boldsymbol{\mathcal{L}}}

\newcommand{\vc}{\mathbf{c}}
\newcommand{\vf}{\mathbf{f}}
\newcommand{\vg}{\mathbf{g}}
\newcommand{\vh}{\mathbf{h}}
\newcommand{\vr}{\mathbf{r}}
\newcommand{\vu}{\mathbf{u}}
\newcommand{\vv}{\mathbf{v}}

\newcommand{\vx}{\mathbf{x}}
\newcommand{\vz}{\mathbf{z}}
\newcommand{\vone}{\mathbf{1}}

\newcommand{\mv}{\mathbf{v}}
\newcommand{\vphi}{\boldsymbol{\phi}}
\newcommand{\vpsi}{\boldsymbol{\psi}}

\renewcommand{\paragraph}[1]{{\vspace{0.3mm}\noindent \bf #1}.}
\newcommand{\paragraphnoper}[1]{{\vspace{0.3mm}\noindent \bf #1}}

\apptocmd{\appendix}{\crefalias{subsubsection}{subappendix}}{}{}

\title{Hodge Spectral Concentration \\in Simplicial Complex Neural Networks}

\author{Iris Liao, Maximilian Krahn, Emre Ulusoy, Lennart Bastian\textsuperscript{\textdagger} \& Tolga Birdal\textsuperscript{\textdagger} \\
Imperial College London           
{\normalfont\small\textsuperscript{\textdagger}Equal senior authorship.}
}

\iclrfinalcopy %
\begin{document}

\maketitle
\lhead{Preprint. Under review.}

\begin{abstract}
Simplicial neural networks extend learning on graphs to higher-order topological structures, but the dynamics of feature propagation in these networks remain understudied. We develop a framework that characterizes these dynamics through the Hodge decomposition: rather than a single energy, as used for over-smoothing and -sharpening in graph neural networks, we track the feature-normalized lower and upper Hodge Rayleigh quotients separately. This decouples spectral concentration from changes in feature magnitude and retains information about the gradient and curl subspaces, allowing us to distinguish three forms of \emph{spectral concentration}: harmonic-dominant, gradient-dominant, and curl-dominant dynamics. For linear simplicial convolutions with independent, symmetric weights, we first show that the dynamics are a gradient flow of a learnable energy and derive three competing growth rates associated with the harmonic and highest-frequency gradient and curl eigenspaces, then show that these rates determine the asymptotic regime for almost every input. We further bound how far each Rayleigh quotient is from its predicted limit at any finite depth, extend this bound to elementwise nonlinearities such as $\tanh$ under certain conditions, and show that the results hold at every simplicial rank. Computed from the learned weights and Hodge frequencies, these rates provide a post-training diagnostic of which eigenspace the normalized features concentrate in. We instantiate our framework with normalized propagation choices, including an adjacency-like form that recovers graph convolution at rank zero. Finally, experiments on synthetic edge classification, ocean-drifter trajectory prediction, and additional real-world datasets with nonlinear flows and other simplicial architectures validate the predictions and illustrate how supervision and the choice of propagation operator select the regime.
Our project page is available 
\href{https://circle-group.github.io/research/HodgeSpectralConcentration}{\small\texttt{here}}.
\end{abstract}

\vspace{-2mm}
\section{Introduction}
\vspace{-2mm}
\noindent Modeling structured data in domains such as molecular property prediction, social networks, and knowledge bases benefits from inductive biases that inherently preserve relationships~\citep{gilmer2017neural,ying2018graph}.
While mainstay methods for relational learning represent interactions using pairwise graphs (e.g., GNNs), scholars have recently proposed numerous architectures that faithfully represent group-wise interactions through higher-order topological structures such as simplicial and cell complexes~\citep{bunch2020simplicial,pmlr-v139-bodnar21a,papamarkou2024position}. 
A growing family of topological neural networks learns from signals supported on nodes, edges, faces, and higher-dimensional cells~\citep{hajij2022topological,yang2022scnn,yang2023sccnn}.

\noindent The theoretical understanding of these higher-order architectures, however, remains considerably less mature than that of their graph counterparts. 
Simplicial neural networks (SNNs) propagate information using operators derived from the Hodge Laplacian, which captures interactions both below and above a given simplicial rank. 
For edge signals, the lowest rank at which both types of interactions occur, Hodge theory provides an orthogonal decomposition into gradient, harmonic, and curl components, representing qualitatively different forms of information flow~\citep{lim2020hodge}. 
Recent work has begun to use this structure to analyze SNN expressivity and over-smoothing~\citep{yang2025hodgeaware,einizade2025continuous}, but there is not yet a comparably established set of definitions and analytical tools for understanding deep feature propagation in SNNs as for GNNs. 
This limits both our understanding of existing architectures and the principled design of higher-order networks.

\noindent \textbf{Contributions.} In this work, we develop a framework for understanding these dynamics. In particular, we characterize deep feature propagation in SNNs through a pair of feature-normalized lower and upper Hodge-theoretic Rayleigh quotients, whose joint behavior distinguishes harmonic-dominant (HD), gradient-dominant (GD), and curl-dominant (CD) regimes (\cref{def:rqdefs,prop:regimeprops}). 
For linear edge-feature flows, we establish a gradient-flow formulation (\cref{thm:gradflow}) and show that the depth-asymptotic analysis reduces to three competing spectral growth rates associated with the harmonic, gradient, and curl components. 
When the harmonic space is nontrivial and one rate is strictly largest, their comparison determines the asymptotic regime for almost every input (\cref{thm:trichotomy}). 
These rates can be computed from the trained channel weights and eigenvalues of the lower and upper Hodge Laplacians, yielding a post-training diagnostic of the Hodge structure amplified by the network. Beyond the limit, we bound the distance of each Rayleigh quotient from its predicted endpoint at any finite depth, and show that this bound extends to elementwise nonlinearities such as $\tanh$ when nonlinear leakage between Hodge subspaces is small (\cref{prop:nonlinear}). All results hold at arbitrary simplicial rank (\cref{rem:rank-main}). 
Experiments on synthetic edge classification and ocean-drifter prediction support these predictions and show how supervision and the propagation operator select among the different regimes, as illustrated by the phase portrait in~\cref{fig:phase-portrait}. Additional experiments with $\tanh$ flows, published SNN architectures (SCoNe, SCNN, SCCNN) and two further real-world datasets show that the paired Rayleigh quotients remain informative beyond the linear setting (\cref{app:additional-experiments}).
\vspace{-2mm}
\section{Preliminaries}
\label{sec:preliminaries}
\vspace{-2mm}
    We review the necessary background on simplicial complexes and Hodge theory, referring to \citet{hatcher2002algebraic} for extended background in algebraic topology and \citet{eckmann1944harmonische} for the classical treatment of combinatorial Hodge theory.
\vspace{-2mm}
\begin{wrapfigure}[24]{r}{0.36\textwidth}
    \centering
    \includegraphics[width=\linewidth]{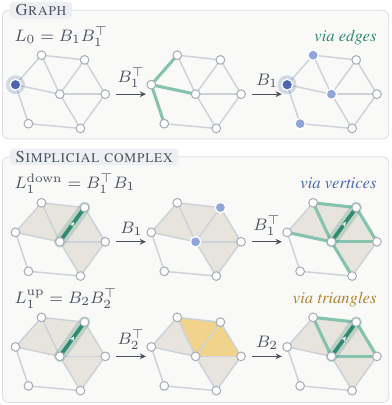}
    \caption{\textbf{Feature propagation in graphs and simplicial complexes (SCs).} On a graph, $\mL_0=\mB_1\mB_1^\top$ sends a node signal to its incident edges and back to the neighboring nodes. On a simplicial complex, an edge signal has two channels: $\mL_1^{\mathrm{down}}=\mB_1^\top \mB_1$ couples edges that share a vertex (gradient), while $\mL_1^{\mathrm{up}}=\mB_2\mB_2^\top$ couples edges that share a filled triangle (curl). The unfilled polygon is a hole ($\beta_1=1$), which supports a harmonic signal.}
    \label{fig:laplacian-ops}
    \vspace{-4mm}
\end{wrapfigure}
\vspace{-4mm}
\subsection{Simplicial Complexes and Hodge Theory}
\label{sec:frequencies}
\vspace{-1mm}

\begin{definition}[Simplicial complex {\citep{eckmann1944harmonische}}]
\label{def:simplicial-complex}
A finite abstract simplicial complex $X$ on a finite vertex set $V$ is a collection of non-empty subsets of $V$ such that $\{v\}\in X$ for every $v\in V$ and $\varnothing\neq\tau\subseteq\sigma\in X$ implies $\tau\in X$. We omit the empty simplex. A simplex with $k+1$ vertices is called a $k$-simplex, or $k$-face. Denote the set of $k$-faces by $X_k$ and let $N_k=|X_k|$. If $k=1$ we can interpret the simplex as a graph.
\end{definition}
Boundary matrices encode how simplices of adjacent ranks are incident (see \cref{fig:laplacian-ops}). 
These give us the Hodge Laplacian:

\begin{definition}[Hodge Laplacian {\citep{hatcher2002algebraic}}]
\label{def:hodgeLaplacian}
\noindent %
For $k\geq1$, let $\mB_k\in\mathbb{R}^{N_{k-1}\times N_k}$ denote the boundary matrix, mapping an oriented $k$-simplex to its $(k-1)$-faces, with $\mB_0=0$. Since $\mB_k\mB_{k+1}=0$, the $k$-th combinatorial Hodge Laplacian is
\[
\mL_k
=
\underbrace{\mB_k^\top\mB_k}_{\mL_k^{\mathrm{down}}}
+
\underbrace{\mB_{k+1}\mB_{k+1}^\top}_{\mL_k^{\mathrm{up}}}.
\]
The lower term couples $k$-simplices sharing a $(k-1)$-face, while the upper term couples those contained in a common $(k+1)$-simplex (\cref{fig:laplacian-ops}). Both are symmetric positive semidefinite and satisfy
$\mL_k^{\mathrm{down}}\mL_k^{\mathrm{up}}
=
\mL_k^{\mathrm{up}}\mL_k^{\mathrm{down}}=0$.
\end{definition}

Based on the Hodge Laplacian, we can decompose a signal space $\mathbb{R}^{N_k}$ on $X$ into distinct subspaces.

\begin{definition}[Hodge decomposition]
\label{def:hodge-decomposition}
The Hodge decomposition of the space of real signals on $k$-faces is the orthogonal direct sum
\[
\mathbb{R}^{N_k}
=
\operatorname{im}(\mB_k^\top)
\oplus
\ker(\mL_k)
\oplus
\operatorname{im}(\mB_{k+1}).
\]
Consequently, every \(\mathbf s^k\in\mathbb R^{N_k}\) decomposes
uniquely as
$\mathbf s^k=\mathbf s_G^k+\mathbf s_C^k+\mathbf s_H^k$
where \(\mathbf s_G^k\), \(\mathbf s_C^k\), and \(\mathbf s_H^k\)
are its gradient, curl, and harmonic components, respectively.
\end{definition}

The kernel of $\mL_k$ captures the \textit{homology} over $\mathcal{X}$. 
$\beta_k=\dim\ker(\mL_k)$ is the $k$-th Betti number, which counts independent topological features: $\beta_0$ is the number of connected components, $\beta_1$ the number of loops that do not bound filled triangles, and $\beta_2$ the number of enclosed voids.

\begin{remark}[Simplicial frequencies]
\label{rem:simpfrec}
Since \(\mL_k^{\mathrm{down}}\mL_k^{\mathrm{up}}=\mathbf{0}\), the lower and upper
Hodge Laplacians admit a common orthonormal eigenbasis. Each
joint eigenvector \(\mv\), satisfying $\mL_k^{\mathrm{down}}\mv=\nu_g\mv$ and $\mL_k^{\mathrm{up}}\mv=\nu_c\mv$,
has an eigenvalue pair of one of the three forms
\[
\underbrace{(\nu_g,0),\ \nu_g>0}_{\text{gradient}}
\qquad
\underbrace{(0,\nu_c),\ \nu_c>0}_{\text{curl}}
\qquad
\underbrace{(0,0)}_{\text{harmonic}}.
\]
Thus, the nonzero eigenvalues of \(\mL_k^{\mathrm{down}}\) and
\(\mL_k^{\mathrm{up}}\) define the gradient and curl frequency families,
respectively~\citep{yang2022simplicialfilters}.
\end{remark}
\begin{wrapfigure}[23]{r}{0.45\textwidth}
    \vspace{-9mm}
    \centering
    \includegraphics[width=\linewidth]{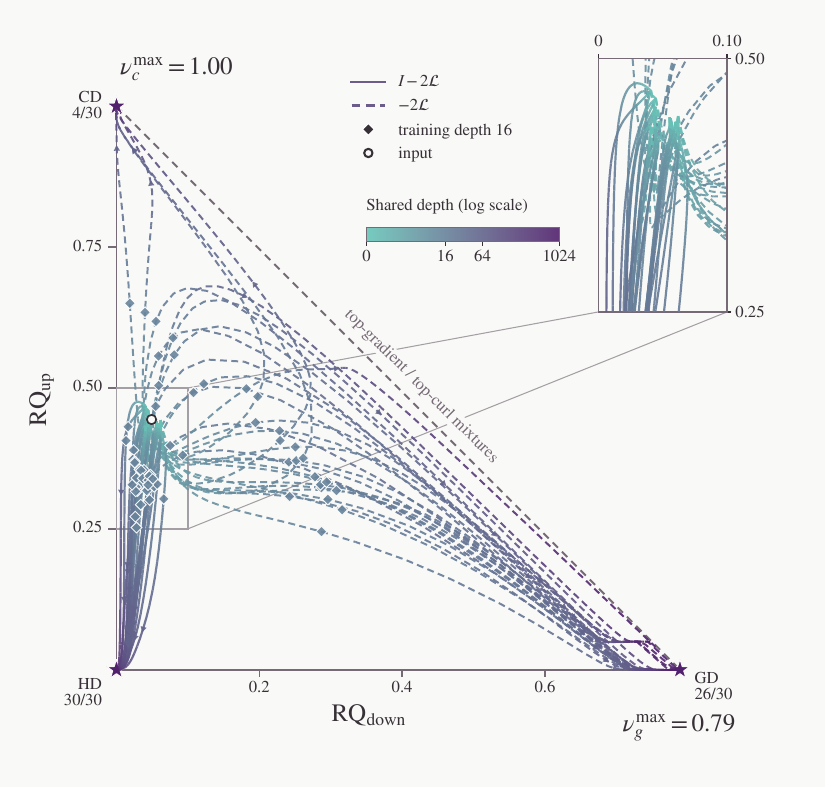}
    \caption{\textbf{Phase diagram depicting transition states during gradient flow of an SNN.} We start from an initial input and evolve over time. We plot the relative spectral concentration measured via up and down Hodge Rayleigh quotients (\cref{def:rqdefs}) for inference runs on an ocean drifter \citep{roddenberry2021scone} dataset showing eventual convergence to one of three regimes (see \cref{prop:regimeprops}).}
    \label{fig:phase-portrait}
    \vspace{-4mm}
\end{wrapfigure}

\noindent For edge signals ($k=1$), the decomposition admits a physical interpretation: the gradient component is curl-free, the curl component is divergence-free, and the harmonic component is both divergence- and curl-free~\citep{yang2022simplicialfilters,lim2020hodge}. 
On a connected graph, harmonic node signals are constant. Harmonic edge signals, however, can describe circulation around holes without being constant.
Whenever the corresponding Hodge subspace is nontrivial, we write
$\nu_g^{\min}\leq\nu_g^{\max}$ for the smallest and largest gradient
frequencies, i.e.\ the eigenvalues of $\mL_k^{\mathrm{down}}$
restricted to $\operatorname{im}(\mB_k^\top)$, and
$\nu_c^{\min}\leq\nu_c^{\max}$ analogously for the curl frequencies of
$\mL_k^{\mathrm{up}}$ restricted to $\operatorname{im}(\mB_{k+1})$.
These frequencies are strictly positive. If $\beta_k>0$, then $\lambda_{\min}(\mL_k)=0$ and the corresponding eigenspace is harmonic. 
If $\beta_k=0$, then all eigenvalues of $\mL_k$ are positive and $\lambda_{\min}(\mL_k)
=
\min\{\nu_g^{\min},\nu_c^{\min}\}$, with the minimum taken over the Hodge families that are present.

\vspace{-1mm}

\subsection{Learning on Simplicial Complexes}
\label{sec:model}
\vspace{-1mm}

We study the gradient flow of a simple SNN.
\begin{proposition}[Rankwise Simplicial Convolutional Neural Network Layer]
\label{def:simplicial-convolutional-neural network}
For a fixed rank $k$ and feature dimension $d$ we consider a first-order layer that updates features $\mF_m\in\mathbb R^{N_k\times d}$ by
\begin{equation}
\label{eq:simplicial-layer}
\mF_{m+1}=\sigma\!\left(\mF_m+\tau\bigl[\mF_m\mW_0^{(m)}
+\mL_k^{\mathrm{down}}\mF_m\mW_1^{(m)}
+\mL_k^{\mathrm{up}}\mF_m\mW_2^{(m)}\bigr]\right),
\end{equation}
where $m$ indexes layers, $\tau>0$ is a step size, $\sigma$ acts elementwise,
and the learnable matrices $\mW_i^{(m)}\in\mathbb R^{d\times d}$ for $i \in \{0,1,2\}$ mix channels.
The three learned terms act on each $k$-simplex itself, on other $k$-simplices
sharing a $(k-1)$-face, and on those sharing a $(k+1)$-coface, respectively.
\end{proposition}

This construction connects closely to widely adopted architectures in the learning literature.
For instance, with combinatorial Hodge operators, matching activations and
constant width,
\cref{eq:simplicial-layer} has the hidden-layer form of SCoNe at rank
$1$~\citep{roddenberry2021scone}, degree-one SCNN~\citep{yang2022scnn},
and degree-one SCCNN without inter-rank terms ~\citep{yang2023sccnn}.
Our theory studies the linear, symmetric, depth-shared restriction defined below.
Linearity preserves every Hodge subspace, symmetric weights yield a gradient flow with real growth rates, and degree-one filters reduce the asymptotic regime to three endpoint rates (\cref{sec:trichotomy}).
\Cref{eq:simplicial-layer} is the single-rank, degree-one instance of a common layer formulation that contains all three architectures (\cref{def:snn}).
We detail the exact correspondence in \cref{app:instantiations}.

\begin{proposition}[Linear simplicial gradient flow]
\label{def:linear-simplicial-flow}
In \eqref{eq:simplicial-layer}, set $\sigma=\mathrm{id}$ and share symmetric weights across depth: $\mW_i^{(m)}=\mW_i=\mW_i^\top$ for $i=0,1,2$.
The resulting network is exactly the forward Euler discretization, with step size $\tau$, of
\begin{equation}
\label{eq:flow}
\dot{\mF}=\mF\mW_0
+\mL_k^{\mathrm{down}}\mF\mW_1
+\mL_k^{\mathrm{up}}\mF\mW_2,
\end{equation}
where $\mF\colon t\mapsto\mF(t)\in\mathbb R^{N_k\times d}$ gives the features at continuous depth $t\ge0$ (argument omitted when clear), $\dot{\mF}=\mathrm d\mF/\mathrm dt$, and $\mF(0)=\mF_0$. Layer $m$ approximates the flow at time $t_m=m\tau$.
\end{proposition}
We call \cref{eq:flow} our \emph{general affine model}: its propagation operator is affine in the lower and upper Laplacians, and it only depends linearly on features.
Because the channel weights are symmetric, this yields a negative gradient flow (\cref{thm:gradflow}).
In this manner, we analyze a moderately restricted case of the network in \cref{eq:simplicial-layer}, with each Hodge subspace invariant. 
The continuous and discrete models have the same asymptotic regime under the step-size condition in \cref{prop:flip}.

\vspace{-2mm}
\section{A Hodge Spectral Concentration Framework}
\label{sec:theory}
\vspace{-1.5mm}
Before analyzing the flow in \cref{eq:flow}, we first develop a general framework for spectral concentration of edge-feature dynamics that applies to any propagation on edges, including the nonlinear layer \cref{eq:simplicial-layer}. 
Throughout, we specialize to rank $k=1$ (features on edges), the lowest rank at which both lower and upper interactions occur, and write $\mF(t)\in\mathbb R^{N_1\times d}$ for the edge features.

We also establish a notational convention that carries over to the instantiation in \cref{sec:instantiation} and the experiments.
Our analysis leverages root-to-leaf normalized Laplacians $\mcL_k^{\mathrm{down}},\mcL_k^{\mathrm{up}}$ of \citet{vigano2026roottoleafpathrandomwalks} (written $\mcL$ to distinguish them from the unnormalized $\mL$ of \cref{sec:preliminaries}), which uniformly normalize the spectra at every rank of a simplicial complex to $[0,1]$.
This provides a common spectral scale across all ranks and different complexes, for instance, across datasets. 
We provide the construction of this normalized Laplacian in \cref{app:r2l} and discuss the importance of normalization, as well as alternative choices in \cref{app:normalization-alternatives}.
$\hat \mB_1,\hat \mB_2$ denote the boundary maps and may refer to combinatorial or normalized versions, subject only to $\hat \mB_1\hat \mB_2=0$, and
\[
\mcL_1^{\mathrm{down}}:=\hat \mB_1^\top\hat \mB_1,
\qquad
\mcL_1^{\mathrm{up}}:=\hat \mB_2\hat \mB_2^\top,
\qquad
\mcL_1:=\mcL_1^{\mathrm{down}}+\mcL_1^{\mathrm{up}}.
\]
Note that the analysis only depends on symmetry, positive semi-definiteness, and $\mcL_1^{\mathrm{down}}\mcL_1^{\mathrm{up}}=0$.

\vspace{-1mm}
\paragraph{Normalized Dirichlet energy definitions of spectral concentration}
We first introduce a \emph{pair} of Rayleigh quotients, one for each of the up and down Hodge Laplacians, with the aim of respecting the natural structure that arises from the Hodge decomposition. 
The pair, as opposed to a single summed quantity, retains information about the gradient and curl subspaces separately and allows us to define three distinct limiting spectral concentration regimes.

\begin{proposition}[Hodge-theoretic Rayleigh quotients]
\label{def:rqdefs}
For $\mF \in \mathbb{R}^{N_1\times d}$, $\mF \neq 0$, define
\begin{equation}
\mathrm{RQ}_{\mathrm{down}}(\mF) := \frac{\mathrm{Tr}\!\big(\mF^\top \mcL_1^{\mathrm{down}}\, \mF\big)}{\|\mF\|^2},
\qquad
\mathrm{RQ}_{\mathrm{up}}(\mF) := \frac{\mathrm{Tr}\!\big(\mF^\top \mcL_1^{\mathrm{up}}\, \mF\big)}{\|\mF\|^2},
\end{equation}
with
\begin{equation}
\mathrm{RQ}(\mF) := \mathrm{RQ}_{\mathrm{down}}(\mF) + \mathrm{RQ}_{\mathrm{up}}(\mF)
= \frac{\mathrm{Tr}\!\big(\mF^\top \mcL_1 \mF\big)}{\|\mF\|^2},
\end{equation}
with $\|\cdot\|$ the Frobenius norm. $\mathrm{RQ}(\mF)$ is the Dirichlet energy normalized by the feature norm, i.e.\ the Rayleigh quotient of $\mcL_1$.
\end{proposition}

\noindent These are channel-summed Rayleigh quotients of symmetric positive
semi-definite operators. Whenever the corresponding Hodge family is nontrivial, $0\leq\mathrm{RQ}_{\mathrm{down}}(\mF)\leq\nu_g^{\max}$, $0\leq\mathrm{RQ}_{\mathrm{up}}(\mF)\leq\nu_c^{\max}$, while in all cases $\lambda_{\min}(\mcL_1)
\leq \mathrm{RQ}(\mF) \leq \lambda_{\max}(\mcL_1)$ (see \cref{eq:rq-masses}).

\begin{assumption}[Nondegeneracy]
\label{ass:nondeg}
Both the gradient and the curl frequency families are nontrivial, i.e.\ the complex has at least one edge and one triangle, and $\nu_g^{\max}>\lambda_{\min}(\mcL_1)$ and $\nu_c^{\max}>\lambda_{\min}(\mcL_1)$.
\end{assumption}
\begin{proposition}[Harmonic-, gradient-, and curl-dominant dynamics]
\label{prop:regimeprops}
Let $t \mapsto \mF(t) \in \mathbb{R}^{N_1\times d}\setminus\{0\}$ be any edge-feature dynamics, in continuous time or in discrete depth (layers). Under \cref{ass:nondeg}, the dynamics is
\begin{itemize}[noitemsep,topsep=0em]
    \item[(i)] \emph{harmonic-dominant} (HD) if  $\mathrm{RQ}(\mF(t)) \to 0$ as $t \to \infty$;
    \item[(ii)] \emph{gradient-dominant} (GD) if  $\mathrm{RQ}_{\mathrm{down}}(\mF(t))\to\nu_g^{\max}$;
    \item[(iii)] \emph{curl-dominant} (CD) if $\mathrm{RQ}_{\mathrm{up}}(\mF(t))\to\nu_c^{\max}$.
\end{itemize}
\end{proposition}

\noindent These limits describe \emph{spectral concentration}: the fraction of squared feature mass outside the selected eigenspace tends to zero.
HD concentrates the harmonic subspace. GD and CD concentrate in the highest-frequency gradient and curl eigenspaces, respectively. 
This gives a direct connection to graph over-smoothing / sharpening, where low- and high-frequency dominance (LFD/HFD) describe concentration in the lowest and highest graph Laplacian eigenspaces~\citep{digiovanni2023understandingconvolutiongraphsenergies} which we describe in \cref{app:graphbg}. 
At rank $0$, the harmonic criterion recovers graph LFD, where features converge to a degree-scaled constant eigenvector. 
For edge features, however, harmonic dominance retains information about global topological holes, which may be helpful or harmful depending on the learning task. Gradient and curl dominance are also distinct behaviors, which do not have natural analogs in the graph case. 
Because each asymptotic regime corresponds to convergence to a Hodge subspace with distinct characteristics, the over-smoothing / sharpening dichotomy no longer applies.
We henceforth use the terms HD, GD, and CD as the three generic forms of spectral concentration.

\noindent These are \emph{model-agnostic} and apply to arbitrary continuous- or discrete-depth dynamics. 
The same construction extends to arbitrary rank, with any absent lower or upper Hodge family omitted at the boundary ranks.
Fundamental properties of the RQ pair, including the feasible region, a bottom-eigenspace characterization of HD, and the mutual exclusion of GD and CD, are in \cref{app:rqlemmas}.

\vspace{-1mm}
\paragraph{Energy and gradient flow}
\noindent We now study the family of linear models with symmetric weights introduced in \cref{sec:model} as the gradient flow of a parametric energy and show that Hodge subspace suppression and amplification can be induced by the positive and negative eigenvalues of the learnable weights. We then derive the exact conditions for each type of behavior.
\noindent We define the parametric multi-particle energy over the edge features as
\begin{equation}
\mathcal{E}(\mF) = -\tfrac12\,\mathrm{Tr}\!\big(\mF^\top \mF\, \mW_0\big) -\tfrac12\,\mathrm{Tr}\!\big(\mF^\top \mcL_1^{\mathrm{down}} \mF\, \mW_1\big) - \tfrac12\,\mathrm{Tr}\!\big(\mF^\top \mcL_1^{\mathrm{up}} \mF\, \mW_2\big).
\end{equation}
\begin{theorem}
\label{thm:gradflow}
If $\mW_0$, $\mW_1$ and $\mW_2$ are symmetric, then the dynamical system in \cref{eq:flow} is the negative gradient flow of $\mathcal{E}(\mF)$, and consequently $\tfrac{d}{dt}\mathcal{E}(\mF(t)) \le 0$.
\end{theorem}

\noindent This gradient-flow property holds for each affine (indeed polynomial) propagation operator built from symmetric Laplacians, whether normalized or not (proof in \cref{app:proofs}).
The last two terms of $\mathcal E$ are quadratic forms in the divergence $\hat \mB_1\mF$ at each node and in the circulation $\hat \mB_2^\top \mF$ around each $2$-simplex. 
At a node, divergence is the signed, normalization-weighted sum of the features on its incident edges. 
A negative eigenvalue of $\mW_1$ (resp.\ $\mW_2$) penalizes divergence (resp.\ circulation) and acts attractively, while a positive one acts repulsively. \Cref{app:energy} develops this interpretation.

\vspace{-1mm}
\subsection{Asymptotic Regimes of the Flow}
\vspace{-1mm}
\label{sec:trichotomy}

\noindent The regimes of \cref{prop:regimeprops} apply to any edge-feature dynamics. 
For the flow in \cref{eq:flow}, we now show that exactly one of HD, GD, or CD occurs for almost every input and present the conditions on the channel weights and Hodge frequencies that determine each regime. 
Throughout this section \cref{ass:nondeg} is in force and we assume that $\beta_1>0$. 
The case $\beta_1=0$ is treated in \cref{app:nobetti}. 
We first decouple the flow into a $d$-dimensional linear system for each joint eigenvector of $\mcL_1^{\mathrm{down}}$ and $\mcL_1^{\mathrm{up}}$, and observe that the normalized features concentrate on the fastest-growing eigenvectors of the flow.

\paragraph{Eigenvectors of the flow}
Since $\mcL_1^{\mathrm{down}}\mcL_1^{\mathrm{up}}=0$ and both operators are symmetric, they share an orthonormal eigenbasis $\{\vu_j\}_{j=1}^{N_1}$, $\mcL_1^{\mathrm{down}}\vu_j=\nud{j}\,\vu_j$ and $\mcL_1^{\mathrm{up}}\vu_j=\nuu{j}\,\vu_j$, 
where $(\nud{j},\nuu{j})$ is $(\nu,0)$, $(0,\nu)$ or $(0,0)$ with $\nu>0$ for a gradient, curl or harmonic eigenvector (\cref{rem:simpfrec}). Since the Laplacians act on the left and the weights on the right, the channel coefficients $\vc_j:=\mF^\top\vu_j\in\mathbb R^d$ of $\mF$ along each $\vu_j$ evolve independently,
\begin{equation}
\label{eq:mode-ode}
\dot\vc_j=\mG_j\,\vc_j,
\qquad
\mG_j:=\mW_0+\nud{j}\mW_1+\nuu{j}\mW_2 ,
\end{equation}
with a \emph{channel generator} $\mG_j$, symmetric because the weights are, that depends on $\vu_j$ only through its frequency pair. Let $\vpsi_{j,r}$, $r=1,\dots,d$, be orthonormal eigenvectors of $\mG_j$ with eigenvalues $\mu_{j,r}$. The $N_1d$ rank-one matrices $\vu_j\vpsi_{j,r}^\top$, each pairing one spatial eigenvector with one channel eigenvector, form an orthonormal eigenbasis of the linear map $\mF\mapsto\mF\mW_0+\mcL_1^{\mathrm{down}}\mF\mW_1+\mcL_1^{\mathrm{up}}\mF\mW_2$ that generates \cref{eq:flow}, with eigenvalues $\mu_{j,r}$. Expanding the input in this basis, $\mF(0)=\sum_{j,r}a_{j,r}\,\vu_j\vpsi_{j,r}^\top$ with $a_{j,r}:=\vu_j^\top\mF(0)\vpsi_{j,r}$, the solution of \cref{eq:flow} is
\begin{equation}
\label{eq:closedform}
\mF(t)=\sum_{j,r}e^{\mu_{j,r}t}\,a_{j,r}\,\vu_j\vpsi_{j,r}^\top .
\end{equation}
The flow rescales each coefficient of the input by $e^{\mu_{j,r}t}$ . Let $\mu_*:=\max_{j,r}\mu_{j,r}$ be the top rate, $T_*:=\{(j,r):\mu_{j,r}=\mu_*\}$, and $E_*:=\operatorname{span}\{\vu_j\vpsi_{j,r}^\top:(j,r)\in T_*\}$ the top eigenspace.

\begin{proposition}[Normalized limit]
\label{prop:closedform}
By orthonormality, $\|\mF(t)\|^2=\sum_{j,r}e^{2\mu_{j,r}t}a_{j,r}^2$, and for every $\mF(0)$ not orthogonal to $E_*$, i.e.\ with $\sum_{(j,r)\in T_*}a_{j,r}^2>0$, the normalized features converge to the normalized orthogonal projection of $\mF(0)$ onto $E_*$:
\begin{equation}
\label{eq:normlimit}
\begin{aligned}
\frac{\mF(t)}{\|\mF(t)\|}
&\longrightarrow
\frac{\sum_{(j,r)\in T_*}a_{j,r}\,\vu_j\vpsi_{j,r}^\top}
     {\big(\sum_{(j,r)\in T_*}a_{j,r}^2\big)^{1/2}},
\\
\big(\mathrm{RQ}_{\mathrm{down}},\mathrm{RQ}_{\mathrm{up}}\big)(\mF(t))
&\longrightarrow
\frac{\sum_{(j,r)\in T_*}a_{j,r}^2\,\big(\nud{j},\nuu{j}\big)}
     {\sum_{(j,r)\in T_*}a_{j,r}^2}.
\end{aligned}
\end{equation}
\end{proposition}
\noindent Every eigenvector with rate below $\mu_*$ is suppressed relative to $\|\mF(t)\|$ by a factor $e^{(\mu_{j,r}-\mu_*)t}$, so the normalized features concentrate on $E_*$, and the Rayleigh-quotient pair converges to the mass-weighted average of the frequency pairs in $T_*$ (proof in \cref{app:normlimit}). The regime therefore corresponds to which eigenvectors $\vu_j$ appear in $E_*$.

\paragraph{Endpoint rates}
Define the three \emph{endpoint rates}
\begin{equation}
\label{eq:rates}
\rho_H:=\lambda_{\max}(\mW_0),
\qquad
\rho_G:=\lambda_{\max}\big(\mW_0+\nu_g^{\max}\mW_1\big),
\qquad
\rho_C:=\lambda_{\max}\big(\mW_0+\nu_c^{\max}\mW_2\big),
\end{equation}
the largest eigenvalues of the channel generators of a harmonic mode and of the top-frequency gradient and curl modes. Because $\lambda_{\max}(\mG_j)$ is a convex function of the frequency of $\vu_j$ within each Hodge family, and $\beta_1>0$ ensures all three endpoint frequencies occur, the top rate is $\mu_*=\max\{\rho_H,\rho_G,\rho_C\}$ (\cref{lem:endpoint}, \cref{app:proofs}).

\begin{theorem}[Spectral concentration trichotomy]
\label{thm:trichotomy}
Suppose the maximum of $\{\rho_H,\rho_G,\rho_C\}$ is attained by exactly one of them. Then, for almost every initial condition $\mF(0)$,
\begin{itemize}[leftmargin=7.5mm,itemsep=0.01em,topsep=0em]
    \item[(i)] if $\rho_H>\max\{\rho_G,\rho_C\}$, the flow is HD:
    $\big(\mathrm{RQ}_{\mathrm{down}},\mathrm{RQ}_{\mathrm{up}}\big)(\mF(t))\longrightarrow(0,0)$;
    \item[(ii)] if $\rho_G>\max\{\rho_H,\rho_C\}$, the flow is GD:
    $\big(\mathrm{RQ}_{\mathrm{down}},\mathrm{RQ}_{\mathrm{up}}\big)(\mF(t))\longrightarrow(\nu_g^{\max},0)$;
    \item[(iii)] if $\rho_C>\max\{\rho_H,\rho_G\}$, the flow is CD:
    $\big(\mathrm{RQ}_{\mathrm{down}},\mathrm{RQ}_{\mathrm{up}}\big)(\mF(t))\longrightarrow(0,\nu_c^{\max})$.
\end{itemize}
Here ``almost every'' excludes the initial conditions orthogonal to the top eigenspace $E_*$. The proof is in \cref{app:proofs}.
\end{theorem}

\paragraph{Consequences}
These results require only the trained weights and the two eigenvalues $\nu_g^{\max},\nu_c^{\max}$, so \cref{thm:trichotomy} can serve as a post-training diagnostic of the Hodge subspace selected by network to amplify. 
If $\mW_1\preceq0$ then $\mW_0+\nu\mW_1\preceq\mW_0$, so $\lambda_{\max}(\mW_0+\nu\mW_1)\le\rho_H$ for every $\nu>0$ and gradient dominance is impossible: GD and CD require positive eigenvalues of $\mW_1$ and $\mW_2$, respectively, that is, a repulsive channel direction in the sense of \cref{app:energy}. 
Finally, by \cref{def:linear-simplicial-flow} the layer update \eqref{eq:simplicial-layer} with $\sigma=\mathrm{id}$ and shared symmetric weights is the forward Euler step of \eqref{eq:flow}, which multiplies the component along $\vu_j\vpsi_{j,r}^\top$ by $1+\tau\mu_{j,r}$ per layer instead of $e^{\mu_{j,r}t}$. Let $\mu_{\min}:=\min_{j,r}\mu_{j,r}$ be the smallest rate (see \cref{lem:endpoint}):
\begin{equation}
\label{eq:mumin}
\mu_{\min}=\min\bigl\{\lambda_{\min}(\mW_0),\ \lambda_{\min}(\mW_0+\nu_g^{\max}\mW_1),\ \lambda_{\min}(\mW_0+\nu_c^{\max}\mW_2)\bigr\}.
\end{equation}
The discrete iteration is dominated by the same top-rate eigenspace, and hence has the same HD, GD or CD regime, whenever $2+\tau(\mu_*+\mu_{\min})>0$; for larger steps the bottom-rate eigenspace dominates instead, with a sign alternating from layer to layer (\cref{prop:flip}, \cref{app:discrete}). 
We verify this step-size condition as part of the spectral diagnostics of the learned flow operators (\cref{app:additional-results}).
\vspace{-1mm}
\subsection{Analysis: Finite Depth and (Non-)Linear Propagation}
\label{sec:finite-depth}
\vspace{-1mm}
We now quantify how close a network of finite depth is to the asymptotic limit for both a linear iteration and an elementwise nonlinearity.
An elementwise nonlinearity in~\cref{eq:simplicial-layer} can couple the Hodge subspaces across layers, but the linear regime prediction remains valid when this leakage between Hodge subspaces is small relative to the linear spectral gap. We state this for activations satisfying
\begin{equation}
\label{eq:activation-bounds}
|\sigma(x)|\le|x|,
\qquad
|\sigma(x)-x|\le M_\sigma|x|^p
\qquad\text{for all }x\in\mathbb R,
\end{equation}
for some $p\ge2$ and $M_\sigma\ge0$. The identity corresponds to $M_\sigma=0$. The odd activation $\sigma=\tanh$ satisfies~\eqref{eq:activation-bounds} with $p=3$ and $M_\sigma=\tfrac13$, as does $\arctan$, and softsign $x/(1+|x|)$ with $p=2$ and $M_\sigma=1$. Write $\mF=\mF^++\mF^\perp$ for the orthogonal decomposition of $\mF$ onto $E_*$ and $E_*^\perp$, and let $\alpha_+:=1+\tau\mu_*>0$ and $\gamma_0:=\|\mF_0^+\|/\|\mF_0\|$. Let $q$ be $\mathrm{RQ}_{\mathrm{down}}$ or
$\mathrm{RQ}_{\mathrm{up}}$, $\nu_{\max}$ be $\nu_g^{\max}$
or $\nu_c^{\max}$, respectively, and $q_\infty$ the corresponding
limit in \cref{thm:trichotomy}. Finally, let
\[
C:=
\frac{\sum_{(j,r)\notin T_*}a_{j,r}^{2}}
     {\sum_{(j,r)\in T_*}a_{j,r}^{2}},
\qquad
\kappa:=
\max_{(j,r)\notin T_*}
\frac{|1+\tau\mu_{j,r}|}{|1+\tau\mu_*|}
<1.
\]
The constant $C$ measures the initial squared mass outside $E_*$
relative to that within it, while $\kappa$ controls the geometric
convergence rate. We now state the result, with its proof in \cref{app:discrete}.
\vspace{-1mm}
\begin{proposition}[Finite-depth spectral concentration]
\label{prop:nonlinear}
Under the hypotheses of \cref{thm:trichotomy} and the step-size condition of \cref{prop:flip}(i), consider~\cref{eq:simplicial-layer} with an activation satisfying~\eqref{eq:activation-bounds}, normalized Laplacians, shared symmetric weights, and $\mF_0\not\perp E_*$. Define
\begin{equation}
\delta:=\frac{M_\sigma\|\mF_0\|^{p-1}}{\gamma_0},
\qquad
\Phi_m:=\sum_{k=0}^{m-1}\alpha_+^{(p-1)(k+1)},
\qquad
\Psi_m:=\sum_{k=0}^{m-1}\kappa^{m-1-k}\alpha_+^{(p-1)(k+1)}.
\end{equation}
\begin{itemize}[leftmargin=*,itemsep=0.01em,topsep=0em]
    \item[(i)] \emph{(Finite-depth bound).} For every depth $m\ge1$ such that $\delta\Phi_m<1$, we have $\mF_m^+\ne0$. Defining $s_m:=\|\mF_m^\perp\|^2/\|\mF_m^+\|^2$, the Rayleigh quotient $q$ satisfies
    \begin{equation}
    \label{eq:nonlinear-finite-depth}
    |q(\mF_m)-q_\infty|\le\nu_{\max}\frac{s_m}{1+s_m},
    \qquad
    s_m\le\left(\frac{\sqrt{C}\,\kappa^m+\delta\Psi_m}{1-\delta\Phi_m}\right)^{\!2}.
    \end{equation}
    \item[(ii)] \emph{(Asymptotic trichotomy for contractive flows).} If $\mu_*<0$ (so that $0<\alpha_+<1$) and $\mF_0$ satisfies $\delta\,\le\, {(1-\alpha_+^{p-1})}/{(2\alpha_+^{p-1})}$,
    then $\delta\Phi_m<\tfrac12$ for all $m\ge1$ and $q(\mF_m)\to q_\infty$ at the geometric rate
    \begin{equation}
    \label{eq:nonlinear-asymptotic}
    |q(\mF_m)-q_\infty|=\mathcal{O}\bigl(m^2\tilde\kappa^{2m}\bigr)\longrightarrow0,
    \qquad
    \tilde\kappa:=\max\{\kappa,\alpha_+^{p-1}\}<1.
    \end{equation}
\end{itemize}
\end{proposition}

\paragraph{Linear case}
For the identity activation, the nonlinear remainder vanishes, so $\delta=0$, part~(i) holds at every depth, and $s_m\le C\kappa^{2m}$. Under the step-size condition above and the hypotheses of \cref{thm:trichotomy}, for $\mF_0\not\perp E_*$ we therefore obtain the finite-depth bound
\begin{equation}
\label{eq:linear-finite-depth}
|q(\mF_m)-q_\infty|
\leq
\nu_{\max}\frac{C\kappa^{2m}}{1+C\kappa^{2m}},
\qquad m\geq 1.
\end{equation}
Since the bound is at most $\nu_{\max}C\kappa^{2m}$, depth $m\ge\log(C\nu_{\max}/\varepsilon)/(2\log(1/\kappa))$ suffices for a tolerance $\varepsilon$, where $C$ reflects how well the input is aligned with $E_*$ and $\kappa$ the gap between the top rate and the rest.

\paragraph{Nonlinear case}
The bound~\eqref{eq:nonlinear-finite-depth} separates the linear prediction from the nonlinear correction: $\sqrt{C}\kappa^m$ is the linear term of~\eqref{eq:linear-finite-depth}, while $\delta$, which grows with the input scale relative to its leading component, measures how much mass the nonlinearity can move between Hodge subspaces. At finite depth, \eqref{eq:nonlinear-finite-depth} quantifies when nonlinear leakage remains small enough for the linear endpoint prediction to be informative. In part~(ii), every input direction with $\mF_0^+\ne0$ preserves the linear HD, GD, or CD endpoint under sufficiently small rescaling of the input. Because $\|\mF_m\|\le\alpha_+^m\|\mF_0\|\to0$ when $\alpha_+<1$, \eqref{eq:nonlinear-asymptotic} characterizes the normalized Hodge composition of a vanishing representation. Consistent with this picture, our $\tanh$ flows largely keep the regime predicted from their weights but can stop short of the linear endpoints (\cref{app:additional-experiments}).

\begin{remark}[Arbitrary ranks]
\label{rem:rank-main}
All results of this section, including the finite-depth bounds, hold for rank-$k$ features $\mF\in\mathbb R^{N_k\times d}$ after replacing $(\mcL_1^{\mathrm{down}},\mcL_1^{\mathrm{up}},\mcL_1,\beta_1)$ by $(\mcL_k^{\mathrm{down}},\mcL_k^{\mathrm{up}},\mcL_k,\beta_k)$ and the gradient and curl subspaces by the rank-$k$ coboundary $\operatorname{im}(\hat\mB_k^\top)$ and boundary $\operatorname{im}(\hat\mB_{k+1})$ subspaces, since the proofs use only the symmetry, positive semidefiniteness and orthogonality of the two operators. At a boundary rank, where one operator vanishes, the trichotomy reduces to a dichotomy between harmonic modes and the remaining highest-frequency family. Indeed, at $k=0$ this recovers the low- and high-frequency dominance of graph networks (\cref{app:graph}). Details are in \cref{app:rank_extension}.
\end{remark}
\vspace{-2mm}
\section{Experiments}
\vspace{-2mm}
We aim to demonstrate whether the spectral conditions predict HD, GD, and CD behavior and whether supervision selects task-relevant regimes.
A more exploratory aim is to identify whether the trained model selects the most helpful limiting behavior for the dataset topology and supervision.  Our two main sets of experiments are: (1) an edge classification task using a synthetic SC, and (2) a trajectory-prediction task on a real-world dataset of ocean drifters. For the main flows of each model (i.e.\ before the readout layer, which differ per task), we use the flow models introduced in \cref{sec:instantiation}.

\vspace{-1mm}
\subsection{Architecture and Propagation Operators}
\vspace{-1mm}
\label{sec:instantiation}
For our experiments, we define two propagation variants of the SNN in \cref{eq:simplicial-layer} using normalized Laplacians and two parameter choices in its linear flow \cref{eq:flow}.
Both variants use two symmetric channel-mixing matrices.

\noindent We consider two parameter choices, each expressed in terms of two symmetric matrices $\mW_{\mathrm{down}},\mW_{\mathrm{up}}\in\mathbb R^{d\times d}$. Our principal instantiation uses $\mW_0=\mW_{\mathrm{down}}+\mW_{\mathrm{up}}$, $\mW_1=-2\mW_{\mathrm{down}}$, $\mW_2=-2\mW_{\mathrm{up}}$, giving the adjacency analogue
\begin{equation}
\dot \mF=\mcA_1^{\mathrm{down}}\mF\mW_{\mathrm{down}}+\mcA_1^{\mathrm{up}}\mF\mW_{\mathrm{up}},
\end{equation}
with $\mcA_1^{\mathrm{down}}:=\mI-2\mcL_1^{\mathrm{down}}$ and $\mcA_1^{\mathrm{up}}:=\mI-2\mcL_1^{\mathrm{up}}$. For comparison, setting $\mW_0=0$, $\mW_1=-2\mW_{\mathrm{down}}$, $\mW_2=-2\mW_{\mathrm{up}}$ yields pure Hodge Laplacian propagation,
\begin{equation}
\dot \mF
=
-2\mcL_1^{\mathrm{down}}\mF\mW_{\mathrm{down}}
-2\mcL_1^{\mathrm{up}}\mF\mW_{\mathrm{up}}.
\end{equation}

The adjacency analogue operator maps the bounded frequency interval of the normalized Laplacian $[0,1]$ onto $[-1,1]$. So the harmonic frequency $0$ has multiplier $1$, while the maximal possible normalized frequency $1$ has multiplier $-1$. With normalization, all complexes are on the same scale, allowing the regime conditions derived above to be compared without additional complex-dependent spectral rescaling.
Under this choice, the feature-space generators of Section~\ref{sec:trichotomy} are $\mG^{\mathrm{harm}}=\mW_{\mathrm{down}}+\mW_{\mathrm{up}}$, $\mG^{\mathrm{grad}}_\nu=(1-2\nu)\mW_{\mathrm{down}}+\mW_{\mathrm{up}}$ and $\mG^{\mathrm{curl}}_\nu=\mW_{\mathrm{down}}+(1-2\nu)\mW_{\mathrm{up}}$, and all of the preceding energy and regime results apply directly after substituting these generators. 
We show recovery of graph-level analogs in \cref{app:graph}.

\paragraph{Implementation}
All models propagate edge features at rank $1$. Our principal model uses the adjacency-like flow above, discretized by forward Euler with step $\tau$ as in \eqref{eq:simplicial-layer}, i.e.\ $\mF_{m+1}=\mF_m+\tau\bigl(\mcA_1^{\mathrm{down}}\mF_m\mW_{\mathrm{down}}+\mcA_1^{\mathrm{up}}\mF_m\mW_{\mathrm{up}}\bigr)$, with pure-Laplacian flow as a comparison. Both experiments use independent symmetric channel matrices shared across
depth. Full training details appear
in \cref{app:models-training}.

\vspace{-1mm}
\subsection{Spectral Edge-Role Classification}
\vspace{-1mm}
\label{sec:fourdisk}
We test whether task-supervision can select the asymptotic regime of a linear flow with symmetric, depth-shared weights. 
On a seeded reconstruction of the four-disk complex of \citet{grande2024hodge} (1,600 vertices, 4,671 edges, 3,071 triangles, $\beta_1=1$), we define four edge-classification tasks with a shared, noisy feature input that mixes low- and high-frequency gradient, curl, and harmonic components. Only the labels varied across the four tasks. 
Following the Hodge spectral clustering of \citet{grande2024hodge}, labels for the first three tasks are $k$-means clusters of the harmonic eigenvector, of the top gradient eigenvectors, and of the top curl eigenvectors, respectively (construction, splits and training details in \cref{app:four-disk-labels}), thereby assigning edge ``roles" according to the Hodge decomposition. The fourth, region task labels each edge by its disk membership. After training we compute the endpoint generators, predict the regime from the largest of $\lambda_{\max}(\mG_S)$, $S\in\{H,G,C\}$, and then track the Rayleigh-quotient pair during inference. 

\begin{figure}[t]
\centering
\includegraphics[width=\linewidth]{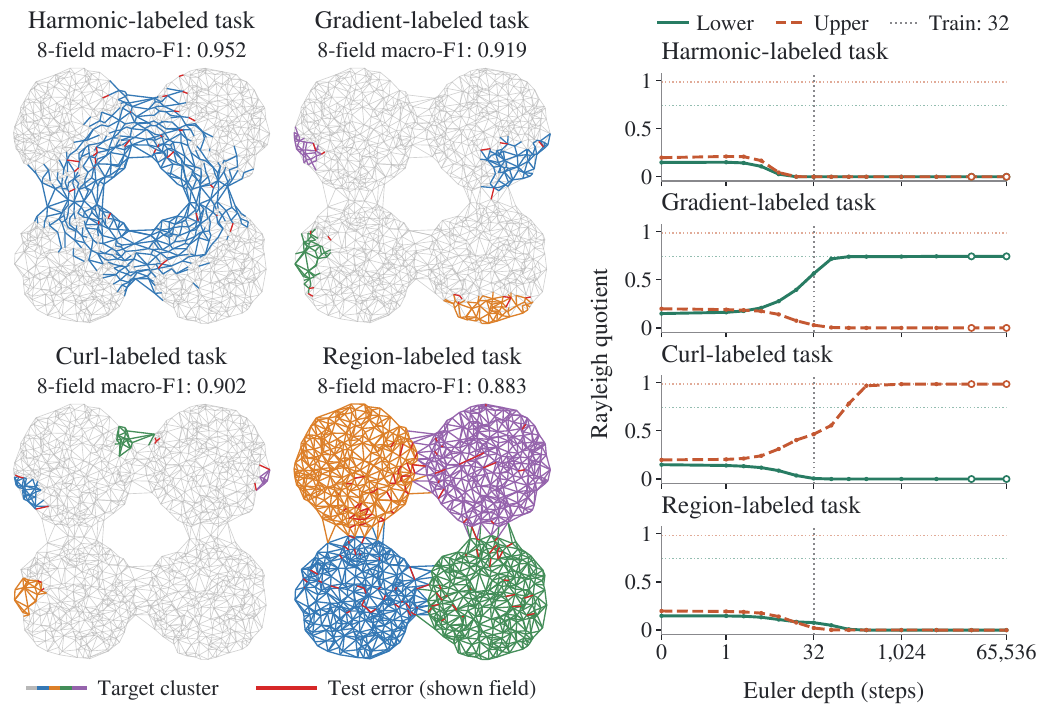}
\vspace{-5.0mm}
\caption{\textbf{(left)} Target labels and prediction error for the four edge-prediction tasks. All tasks share the complex and input features. The first three label edges by harmonic, gradient, and curl roles, the fourth by disk (region). Colors mark classes within each task. Gray is the remaining supervised class, not unlabeled edges. \textbf{(right)} Lower and upper Rayleigh quotients converging to their limits during inference. Horizontal dotted lines mark the maximum Laplacian eigenvalues.\vspace{-4mm}}
\label{fig:fourdisk-labels}
\end{figure}

\paragraph{Results}
The predicted regime matches the task in each of the three Hodge-role tasks (Tab.~\ref{tab:fourdisk}): the harmonic, upper-gradient and upper-curl tasks yield $\eta_H$, $\eta_G$ and $\eta_C$ as the leading rate, respectively, and the Rayleigh-quotient pairs move from the first layers toward the predicted limits $(0,0)$, $(\nu_g^{\max},0)$ and $(0,\nu_c^{\max})$, reaching them to three digits by depth 4,096 (Fig.~\ref{fig:fourdisk-labels}, right), while the learned classifications track the target labels (Fig.~\ref{fig:fourdisk-labels}, left). For the region task, the model chose to amplify harmonic information, with $\eta_H$ as the leading rate. 
Since the labels are built from the same normalized Laplacian's eigenvectors as the flow, this tests whether the three regimes are \emph{selectable}.

\begin{table}[t]
\vspace{-2mm}
\centering
\caption{Leading eigenvalues $\eta_S$ of the learned endpoint generators (seed 0); bold marks the largest per row, which gives the predicted regime. RQ: Rayleigh quotients after 4{,}096 Euler steps on the first held-out noise realization ($\nu_g^{\max}=0.7413$, $\nu_c^{\max}=0.9836$).}
\vspace{-2.5mm}
\label{tab:fourdisk}
\footnotesize
\begin{tabular*}{\linewidth}{@{\extracolsep{\fill}}l ccc l cc@{}}
\toprule
 & \multicolumn{3}{c}{Eigenvalue} & & \multicolumn{2}{c}{RQ} \\
\cmidrule(lr){2-4} \cmidrule(l){6-7}
Task & $\eta_H$ & $\eta_G$ & $\eta_C$ & Predicted & down & up \\
\midrule
Harmonic       & \textbf{1.493} & 0.449          & 0.339          & Harmonic & $2.5{\times}10^{-7}$ & $9.6{\times}10^{-8}$ \\
Upper-gradient & 0.520          & \textbf{0.726} & 0.540          & Gradient & $0.7412$             & $\approx 0$          \\
Upper-curl     & 0.600          & 0.385          & \textbf{0.652} & Curl     & $\approx 0$          & $0.9833$             \\
Region         & \textbf{0.562} & 0.488          & 0.302          & Harmonic & $6.6{\times}10^{-5}$ & $1.1{\times}10^{-3}$ \\
\bottomrule
\end{tabular*}
\vspace{-3mm}
\end{table}

\vspace{-1.65mm}
\subsection{Trajectory Prediction on Ocean Drifters}
\label{sec:drifters}
\vspace{-1.7mm}
We use the ocean-drifter dataset of SCoNe~\citep{roddenberry2021scone}: 200 trajectories on a fixed oriented complex around Madagascar (133 vertices, 320 edges, 186 triangles) whose 2-dimensional harmonic space reflects Madagascar and a smaller neighboring island. Each trajectory prefix is encoded as its signed net edge-traversal vector $\vx\in\mathbb R^{320}$, and the task is to predict the next vertex among its endpoint's neighbors. We train the adjacency-like and pure-Laplacian flows of \cref{sec:instantiation} with a nonlinear decoder, freeze each model, predict its regime from the learned generators, and extend propagation to depth 1,024 on held-out examples, tracking the Rayleigh-quotient pair and the feature mass in the predicted eigenspace (\cref{app:experimental-details}).

\begin{figure}[t]
\vspace{-5mm}
\centering
\includegraphics[width=\linewidth]{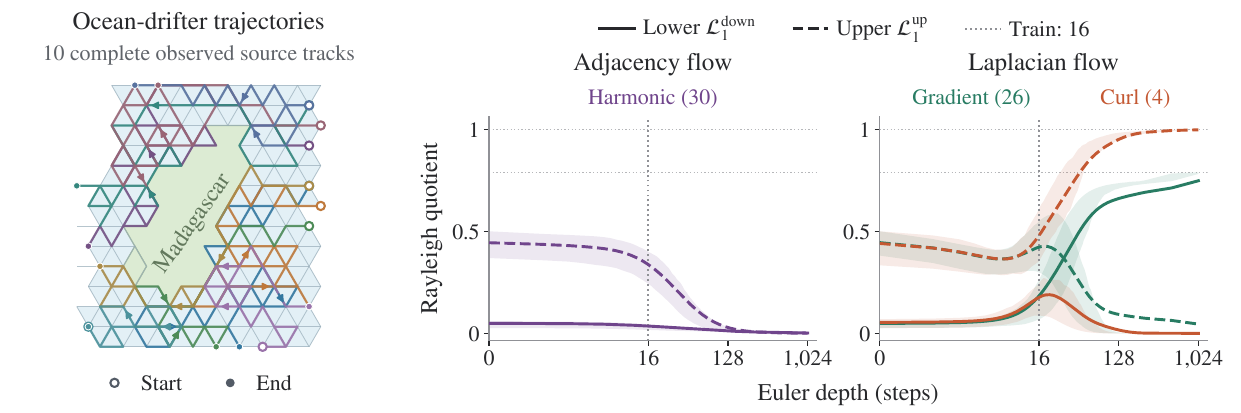}
\caption{\textbf{(left)} Sample trajectories in the ocean drifter dataset. \textbf{(right)} Frozen-flow spectral evolution on the ocean-drifter complex for adjacency-based and Laplacian flow, grouped by its predicted endpoint. Lines mark means with shaded interquartile ranges across test occurrences. Horizontal dotted lines in the right panel indicate the predicted limiting Rayleigh quotients. Adjacency-trained flows (30) amplify the harmonic space, while Laplacian flows select either gradient (26) or curl (4).}
\label{fig:drifters}
\vspace{-4mm}
\end{figure}

\paragraph{Results}
The two architectures select different regimes (Fig.~\ref{fig:drifters}). Every adjacency-like model is predicted harmonic-dominant, and both Rayleigh quotients decay toward zero along the frozen trajectories. 
The pure-Laplacian flow, whose identity-free generator cannot act on the harmonic sector (\cref{app:AvsL}), is instead predicted top-gradient in 26 models and top-curl in four, and its Rayleigh quotients approach these endpoints. 
Because drifter trajectories circulate around the two holes, the task-relevant signal has a large harmonic component~\citep{schaub2020random}, which only the adjacency-like flow can amplify. This suggests harmonic information may help on this topology, and an orientation-equivariant $\tanh$ leaves these dynamics largely intact (\cref{app:additional-ocean-drifters}).

\vspace{-1.5mm}
\section{Related Works}
\vspace{-2.5mm}

Simplicial complexes extend graphs through higher-dimensional simplices, allowing SNNs and other topological deep learning (TDL) models to learn features on higher-order structures~\citep{pmlr-v139-bodnar21a,papamarkou2024position}.
In GNNs, repeated propagation can concentrate node features in low-frequency subspaces, a phenomenon described as over-smoothing that motivates asking how depth affects features at higher simplicial ranks~\citep{oono2020expressive,cai2020oversmoothing,rusch2023oversmoothing}.
Prior works on SNNs have studied simplicial over-smoothing through decay of the unnormalized Hodge Dirichlet energy \citep{yang2025hodgeaware,einizade2025continuous}, while binarized propagation has been proposed to mitigate over-smoothing \citep{yan2025binarized}.
The Hodge decomposition has been employed to inform representations, learning objectives, and architectures that distinguish gradient, curl, and harmonic components~\citep{roddenberry2022hodgelets,battiloro2023weighted,mollers2024hodgeaware,alain2026hodgelet,bastian2026topological}.
This work extends the feature-normalized graph criterion of \citet{digiovanni2023understandingconvolutiongraphsenergies} to paired Hodge Rayleigh quotients to study spectral concentration in SNNs.
We derive conditions under which normalized edge features concentrate asymptotically in the harmonic space or the highest-frequency gradient or curl eigenspace, allowing a trained propagation operator to be diagnosed without running it to large depth, and bound the deviation from these limits at finite depth.
Further related work is in~\cref{app:graphbg}.

\vspace{-2.5mm}
\section{Concluding Remarks}
\vspace{-2.5mm}

We developed a spectral framework for characterizing feature propagation in Hodge-Laplacian-based simplicial neural networks. Using a pair of feature-normalized lower and upper Hodge Rayleigh quotients, we established three asymptotic regimes: harmonic dominance, gradient dominance, and curl dominance. 
For linear edge-feature flows with independent lower and upper channel mixing, we showed that the asymptotic behavior is governed by three competing spectral growth rates, which can be computed from the trained weights and Hodge eigenvalues to provide a post-training interpretability tool. 
We further bounded the distance to the predicted regime at finite depth, extended this bound to elementwise nonlinearities under certain conditions, and showed that the results hold at every simplicial rank.
Our experiments support these predictions, illustrate how supervision and the propagation operator influence the selected regime, and indicate that the paired Rayleigh quotients remain informative for nonlinear flows and published architectures.

\paragraph{Limitations and future work}
Our main theoretical results characterize the precise regime for linear SNN dynamics with depth-shared, symmetric channel-mixing operators, while our nonlinear guarantees require the leakage between Hodge subspaces to remain small, which holds at every depth only for contractive flows or sufficiently small inputs. Nonlinear flows that amplify their leading eigenspace, and depth-varying or non-symmetric weights, remain open for future work. In particular, analyzing our spectral framework as a dynamical system could recast the HD, GD, and CD limits as attractors, allowing us to study their stability, basins of attraction, and transitions. Finally, extending our results from simplicial complexes to more general cell complexes would broaden the framework to a larger class of higher-order relational domains.

\clearpage

\paragraph{Acknowledgements}
TB acknowledges support from the UKRI Engineering and Physical Sciences Research Council (EPSRC) through the Future Leaders Fellowship [grant number MR/Y018818/1]. 
LB was supported by the UK Royal Society through grant NIF/R1/254128. 

\paragraph{AI use statement}
In this work, we used generative AI tools for language editing, code editing, debugging, validating our synthetic data simulation, and visualization. 
All AI-assisted text and code were reviewed by the authors. 
The authors take full responsibility for the final content.

\paragraph{Ethics and reproducibility}
Our work is theoretical in nature; we thus do not anticipate any direct adverse consequences from the insights into the spectral concentration of SNNs. 
Experiments use public data, thus no human subjects or personal data are involved.
Training protocols are documented in \cref{app:experimental-details}, with code available at \url{https://github.com/irisdliao/hodge-spectral-concentration}.

\bibliography{iclr2027_conference}

\begin{thebibliography}{49}
\providecommand{\natexlab}[1]{#1}
\providecommand{\url}[1]{\texttt{#1}}
\expandafter\ifx\csname urlstyle\endcsname\relax
  \providecommand{\doi}[1]{doi: #1}\else
  \providecommand{\doi}{doi: \begingroup \urlstyle{rm}\Url}\fi

\bibitem[Alain et~al.(2026)Alain, Takao, Dong, Rieck, and Noutahi]{alain2026hodgelet}
Mathieu Alain, So~Takao, Xiaowen Dong, Bastian Rieck, and Emmanuel Noutahi.
\newblock Graph and simplicial complex prediction {G}aussian process via the {H}odgelet representations.
\newblock In \emph{Advances in Neural Information Processing Systems}, volume~39, 2026.
\newblock In press.

\bibitem[Barsbey et~al.(2025)Barsbey, Ballester, Demir, Casacuberta, Hern{\'a}ndez-Garc{\'\i}a, Pujol-Perich, Yurtseven, Escalera, Battiloro, Hajij, et~al.]{barsbey2025higher}
Melih Barsbey, Rub{\'e}n Ballester, Andac Demir, Carles Casacuberta, Pablo Hern{\'a}ndez-Garc{\'\i}a, David Pujol-Perich, Sarper Yurtseven, Sergio Escalera, Claudio Battiloro, Mustafa Hajij, et~al.
\newblock Higher-order molecular learning: The cellular transformer.
\newblock In \emph{ICLR 2025 Workshop on Generative and Experimental Perspectives for Biomolecular Design}, 2025.

\bibitem[Bastian et~al.(2026)Bastian, Leventhal, Hajij, and Birdal]{bastian2026topological}
Lennart Bastian, Samuel Leventhal, Mustafa Hajij, and Tolga Birdal.
\newblock Topological neural operators, 2026.
\newblock URL \url{https://arxiv.org/abs/2606.09806}.

\bibitem[Battiloro et~al.(2023)Battiloro, Sardellitti, Barbarossa, and Di~Lorenzo]{battiloro2023weighted}
Claudio Battiloro, Stefania Sardellitti, Sergio Barbarossa, and Paolo Di~Lorenzo.
\newblock Topological signal processing over weighted simplicial complexes.
\newblock In \emph{2023 IEEE International Conference on Acoustics, Speech and Signal Processing (ICASSP)}, pp.\  1--5. IEEE, 2023.
\newblock \doi{10.1109/ICASSP49357.2023.10094927}.

\bibitem[Bodnar et~al.(2021)Bodnar, Frasca, Wang, Otter, Montufar, Li{\'o}, and Bronstein]{pmlr-v139-bodnar21a}
Cristian Bodnar, Fabrizio Frasca, Yuguang Wang, Nina Otter, Guido~F Montufar, Pietro Li{\'o}, and Michael Bronstein.
\newblock Weisfeiler and lehman go topological: Message passing simplicial networks.
\newblock In Marina Meila and Tong Zhang (eds.), \emph{Proceedings of the 38th International Conference on Machine Learning}, volume 139 of \emph{Proceedings of Machine Learning Research}, pp.\  1026--1037. PMLR, 18--24 Jul 2021.

\bibitem[Bodnar et~al.(2022)Bodnar, Di~Giovanni, Chamberlain, Lio, and Bronstein]{bodnar2022neural}
Cristian Bodnar, Francesco Di~Giovanni, Benjamin Chamberlain, Pietro Lio, and Michael Bronstein.
\newblock Neural sheaf diffusion: A topological perspective on heterophily and oversmoothing in gnns.
\newblock \emph{Advances in Neural Information Processing Systems}, 35:\penalty0 18527--18541, 2022.

\bibitem[Bunch et~al.(2020)Bunch, You, Fung, and Singh]{bunch2020simplicial}
Eric Bunch, Qian You, Glenn Fung, and Vikas Singh.
\newblock Simplicial 2-complex convolutional neural networks.
\newblock In \emph{NeurIPS 2020 Workshop on Topological Data Analysis and Beyond}, 2020.
\newblock URL \url{https://openreview.net/forum?id=TLbnsKrt6J-}.

\bibitem[Cai \& Wang(2020)Cai and Wang]{cai2020oversmoothing}
Chen Cai and Yusu Wang.
\newblock A note on over-smoothing for graph neural networks.
\newblock \emph{arXiv preprint arXiv:2006.13318}, 2020.
\newblock \doi{10.48550/arXiv.2006.13318}.
\newblock URL \url{https://arxiv.org/abs/2006.13318}.

\bibitem[Celma(2010)]{Celma2010lastfm}
O.~Celma.
\newblock \emph{{Music Recommendation and Discovery in the Long Tail}}.
\newblock Springer, 2010.

\bibitem[Di~Giovanni et~al.(2023)Di~Giovanni, Rowbottom, Chamberlain, Markovich, and Bronstein]{digiovanni2023understandingconvolutiongraphsenergies}
Francesco Di~Giovanni, James Rowbottom, Benjamin~P. Chamberlain, Thomas Markovich, and Michael~M. Bronstein.
\newblock Understanding convolution on graphs via energies, 2023.
\newblock URL \url{https://arxiv.org/abs/2206.10991}.

\bibitem[Ebli et~al.(2020)Ebli, Defferrard, and Spreemann]{ebli2020simplicial}
Stefania Ebli, Micha{\"e}l Defferrard, and Gard Spreemann.
\newblock Simplicial neural networks.
\newblock \emph{arXiv preprint arXiv:2010.03633}, 2020.
\newblock \doi{10.48550/arXiv.2010.03633}.
\newblock URL \url{https://arxiv.org/abs/2010.03633}.

\bibitem[Eckmann(1944)]{eckmann1944harmonische}
Beno Eckmann.
\newblock {Harmonische Funktionen und Randwertaufgaben in einem Komplex}.
\newblock \emph{Commentarii Mathematici Helvetici}, 17\penalty0 (1):\penalty0 240--255, 1944.
\newblock \doi{10.1007/BF02566245}.

\bibitem[Einizade et~al.(2025)Einizade, Thanou, Malliaros, and Giraldo]{einizade2025continuous}
Aref Einizade, Dorina Thanou, Fragkiskos~D. Malliaros, and Jhony~H. Giraldo.
\newblock Continuous simplicial neural networks.
\newblock In \emph{The Thirty-ninth Annual Conference on Neural Information Processing Systems}, 2025.
\newblock URL \url{https://openreview.net/forum?id=fPPfFMVTZo}.

\bibitem[Epping et~al.(2024)Epping, Ren{\'e}, Helias, and Schaub]{epping2024donotalways}
Bastian Epping, Alexandre Ren{\'e}, Moritz Helias, and Michael~T. Schaub.
\newblock Graph neural networks do not always oversmooth.
\newblock In \emph{Advances in Neural Information Processing Systems}, volume~37, pp.\  48164--48188, 2024.
\newblock URL \url{https://proceedings.neurips.cc/paper_files/paper/2024/hash/5623c35f3ab5e2c72aeb3abce27dc28f-Abstract-Conference.html}.

\bibitem[Gilmer et~al.(2017)Gilmer, Schoenholz, Riley, Vinyals, and Dahl]{gilmer2017neural}
Justin Gilmer, Samuel~S. Schoenholz, Patrick~F. Riley, Oriol Vinyals, and George~E. Dahl.
\newblock Neural message passing for quantum chemistry.
\newblock In \emph{Proceedings of the 34th International Conference on Machine Learning}, volume~70 of \emph{Proceedings of Machine Learning Research}, pp.\  1263--1272. PMLR, 2017.
\newblock URL \url{https://proceedings.mlr.press/v70/gilmer17a.html}.

\bibitem[Grande \& Schaub(2024)Grande and Schaub]{grande2024hodge}
Vincent~P. Grande and Michael~T. Schaub.
\newblock Disentangling the spectral properties of the hodge laplacian: Not all small eigenvalues are equal.
\newblock In \emph{2024 IEEE International Conference on Acoustics, Speech and Signal Processing (ICASSP)}. IEEE, 2024.
\newblock \doi{10.1109/ICASSP48485.2024.10446051}.

\bibitem[Hajij et~al.(2022)Hajij, Zamzmi, Papamarkou, Miolane, Guzm{\'a}n-S{\'a}enz, Ramamurthy, Birdal, Dey, Mukherjee, Samaga, Livesay, Walters, Rosen, and Schaub]{hajij2022topological}
Mustafa Hajij, Ghada Zamzmi, Theodore Papamarkou, Nina Miolane, Aldo Guzm{\'a}n-S{\'a}enz, Karthikeyan~Natesan Ramamurthy, Tolga Birdal, Tamal~K. Dey, Soham Mukherjee, Shreyas~N. Samaga, Neal Livesay, Robin Walters, Paul Rosen, and Michael~T. Schaub.
\newblock Topological deep learning: Going beyond graph data, 2022.
\newblock URL \url{https://arxiv.org/abs/2206.00606}.

\bibitem[Hajij et~al.(2026)Hajij, Bastian, Osentoski, Kabaria, Davenport, Cherukuri, Kocheemoolayil, Shahmansouri, Lew, Papamarkou, et~al.]{hajij2026copresheaf}
Mustafa Hajij, Lennart Bastian, Sarah Osentoski, Hardik Kabaria, John Davenport, Balaji Cherukuri, Joseph Kocheemoolayil, Nastaran Shahmansouri, Adrian Lew, Theodore Papamarkou, et~al.
\newblock Copresheaf topological neural networks: A generalized deep learning framework.
\newblock \emph{Advances in Neural Information Processing Systems}, 38:\penalty0 149759--149803, 2026.

\bibitem[Hansen \& Gebhart(2020)Hansen and Gebhart]{hansen2020sheaf}
Jakob Hansen and Thomas Gebhart.
\newblock Sheaf neural networks.
\newblock \emph{arXiv preprint arXiv:2012.06333}, 2020.

\bibitem[Hatcher(2002)]{hatcher2002algebraic}
Allen Hatcher.
\newblock \emph{Algebraic Topology}.
\newblock Cambridge University Press, 2002.
\newblock URL \url{https://pi.math.cornell.edu/~hatcher/AT/ATpage.html}.

\bibitem[Huang \& Birdal(2026)Huang and Birdal]{huang2026hog}
Yiming Huang and Tolga Birdal.
\newblock Hog-diff: Higher-order guided diffusion for graph generation.
\newblock In \emph{International Conference on Learning Representations}, volume 2026, pp.\  111572--111614, 2026.

\bibitem[Huang et~al.(2026)Huang, Zeng, Dwivedi, Foti, Wang, Leskovec, and Birdal]{huang2026higher}
Yiming Huang, Yujie Zeng, Vijay~Prakash Dwivedi, Simone Foti, Jianmin Wang, Jure Leskovec, and Tolga Birdal.
\newblock Higher-order molecular grammars for generative and foundation models in chemistry.
\newblock \emph{arXiv preprint arXiv:2610.02186}, 2026.

\bibitem[Jia et~al.(2019)Jia, Schaub, Segarra, and Benson]{jia2019graph}
Junteng Jia, Michael~T Schaub, Santiago Segarra, and Austin~R Benson.
\newblock Graph-based semi-supervised \& active learning for edge flows.
\newblock In \emph{Proceedings of the 25th ACM SIGKDD international conference on knowledge discovery \& data mining}, pp.\  761--771, 2019.

\bibitem[Keriven(2022)]{keriven2022not}
Nicolas Keriven.
\newblock Not too little, not too much: a theoretical analysis of graph (over) smoothing.
\newblock \emph{Advances in Neural Information Processing Systems}, 35:\penalty0 2268--2281, 2022.

\bibitem[Kipf \& Welling(2017)Kipf and Welling]{kipf2017semisupervised}
Thomas~N. Kipf and Max Welling.
\newblock Semi-supervised classification with graph convolutional networks.
\newblock In \emph{International Conference on Learning Representations (ICLR)}, 2017.
\newblock URL \url{https://arxiv.org/abs/1609.02907}.

\bibitem[Korkmaz et~al.(2025)Korkmaz, Nuwagira, Coskunuzer, and Birdal]{korkmaz2025cumperlay}
Caner Korkmaz, Brighton Nuwagira, Baris Coskunuzer, and Tolga Birdal.
\newblock Cumperlay: Learning cubical multiparameter persistence vectorizations.
\newblock In \emph{Proceedings of the IEEE/CVF International Conference on Computer Vision}, pp.\  27084--27094, 2025.

\bibitem[Krahn et~al.(2026)Krahn, Bastian, Garg, Schuller, and Birdal]{krahn2026collapsed}
Maximilian Krahn, Lennart Bastian, Vikas~K Garg, Bj{\"o}rn Schuller, and Tolga Birdal.
\newblock Collapsed effective operators for higher-order structures.
\newblock In \emph{Forty-third International Conference on Machine Learning}, 2026.
\newblock URL \url{https://openreview.net/forum?id=arc2pWtZLN}.

\bibitem[Li et~al.(2018)Li, Han, and Wu]{li2018deeper}
Qimai Li, Zhichao Han, and Xiao-Ming Wu.
\newblock Deeper insights into graph convolutional networks for semi-supervised learning.
\newblock In \emph{Proceedings of the AAAI Conference on Artificial Intelligence}, 2018.
\newblock URL \url{https://arxiv.org/abs/1801.07606}.

\bibitem[Lim(2020)]{lim2020hodge}
Lek-Heng Lim.
\newblock Hodge laplacians on graphs.
\newblock \emph{SIAM Review}, 62\penalty0 (3):\penalty0 685--715, 2020.
\newblock \doi{10.1137/18M1223101}.
\newblock URL \url{https://arxiv.org/abs/1507.05379}.

\bibitem[M{\"o}llers et~al.(2024)M{\"o}llers, Immer, Fortuin, and Isufi]{mollers2024hodgeaware}
Alexander M{\"o}llers, Alexander Immer, Vincent Fortuin, and Elvin Isufi.
\newblock {H}odge-aware contrastive learning.
\newblock In \emph{2024 IEEE International Conference on Acoustics, Speech and Signal Processing (ICASSP)}, pp.\  9746--9750. IEEE, 2024.
\newblock \doi{10.1109/ICASSP48485.2024.10447015}.

\bibitem[Oono \& Suzuki(2020)Oono and Suzuki]{oono2020expressive}
Kenta Oono and Taiji Suzuki.
\newblock Graph neural networks exponentially lose expressive power for node classification.
\newblock In \emph{International Conference on Learning Representations}, 2020.
\newblock URL \url{https://openreview.net/forum?id=S1ldO2EFPr}.

\bibitem[Papamarkou et~al.(2024)Papamarkou, Birdal, Bronstein, Carlsson, Curry, Gao, Hajij, Kwitt, Lio, Di~Lorenzo, Maroulas, Miolane, Nasrin, Natesan~Ramamurthy, Rieck, Scardapane, Schaub, Veli\v{c}kovi\'{c}, Wang, Wang, Wei, and Zamzmi]{papamarkou2024position}
Theodore Papamarkou, Tolga Birdal, Michael~M. Bronstein, Gunnar~E. Carlsson, Justin Curry, Yue Gao, Mustafa Hajij, Roland Kwitt, Pietro Lio, Paolo Di~Lorenzo, Vasileios Maroulas, Nina Miolane, Farzana Nasrin, Karthikeyan Natesan~Ramamurthy, Bastian Rieck, Simone Scardapane, Michael~T. Schaub, Petar Veli\v{c}kovi\'{c}, Bei Wang, Yusu Wang, Guowei Wei, and Ghada Zamzmi.
\newblock Position: Topological deep learning is the new frontier for relational learning.
\newblock In \emph{Proceedings of the 41st International Conference on Machine Learning}, volume 235 of \emph{Proceedings of Machine Learning Research}, pp.\  39529--39555. PMLR, 2024.
\newblock URL \url{https://proceedings.mlr.press/v235/papamarkou24a.html}.

\bibitem[Papillon et~al.(2025)Papillon, Bernardez, Battiloro, and Miolane]{papillon2025topotune}
Mathilde Papillon, Guillermo Bernardez, Claudio Battiloro, and Nina Miolane.
\newblock Topotune: A framework for generalized combinatorial complex neural networks.
\newblock In \emph{Forty-second International Conference on Machine Learning}, 2025.
\newblock URL \url{https://openreview.net/forum?id=S5njonQdBf}.

\bibitem[Roddenberry et~al.(2021)Roddenberry, Glaze, and Segarra]{roddenberry2021scone}
T.~Mitchell Roddenberry, Nicholas Glaze, and Santiago Segarra.
\newblock Principled simplicial neural networks for trajectory prediction.
\newblock In \emph{Proceedings of the 38th International Conference on Machine Learning}, volume 139 of \emph{Proceedings of Machine Learning Research}, pp.\  9020--9029. PMLR, 2021.
\newblock URL \url{https://proceedings.mlr.press/v139/roddenberry21a.html}.

\bibitem[Roddenberry et~al.(2022)Roddenberry, Frantzen, Schaub, and Segarra]{roddenberry2022hodgelets}
T.~Mitchell Roddenberry, Florian Frantzen, Michael~T. Schaub, and Santiago Segarra.
\newblock {H}odgelets: Localized spectral representations of flows on simplicial complexes.
\newblock In \emph{2022 IEEE International Conference on Acoustics, Speech and Signal Processing (ICASSP)}, pp.\  5922--5926. IEEE, 2022.
\newblock \doi{10.1109/ICASSP43922.2022.9747203}.

\bibitem[Rusch et~al.(2023)Rusch, Bronstein, and Mishra]{rusch2023oversmoothing}
T.~Konstantin Rusch, Michael~M. Bronstein, and Siddhartha Mishra.
\newblock A survey on oversmoothing in graph neural networks.
\newblock \emph{arXiv preprint arXiv:2303.10993}, 2023.
\newblock \doi{10.48550/arXiv.2303.10993}.
\newblock URL \url{https://arxiv.org/abs/2303.10993}.

\bibitem[Schaub et~al.(2020)Schaub, Benson, Horn, Lippner, and Jadbabaie]{schaub2020random}
Michael~T Schaub, Austin~R Benson, Paul Horn, Gabor Lippner, and Ali Jadbabaie.
\newblock Random walks on simplicial complexes and the normalized hodge 1-laplacian.
\newblock \emph{SIAM Review}, 62\penalty0 (2):\penalty0 353--391, 2020.

\bibitem[Taha et~al.(2025)Taha, Chapman, Eidi, Devriendt, and Montufar]{taha2025demystifying}
Diaaeldin Taha, James Chapman, Marzieh Eidi, Karel Devriendt, and Guido Montufar.
\newblock Demystifying topological message-passing with relational structures: A case study on oversquashing in simplicial message-passing.
\newblock In \emph{The Thirteenth International Conference on Learning Representations}, 2025.
\newblock URL \url{https://openreview.net/forum?id=QC2qE1tcmd}.

\bibitem[Telyatnikov et~al.(2024)Telyatnikov, Bernardez, Montagna, Hajij, Carrasco, Vasylenko, Papillon, Zamzmi, Schaub, Verhellen, et~al.]{telyatnikov2024topobench}
Lev Telyatnikov, Guillermo Bernardez, Marco Montagna, Mustafa Hajij, Martin Carrasco, Pavlo Vasylenko, Mathilde Papillon, Ghada Zamzmi, Michael~T Schaub, Jonas Verhellen, et~al.
\newblock Topobench: A framework for benchmarking topological deep learning.
\newblock \emph{arXiv preprint arXiv:2406.06642}, 2024.

\bibitem[Vigan{\`o} et~al.(2026)Vigan{\`o}, Birdal, Schaub, and Barahona]{vigano2026roottoleafpathrandomwalks}
Francesco Vigan{\`o}, Tolga Birdal, Michael~T. Schaub, and Mauricio Barahona.
\newblock Root-to-leaf path random walks, normalized hodge laplacians, and cheeger inequalities on simplicial complexes.
\newblock \emph{arXiv preprint arXiv:2604.27241}, 2026.

\bibitem[Wang et~al.(2026)Wang, Kim, Birdal, Navab, and Bastian]{danjun2026topoor}
Tony~Danjun Wang, Ka~Young Kim, Tolga Birdal, Nassir Navab, and Lennart Bastian.
\newblock Topoor: A unified topological scene representation for the operating room.
\newblock \emph{arXiv e-prints}, pp.\  arXiv--2603, 2026.

\bibitem[Yan \& Kuruoglu(2025)Yan and Kuruoglu]{yan2025binarized}
Yi~Yan and Ercan~Engin Kuruoglu.
\newblock Binarized simplicial convolutional neural networks.
\newblock \emph{Neural Networks}, 183:\penalty0 106928, 2025.
\newblock \doi{10.1016/j.neunet.2024.106928}.

\bibitem[Yang \& Isufi(2023)Yang and Isufi]{yang2023sccnn}
Maosheng Yang and Elvin Isufi.
\newblock Convolutional learning on simplicial complexes.
\newblock \emph{arXiv preprint arXiv:2301.11163}, 2023.
\newblock \doi{10.48550/arXiv.2301.11163}.
\newblock URL \url{https://arxiv.org/abs/2301.11163}.

\bibitem[Yang et~al.(2022{\natexlab{a}})Yang, Isufi, and Leus]{yang2022scnn}
Maosheng Yang, Elvin Isufi, and Geert Leus.
\newblock Simplicial convolutional neural networks.
\newblock In \emph{2022 IEEE International Conference on Acoustics, Speech and Signal Processing (ICASSP)}, pp.\  8847--8851. IEEE, 2022{\natexlab{a}}.
\newblock \doi{10.1109/ICASSP43922.2022.9746017}.

\bibitem[Yang et~al.(2022{\natexlab{b}})Yang, Isufi, Schaub, and Leus]{yang2022simplicialfilters}
Maosheng Yang, Elvin Isufi, Michael~T. Schaub, and Geert Leus.
\newblock Simplicial convolutional filters.
\newblock \emph{IEEE Transactions on Signal Processing}, 70:\penalty0 4633--4648, 2022{\natexlab{b}}.
\newblock \doi{10.1109/TSP.2022.3207045}.

\bibitem[Yang et~al.(2025)Yang, Leus, and Isufi]{yang2025hodgeaware}
Maosheng Yang, Geert Leus, and Elvin Isufi.
\newblock Hodge-aware convolutional learning on simplicial complexes.
\newblock \emph{Transactions on Machine Learning Research}, 2025.
\newblock ISSN 2835-8856.
\newblock URL \url{https://openreview.net/forum?id=Nm5sp09Q25}.

\bibitem[Ying et~al.(2018)Ying, He, Chen, Eksombatchai, Hamilton, and Leskovec]{ying2018graph}
Rex Ying, Ruining He, Kaifeng Chen, Pong Eksombatchai, William~L. Hamilton, and Jure Leskovec.
\newblock Graph convolutional neural networks for web-scale recommender systems.
\newblock In \emph{Proceedings of the 24th ACM SIGKDD International Conference on Knowledge Discovery and Data Mining}, pp.\  974--983, 2018.
\newblock \doi{10.1145/3219819.3219890}.
\newblock URL \url{https://arxiv.org/abs/1806.01973}.

\bibitem[Zhang et~al.(2026)Zhang, Deidda, Higham, and Tudisco]{zhang2026measuring}
Kaicheng Zhang, Piero Deidda, Desmond Higham, and Francesco Tudisco.
\newblock Are we measuring oversmoothing in graph neural networks correctly?
\newblock In \emph{International Conference on Learning Representations}, 2026.
\newblock URL \url{https://openreview.net/forum?id=r3DOISnvHD}.

\bibitem[Zhuo et~al.(2025)Zhuo, Wang, Ma, and Wang]{zhuo2025escape}
Zhijian Zhuo, Yifei Wang, Jinwen Ma, and Yisen Wang.
\newblock Graph neural networks (with proper weights) can escape oversmoothing.
\newblock In Vu~Nguyen and Hsuan-Tien Lin (eds.), \emph{Proceedings of the 16th Asian Conference on Machine Learning}, volume 260 of \emph{Proceedings of Machine Learning Research}, pp.\  17--32. PMLR, 2025.
\newblock URL \url{https://proceedings.mlr.press/v260/zhuo25a.html}.

\end{thebibliography}
\bibliographystyle{iclr2027_conference}
\newpage

\clearpage
\appendix

\section*{Appendix}
We now supplement our main paper by providing additional background, the deferred proofs, and extensions of the results in the main paper.
\vspace{1mm}

\addtocontents{toc}{\protect\setcounter{tocdepth}{2}}
\setcounter{tocdepth}{-1}
\begingroup
\makeatletter
\small
\setlength{\parskip}{0pt}
\renewcommand*\l@section[2]{%
  \addpenalty{-\@highpenalty}\addvspace{3pt plus 1pt}%
  {\bfseries\@dottedtocline{1}{0em}{1.6em}{#1}{#2}}}
\renewcommand*\l@subsection{\@dottedtocline{2}{1.6em}{2.2em}}
\@starttoc{toc}
\makeatother
\endgroup
\setcounter{tocdepth}{3}

\section{Additional Related Work and Graph Background}
\label{app:graphbg}
\subsection{Over-smoothing and over-sharpening on graphs}
\noindent Much research has been devoted to understanding message passing in neural networks on graph-structured data, and more recently on simplicial and cellular complexes~\citep{taha2025demystifying}. In Graph Convolutional Networks (GCNs), node features are transformed by a shared, learned mapping and propagated across edges~\citep{kipf2017semisupervised}. Despite their success on various tasks involving graph-structured data, GCNs tend to struggle on \emph{heterophilic graphs} and suffer from \emph{over-smoothing} as the number of layers in the neural network increases. Heterophily (homophily) is a property of labeled graphs that measures the tendency for nodes with different (the same) labels to share edges. Over-smoothing refers to node features becoming increasingly indistinguishable with depth because each propagation step acts as a form of Laplacian smoothing~\citep{li2018deeper}, so that features converge exponentially to a subspace determined by the connected components of the graph~\citep{oono2020expressive} and their Dirichlet energy decays to zero~\citep{cai2020oversmoothing, rusch2023oversmoothing}. Subsequent works refine and further characterize these dynamics. In particular, \citet{keriven2022not} show that a moderate amount of smoothing can be beneficial before it becomes harmful; \citet{epping2024donotalways} show, in the infinite-width (Gaussian-process) limit, that GCNs initialized with sufficiently large weight variance enter a non-over-smoothing phase in which node features remain informative at depth; and \citet{zhuo2025escape} attribute over-smoothing to inadequately learned weights rather than to the architecture itself, showing that suitable weight matrices allow vanilla GNNs to avoid it even after infinitely many propagation steps. Other measures of over-smoothing, such as the numerical rank of the feature matrix, have also been proposed \citet{zhang2026measuring}. \citet{digiovanni2023understandingconvolutiongraphsenergies} investigate these behaviors through a dynamical systems framework, where node features evolve according to the gradient flow of an energy functional related to the Dirichlet energy. We restate their framework here.

\paragraph{The graph Laplacian and adjacency}
\noindent For a graph $G=(V,E)$ with normalized adjacency $\tilde{\mA}$, the normalized graph Laplacian is $\mDelta = \mI - \tilde{\mA}$. Low frequencies of $\mDelta$ correspond to smooth signals over the graph, while high frequencies correspond to rapid variation between neighboring nodes. Many graph-based message-passing architectures use these operators to propagate information across edges; GCNs in particular use the normalized adjacency.

\paragraph{Dirichlet energy and gradient flows}
\noindent The graph Dirichlet energy measures the smoothness of the feature signal across edges:
$$E_{\mathrm{Dir}}(\mF) = \mathrm{trace}(\mF^\top \mDelta \mF).$$
\noindent Since low graph frequencies correspond to smoother signals, graph convolutions are often seen as low-pass filters that progressively reduce the Dirichlet energy and smooth features. To quantify and study this behavior, \citet{digiovanni2023understandingconvolutiongraphsenergies} introduce a parametric energy functional incorporating symmetric learnable channel-mixing weights $\mOmega, \mW$:
$$E_{\theta}(\mF) = \sum_i \langle \vf_i, \mOmega \vf_i \rangle - \sum_{i,j} A_{ij}\langle \vf_i, \mW \vf_j \rangle + \phi^0(\mF, \mF(0)).$$
\noindent Positive eigenvalues of $\mW$ induce attractive forces (smoothing), while negative eigenvalues induce repulsive forces (sharpening). Graph convolutional networks can be interpreted as gradient flows of this functional:
$$\dot{\mF}(t) = -\mF(t)\mOmega + \mA \mF(t)\mW - \mF(0)\tilde{\mW}.$$
\noindent The above energy formulation allows for both over-smoothing and the opposite behavior, over-sharpening, which is noteworthy since GCNs were thought to only exhibit over-smoothing.

\paragraph{Over-smoothing and over-sharpening: LFD and HFD}
\noindent For completeness, we restate the criteria of \citet[Def.~4.2]{digiovanni2023understandingconvolutiongraphsenergies}. For nonzero features $\mF(t)$, with $t$ denoting continuous depth or a discrete layer index, \emph{low-frequency dominance (LFD)} is characterized by
$$\lim_{t\to\infty} \frac{E_{\mathrm{Dir}}(\mF(t))}{\|\mF(t)\|_F^2} = 0.$$
\noindent Conversely, \emph{high-frequency dominance (HFD)} indicates over-sharpening:
$$\lim_{t\to\infty} \frac{E_{\mathrm{Dir}}(\mF(t))}{\|\mF(t)\|_F^2} = \lambda_{n-1},$$
where $\|\cdot\|_F$ is the Frobenius norm and $\lambda_{n-1}$ is the largest eigenvalue of $\mDelta$. These criteria give the spectral characterizations of over-smoothing and over-sharpening discussed in \cref{prop:regimeprops}; at rank $0$ on a connected graph, LFD coincides with the harmonic-dominant regime of \cref{prop:regimeprops}.

\noindent For the simplified continuous-time graph convolution considered by Di Giovanni et al.,
\[
\dot \mF(t) = \mA \mF(t) \mW,
\qquad
\mA = \mD^{-1/2} \mA_{\mathrm{graph}} \mD^{-1/2} = \mI-\mDelta,
\]
assume that the graph is simple, undirected, and connected, and that the channel-mixing matrix $\mW$ is symmetric, with eigenvalues
\[
\mu_0 < \mu_1 \le \cdots \le \mu_{d-1}.
\]
Let
\[
0=\lambda_0\le \cdots\le\lambda_{n-1}
\]
denote the eigenvalues of the normalized graph Laplacian $\mDelta$. Then, for almost every initial feature matrix $\mF(0)$,
\[
|\mu_0|(\lambda_{n-1}-1)>\mu_{d-1}
\]
implies HFD dynamics, whereas
\[
|\mu_0|(\lambda_{n-1}-1)<\mu_{d-1}
\]
implies LFD dynamics~\citep[Thm.~4.3]{digiovanni2023understandingconvolutiongraphsenergies}. \Cref{app:graph} shows that this dichotomy is the rank-$0$ case of our trichotomy under the root-to-leaf normalization.

\subsection{Broader view over topological deep learning approaches}
\label{app:tdlrelated}

\paragraph{Architectures and propagation operators}
Early simplicial neural networks use Hodge-based convolutions to
process signals supported on simplices
\citep{ebli2020simplicial,bunch2020simplicial}.
The broader architectural landscape includes sheaf-based models
\citep{hansen2020sheaf,bodnar2022neural} and copresheaf formulations
\citep{hajij2026copresheaf}.
Other approaches investigate cellular transformers for molecular
learning \citep{barsbey2025higher}, neural operators on cell complexes
\citep{bastian2026topological}, and collapsed effective operators
that encode higher-order structure in operators acting on vertices
\citep{krahn2026collapsed}.
\citet{taha2025demystifying} view simplicial and cellular complexes
as relational structures, so that simplicial message passing becomes
an instance of relational message passing, and use this view to extend
graph-based sensitivity and curvature analyses of over-squashing, as
well as graph rewiring methods, to simplicial networks.

\paragraph{Representations, applications, and frameworks}
Complementary work incorporates topology through learned
representations and task-specific models.
CuMPerLay learns vectorizations of cubical multiparameter
persistence \citep{korkmaz2025cumperlay}, while HOG-Diff uses
higher-order topology to guide diffusion-based graph generation
\citep{huang2026hog,huang2026higher}.
TopoOR uses combinatorial complexes to represent interactions
in operating-room scenes \citep{danjun2026topoor}.
Furthermore TopoTune provides a framework for
generalized combinatorial complex neural networks
\citep{papillon2025topotune} and TopoBench provides a modular
framework for benchmarking topological deep learning
\citep{telyatnikov2024topobench}.

\section{The root-to-leaf normalization}
\label{app:r2l}

\noindent Most existing simplicial convolutional architectures are analyzed with unnormalized Hodge Laplacians, whereas the graph convolutional networks studied by \citet{kipf2017semisupervised} and \citet{digiovanni2023understandingconvolutiongraphsenergies} propagate with the symmetrically normalized adjacency. Since there is no canonical choice of normalization for higher-order Laplacians, we choose one that preserves the Hodge structure on which the analysis of \cref{sec:theory} rests: the root-to-leaf normalization of \citet{vigano2026roottoleafpathrandomwalks}, which produces normalized Hodge Laplacians whose spectra are bounded uniformly across all complexes, the higher-order analogue of the $[0,2]$ bound for the normalized graph Laplacian.

\subsection{Definition and basic properties}
\label{app:r2l-def}
\noindent For a $k$-face $\sigma$, the \emph{leaf-path function} $\mathrm{LP}(\sigma)$ is the number of ascending chains $\sigma=\sigma_k\subset\sigma_{k+1}\subset\cdots\subset\sigma_m$ of faces of consecutive dimensions that end at a maximal face of $X$, and the \emph{root-path function} $\mathrm{RP}(\sigma)$ is the number of descending chains from $\sigma$ to a vertex; since such a chain removes one vertex at a time, $\mathrm{RP}(\sigma)=(k+1)!$ on every $k$-face. Equivalently, $\mathrm{LP}(\sigma)=\sum_{\phi}(\dim\phi-\dim\sigma)!$, the sum running over the maximal faces $\phi\supseteq\sigma$ \citep[Lemma~2.1]{vigano2026roottoleafpathrandomwalks}. We fix for each dimension $k$ the diagonal, strictly positive weight matrix $\mM_k\in\mathbb R^{N_k\times N_k}$ (denoted $\mW_k$ in \citet{vigano2026roottoleafpathrandomwalks}; we write $\mM_k$ to avoid a clash with the channel weights) whose $(\sigma,\sigma)$ entry is the ratio of the leaf-path to the root-path function,
$$(\mM_k)_{\sigma\sigma} = H(\sigma) = \frac{\mathrm{LP}(\sigma)}{\mathrm{RP}(\sigma)} = \frac{\mathrm{LP}(\sigma)}{(k+1)!}.$$
We define the \emph{normalized} boundary operators by conjugating the combinatorial boundaries with the square roots of these weights,
$$\hat{\mB}_k = \mM_{k-1}^{-1/2}\, \mB_k\, \mM_k^{1/2},$$
equivalently the normalized coboundary operators $\delta_{k-1} = \hat{\mB}_k^\top = \mM_k^{1/2}\,\partial_{k-1}^*\, \mM_{k-1}^{-1/2}$ of~\citet{vigano2026roottoleafpathrandomwalks}, with $\partial_{k-1}^* = \mB_k^\top$.

\paragraph{Hodge preservation.} When multiplying boundary matrices, the inner weight matrices cancel and the combinatorial homology is preserved:
$$\hat{\mB}_k \hat{\mB}_{k+1} = \big(\mM_{k-1}^{-1/2}\mB_k \mM_k^{1/2}\big)\big(\mM_k^{-1/2}\mB_{k+1}\mM_{k+1}^{1/2}\big) = \mM_{k-1}^{-1/2}(\mB_k \mB_{k+1})\mM_{k+1}^{1/2} = 0.$$

\noindent The \emph{normalized} Hodge Laplacians are then
$$\mcL_k^{\mathrm{down}} = \hat{\mB}_k^\top \hat{\mB}_k = \mM_k^{1/2}\mB_k^\top \mM_{k-1}^{-1}\mB_k \mM_k^{1/2},$$
$$\mcL_k^{\mathrm{up}} = \hat{\mB}_{k+1}\hat{\mB}_{k+1}^\top = \mM_k^{-1/2}\mB_{k+1}\mM_{k+1}\mB_{k+1}^\top \mM_k^{-1/2},$$
which inherit properties from combinatorial Hodge theory:
\begin{itemize}
    \item[\textbf{(NH1)}] $\mcL_k^{\mathrm{down}}$ and $\mcL_k^{\mathrm{up}}$ are symmetric and positive semi-definite; and $\hat{\mB}_k \hat{\mB}_{k+1}=0$ gives $\mcL_k^{\mathrm{down}}\mcL_k^{\mathrm{up}}=0$, so they are simultaneously diagonalizable, admit a normalized Hodge decomposition, and have harmonic space isomorphic to the $k$-th cohomology~\citep{vigano2026roottoleafpathrandomwalks}.
    \item[\textbf{(NH5)}] Their spectra are bounded uniformly over all complexes by $1$: $\mathrm{Spec}(\mcL_k^{\mathrm{down}}),\, \mathrm{Spec}(\mcL_k^{\mathrm{up}}) \subseteq [0,1]$~\citep{vigano2026roottoleafpathrandomwalks}.
\end{itemize}

\subsection{Why root-to-leaf normalization?}
\label{app:normalization-alternatives}
Normalization fixes the spectral scale of the Hodge operators, allowing frequency
responses and asymptotic thresholds to be compared across complexes. We use the
root-to-leaf normalization of~\citet{vigano2026roottoleafpathrandomwalks}
because it preserves the basic Hodge structure (in particular,
$\hat \mB_k\hat \mB_{k+1}=0$, the Hodge decomposition, and the nonzero spectral
correspondence between adjacent up- and down-Laplacians) while producing symmetric positive-semidefinite operators with spectra uniformly contained in $[0,1]$. Random-walk normalizations such as that of~\citet{schaub2020random} instead arise naturally in a degree-weighted geometry (and admit symmetric representatives after a change of coordinates). The root-to-leaf construction is therefore more convenient here due to its symmetry, which lets us use standard Euclidean gradient flows and Rayleigh
quotients for our analysis directly. Because the spectra of all ranks lie in the same interval $[0,1]$, the regime conditions of \cref{thm:trichotomy} and the limits of the Rayleigh-quotient pair are moreover on a common scale across ranks and across complexes; with unnormalized operators the spectra of different ranks and different complexes need not be comparable, so a notion of dominance ``at maximal frequency'' would have no complex-independent meaning.

\subsection{Recovery of the graph setting}
\label{app:graph}
\noindent When the simplicial complex is a connected graph $G=(V,E)$ viewed as a one-dimensional complex, every edge is a maximal face, so $\mathrm{LP}(e)=1$ and $\mathrm{LP}(v)=\deg(v)$, and the root-to-leaf weights are
\[
\mM_0=\mD,
\qquad
\mM_1=\tfrac12 \mI,
\]
with $\mD$ the degree matrix. Hence
\[
\mcL_0^{\mathrm{up}}
=\hat\mB_1\hat\mB_1^\top
=\tfrac12 \mD^{-1/2}\mB_1\mB_1^\top \mD^{-1/2}
=\tfrac12\mDelta,
\qquad
\mDelta=\mI-\mD^{-1/2}\mA_{\mathrm{graph}}\mD^{-1/2},
\]
is one half of the symmetric normalized graph Laplacian, and the adjacency-like operator of \cref{sec:instantiation} is
\[
\mcA_0^{\mathrm{up}}
=
\mI-2\mcL_0^{\mathrm{up}}
=
\mI-\mDelta
=
\mD^{-1/2}\mA_{\mathrm{graph}}\mD^{-1/2}.
\]
At rank $0$ there is no lower term, since $\mB_0=0$, so the adjacency-like flow of \cref{sec:instantiation} has a single channel matrix $\mW:=\mW_{0,\mathrm{up}}$ (in the notation of \cref{sec:trichotomy}, $\mW_0=\mW$ and $\mW_2=-2\mW$) and reads
\[
\dot \mF_0
=
\mD^{-1/2}\mA_{\mathrm{graph}}\mD^{-1/2}\,\mF_0\,\mW,
\]
which is exactly the simplified graph convolution studied by
\citet{digiovanni2023understandingconvolutiongraphsenergies} and restated in \cref{app:graphbg}.

\noindent The spectral conditions reduce accordingly. If
$0=\lambda_0\leq\cdots\leq\lambda_{n-1}$
are the eigenvalues of $\mDelta$, the root-to-leaf frequencies of the rank-$0$ upper family are $\nu=\lambda/2$, so that $1-2\nu=1-\lambda$, and the channel generator \eqref{eq:mode-ode} of a mode of frequency $\nu$ is $\mW_0+\nu\mW_2=(1-2\nu)\mW=(1-\lambda)\mW$. The harmonic generator is therefore $\mW$ and the top-frequency generator is $(1-\lambda_{n-1})\mW$. Since $G$ is connected with $n\ge2$ nodes, $\lambda_{n-1}>1$: the eigenvalues of $\mDelta$ sum to $\operatorname{Tr}\mDelta=n$ while $\lambda_0=0$, so the largest is at least $n/(n-1)$. Writing
$\mu_0<\mu_1\leq\cdots\leq\mu_{d-1}$
for the eigenvalues of $\mW\ne0$, the two endpoint rates \eqref{eq:rates} are therefore
\[
\rho_H=\lambda_{\max}(\mW)=\mu_{d-1},
\qquad
\rho_{\mathrm{top}}
=
\lambda_{\max}\bigl((1-\lambda_{n-1})\mW\bigr)
=
(\lambda_{n-1}-1)\,(-\mu_0).
\]
If $\mu_0\ge0$ then $\rho_{\mathrm{top}}\le0<\rho_H$ and the flow is LFD. If $\mu_0<0$ then $\rho_{\mathrm{top}}=|\mu_0|(\lambda_{n-1}-1)$, and the comparison of $|\mu_0|(\lambda_{n-1}-1)$ with $\mu_{d-1}$ recovers the LFD/HFD dichotomy of
\citet[Thm.~4.3]{digiovanni2023understandingconvolutiongraphsenergies}: on a connected graph the harmonic space of $\mDelta$ is $\operatorname{span}(\mD^{1/2}\vone)$, the degree-scaled constant signal (constant only for regular graphs), so harmonic dominance is LFD, and dominance of the top upper frequency is HFD.

\noindent Thus we extend the low-/high frequency dynamic studied by \citet{digiovanni2023understandingconvolutiongraphsenergies}: our framework both allows us to recover those dynamics at rank 0 and observe distinct behavior at rank 1, i.e. the harmonic, top-gradient, and top-curl concentration trichotomy. In particular, unlike low-frequency concentration in the graph case, harmonic dominance is not collapse to a constant representation because it retains information on the global topological holes.
\section{Connections to existing simplicial architectures}
\label{app:instantiations}
We compare hidden propagation layers using the combinatorial boundary maps and Hodge Laplacians of \cref{sec:preliminaries}. 
When comparing architectures, we assume that activations and feature sizes are matched when ignoring task-specific prediction heads. 
Normalized variants require compatible boundary maps and symmetric lower and upper Laplacians whose product is zero, which the root-to-leaf operators of \cref{app:r2l} provide.

\subsection{A common layer formulation}
\begin{definition}[Simplicial neural network layer]
\label{def:snn}
Let $\mF_k^{(m)}\in\mathbb R^{N_k\times d_m}$ be the rank-$k$ features at layer $m$, $\sigma$ an elementwise activation, and $T_d,T_u\ge0$ filter degrees. 
Omit the layer index on the weights, and consider
\begin{equation}
\label{eq:polynomial-layer}
\begin{aligned}
\mF_k^{(m+1)}=\sigma\Bigl(
&\mF_k^{(m)}\mW_{k,0}
+\sum_{r=1}^{T_d}(\mL_k^{\mathrm{down}})^r\mF_k^{(m)}\mW_{k,d,r}
+\sum_{s=1}^{T_u}(\mL_k^{\mathrm{up}})^s\mF_k^{(m)}\mW_{k,u,s}\\
&+\sum_{r=0}^{T_d}(\mL_k^{\mathrm{down}})^r\mB_k^\top\mF_{k-1}^{(m)}\mathbf V_{k,d,r}
+\sum_{s=0}^{T_u}(\mL_k^{\mathrm{up}})^s\mB_{k+1}\mF_{k+1}^{(m)}\mathbf V_{k,u,s}\Bigr),
\end{aligned}
\end{equation}
with learnable channel matrices $\mW,\mathbf V\in\mathbb R^{d_m\times d_{m+1}}$. Nonexistent rank inputs are omitted.
\end{definition}
The first line filters features within rank $k$; the second filters adjacent-rank features after boundary projection, which needs only one-sided filters since $\mL_k^{\mathrm{up}}\mB_k^\top=0$ and $\mL_k^{\mathrm{down}}\mB_{k+1}=0$. On a fixed complex, any spectral filter $\phi(\mL_k^{\mathrm{down}},\mL_k^{\mathrm{up}})$ equals $c\mI+P(\mL_k^{\mathrm{down}})+Q(\mL_k^{\mathrm{up}})$ for polynomials $P,Q$ without constant term (interpolate on the joint spectrum), so the polynomial form loses no generality.

The main-text layer in \cref{eq:simplicial-layer} is the case $\mathbf V=0$, $T_d=T_u=1$, constant width $d$, and
\[
\mW_{k,0}=\mI_d+\tau\mW_0^{(m)},\qquad
\mW_{k,d,1}=\tau\mW_1^{(m)},\qquad \mW_{k,u,1}=\tau\mW_2^{(m)}.
\]
Since $\tau>0$, this reparameterization is invertible, so every single-rank, degree-one instance with square channel matrices is of the form \cref{eq:simplicial-layer}. 
The correspondence requires neither symmetric nor depth-shared weights.

\subsection{SCoNe, SCNN, and SCCNN}
\paragraph{SCoNe}
The edge hidden layer of \citet[Eq.~(7)]{roddenberry2021scone} is the case $k=1$, $T_d=T_u=1$, $\mathbf V=0$, and hence, precisely an instance of \cref{eq:simplicial-layer}. 
Their weights $\mW_1,\mW_0,\mW_2$ correspond to our identity, lower, and upper coefficients, respectively.
Their final node readout is separate; with identity activation it loses triangle dependence, as their Eq.~(9) shows.

\paragraph{SCNN}
The filter bank of \citet[Eqs.~(1)--(4)]{yang2022scnn} is the case $\mathbf V=0$, with each channel matrix collecting the corresponding scalar filter coefficients across input--output channel pairs. At rank $1$, degree one recovers SCoNe's hidden-layer form.

\paragraph{SCCNN}
The multichannel layer of \citet[Eq.~(B.1)]{yang2023sccnn} is the full form \eqref{eq:polynomial-layer}; its two intra-rank zeroth-order weights sum to $\mW_{k,0}$, and setting $\mathbf V=0$ recovers SCNN. 
The same architecture is presented in \citet[Eqs.~(4)--(5)]{yang2025hodgeaware}.

\subsection{Restriction to the analyzed flow}
Taking identity activation and symmetric channel weights shared across depth in \eqref{eq:simplicial-layer} gives
\begin{equation}
\label{eq:linear-layer}
\mF_{m+1}=\mF_m+\tau\bigl(\mF_m\mW_0
+\mL_k^{\mathrm{down}}\mF_m\mW_1
+\mL_k^{\mathrm{up}}\mF_m\mW_2\bigr),
\end{equation}
the forward Euler step of \cref{eq:flow}. These restrictions are additional to the source architectures.
Continuous growth is governed by generator eigenvalues; discrete growth
by the magnitudes of the Euler multipliers. The condition in
\cref{prop:flip} ensures the same dominant eigenspace.

\paragraph{The broader polynomial family}
Higher-degree filters lead to the fixed-rank linear flow
\begin{equation}
\label{eq:generalflow}
\dot\mF=\sum_{i=0}^{p}\mA_i\mF\mW_i,
\qquad
\mA_i=\alpha_i\mI+P_i(\mL_k^{\mathrm{down}})+Q_i(\mL_k^{\mathrm{up}}),
\qquad \mW_i=\mW_i^\top,
\end{equation}
with real scalars $\alpha_i$, polynomials $P_i,Q_i$ having zero constant
term, and channel weights fixed across depth. Equation~\eqref{eq:flow}
is the case $p=2$ with $\mA_0=\mI$, $\mA_1=\mL_k^{\mathrm{down}}$,
and $\mA_2=\mL_k^{\mathrm{up}}$. More generally, each $\mA_i$ is diagonal
in the joint eigenbasis: on frequency $(\nu_d,\nu_u)$ its multiplier is
$\alpha_i+P_i(\nu_d)+Q_i(\nu_u)$. The decoupling into one $d$-dimensional system per joint eigenmode uses only this diagonal structure, so it holds for any linear layer of this form, even with non-symmetric or depth-dependent weights. The gradient-flow property (\cref{thm:gradflow}) and the normalized limit (\cref{prop:closedform}) additionally use the symmetry of the weights, and carry over verbatim. The endpoint reduction of \cref{lem:endpoint} requires both symmetry and degree one, and fails without either: the spectral abscissa of non-symmetric weights need not be convex in the frequency, and higher-degree responses can peak at an interior frequency (e.g.\ $\dot\mF=2a\,\mL_k^{\mathrm{down}}\mF-(\mL_k^{\mathrm{down}})^2\mF$ selects the gradient frequency closest to $a$). %

\section{Properties of the Rayleigh-quotient pair}
\label{app:rqlemmas}

\noindent Throughout this appendix $\mcL_1^{\mathrm{down}}$, $\mcL_1^{\mathrm{up}}$ and $\mcL_1$ are the rank-$1$ operators of \cref{sec:theory}, and \cref{ass:nondeg} is in force. Write
\[
\Sigma_g:=\operatorname{Spec}(\mcL_1^{\mathrm{down}})\setminus\{0\},
\qquad
\Sigma_c:=\operatorname{Spec}(\mcL_1^{\mathrm{up}})\setminus\{0\}
\]
for the sets of gradient and curl frequencies, and fix the common orthonormal eigenbasis $\{\vu_j\}_{j=1}^{N_1}$ of $\mcL_1^{\mathrm{down}}$ and $\mcL_1^{\mathrm{up}}$ with frequency pairs $(\nud{j},\nuu{j})$. For a nonzero feature matrix $\mF$, let $\vc_j:=\mF^\top\vu_j\in\mathbb R^d$ be its channel coefficients along $\vu_j$, so that $\mF=\sum_j\vu_j\vc_j^\top$, and define the \emph{mode masses}
\begin{equation}
\label{eq:modemass}
m_j(\mF):=\frac{\|\vc_j\|_2^2}{\|\mF\|_F^2}\ \ge 0,
\qquad
\sum_{j=1}^{N_1}m_j(\mF)=1 ,
\end{equation}
where the normalization follows from orthonormality of the $\vu_j$. Since $\mcL_1^{\mathrm{down}}\mF=\sum_j\nud{j}\,\vu_j\vc_j^\top$ and $\mcL_1^{\mathrm{up}}\mF=\sum_j\nuu{j}\,\vu_j\vc_j^\top$, the same expansion gives
\begin{align}
\label{eq:rq-masses}
\mathrm{RQ}_{\mathrm{down}}(\mF)=\sum_{j}\nud{j}\,m_j(\mF), 
\mathrm{RQ}_{\mathrm{up}}(\mF)=\sum_{j}\nuu{j}\,m_j(\mF), 
\mathrm{RQ}(\mF)=\sum_{j}\bigl(\nud{j}+\nuu{j}\bigr)\,m_j(\mF),
\end{align}
so that each Rayleigh quotient is the mass-weighted average of the corresponding frequencies. Both lemmas below are direct consequences of \eqref{eq:rq-masses}.

\subsection{Eigenspace characterizations of the three regimes}
\label{app:eigenspace}
\noindent Each regime of \cref{prop:regimeprops} is defined through the limit of a scalar Rayleigh quotient. The following lemma shows that each such limit is equivalent to concentration of the normalized feature mass on a single eigenspace, and quantifies the equivalence. Let $\mP_{\mathrm{harm}}$ be the orthogonal projector onto $\mathcal H_1$, and let $\mP_g$ and $\mP_c$ be the orthogonal projectors onto the top gradient and top curl eigenspaces $\ker(\mcL_1^{\mathrm{down}}-\nu_g^{\max}\mI)$ and $\ker(\mcL_1^{\mathrm{up}}-\nu_c^{\max}\mI)$. Define the spectral gaps
$$
\gamma_H:=\min\bigl(\Sigma_g\cup\Sigma_c\bigr),$$
$$
\gamma_g:=\nu_g^{\max}-\max\bigl(\{\nu\in\Sigma_g:\nu<\nu_g^{\max}\}\cup\{0\}\bigr),$$
$$
\gamma_c:=\nu_c^{\max}-\max\bigl(\{\nu\in\Sigma_c:\nu<\nu_c^{\max}\}\cup\{0\}\bigr),$$
all of which are strictly positive.

\begin{lemma}[Eigenspace characterizations of HD, GD and CD]
\label{lem:hd}
Let $\beta_1>0$, so that $\lambda_{\min}(\mcL_1)=0$ with eigenspace $\mathcal H_1$. For every $\mF\neq0$,
\begin{align}
\gamma_H\,\frac{\|(\mI-\mP_{\mathrm{harm}})\mF\|_F^2}{\|\mF\|_F^2}
&\;\le\;\mathrm{RQ}(\mF)\;\le\;
\lambda_{\max}(\mcL_1)\,\frac{\|(\mI-\mP_{\mathrm{harm}})\mF\|_F^2}{\|\mF\|_F^2},
\label{eq:hd-sandwich}\\
\gamma_g\,\frac{\|(\mI-\mP_g)\mF\|_F^2}{\|\mF\|_F^2}
&\;\le\;\nu_g^{\max}-\mathrm{RQ}_{\mathrm{down}}(\mF)\;\le\;
\nu_g^{\max}\,\frac{\|(\mI-\mP_g)\mF\|_F^2}{\|\mF\|_F^2},
\label{eq:gd-sandwich}
\end{align}
and the analogue of \eqref{eq:gd-sandwich} holds for $\nu_c^{\max}-\mathrm{RQ}_{\mathrm{up}}(\mF)$ with $\gamma_c$ and $\mP_c$. In particular $\mathrm{RQ}(\mF)\ge0$, with equality if and only if every column of $\mF$ lies in $\mathcal H_1$, and a dynamics $\mF(t)\neq0$ is HD, GD or CD if and only if, respectively,
\[
\frac{\|(\mI-\mP_{\mathrm{harm}})\mF(t)\|_F}{\|\mF(t)\|_F}\longrightarrow0,
\qquad
\frac{\|(\mI-\mP_g)\mF(t)\|_F}{\|\mF(t)\|_F}\longrightarrow0,
\qquad
\frac{\|(\mI-\mP_c)\mF(t)\|_F}{\|\mF(t)\|_F}\longrightarrow0 .
\]
\end{lemma}
\begin{proof}
By \eqref{eq:rq-masses}, $\mathrm{RQ}(\mF)=\sum_j\lambda_j m_j$ with $\lambda_j:=\nud{j}+\nuu{j}$ the eigenvalue of $\mcL_1$ on $\vu_j$. Since $(\mI-\mP_{\mathrm{harm}})\mF=\sum_{j:\lambda_j>0}\vu_j\vc_j^\top$, orthonormality gives $\|(\mI-\mP_{\mathrm{harm}})\mF\|_F^2/\|\mF\|_F^2=\sum_{j:\lambda_j>0}m_j$, and every $\lambda_j>0$ lies in $[\gamma_H,\lambda_{\max}(\mcL_1)]$ from \eqref{eq:hd-sandwich}. For \eqref{eq:gd-sandwich}, write $\nu_g^{\max}-\mathrm{RQ}_{\mathrm{down}}(\mF)=\sum_j(\nu_g^{\max}-\nud{j})\,m_j$. The coefficient vanishes when $\vu_j$ lies in the top gradient eigenspace, and for every other $j$ it lies in $[\gamma_g,\nu_g^{\max}]$, because $\nud{j}$ is then either $0$ (harmonic or curl mode) or a gradient frequency below $\nu_g^{\max}$; summing as before gives \eqref{eq:gd-sandwich}. The curl case is identical. The equality case and the three equivalences follow by letting the sandwiched quantities tend to zero.
\end{proof}

\begin{remark}[Bottom eigenspace for general $\beta_1$]
\label{rem:bottom}
The proof of \eqref{eq:hd-sandwich} uses only that $\mathcal H_1$ is the eigenspace of the smallest eigenvalue of $\mcL_1$. For arbitrary $\beta_1$, let $E_{\min}:=\ker(\mcL_1-\lambda_{\min}(\mcL_1)\mI)$ with orthogonal projector $\mP_{\min}$, and let $\gamma_{\min}:=\min\{\lambda\in\operatorname{Spec}(\mcL_1):\lambda>\lambda_{\min}(\mcL_1)\}-\lambda_{\min}(\mcL_1)$. The same argument gives
\[
\gamma_{\min}\,\frac{\|(\mI-\mP_{\min})\mF\|_F^2}{\|\mF\|_F^2}
\le
\mathrm{RQ}(\mF)-\lambda_{\min}(\mcL_1)
\le
\bigl(\lambda_{\max}(\mcL_1)-\lambda_{\min}(\mcL_1)\bigr)\frac{\|(\mI-\mP_{\min})\mF\|_F^2}{\|\mF\|_F^2},
\]
so $\mathrm{RQ}(\mF(t))\to\lambda_{\min}(\mcL_1)$ if and only if the normalized mass concentrates on $E_{\min}$. This is the form used for $\beta_1=0$ in \cref{app:nobetti}.
\end{remark}

\subsection{Feasible region and mutual exclusion}
\label{app:polygon}
\begin{lemma}[Feasible polygon; mutual exclusion of the regimes]
\label{lem:gdcd}
For every $\mF\neq0$,
\[
\bigl(\mathrm{RQ}_{\mathrm{down}}(\mF),\mathrm{RQ}_{\mathrm{up}}(\mF)\bigr)
=\sum_{j=1}^{N_1}m_j(\mF)\,\bigl(\nud{j},\nuu{j}\bigr)
\;\in\;
\mathcal P
\]
with
$$\mathcal P:=\operatorname{conv}\Bigl(\{(\nu,0):\nu\in\Sigma_g\}\cup\{(0,\nu):\nu\in\Sigma_c\}\cup\{(0,0)\ \text{if }\beta_1>0\}\Bigr).
$$
If $\beta_1>0$, $\mathcal P$ is the triangle with vertices $(0,0)$, $(\nu_g^{\max},0)$, $(0,\nu_c^{\max})$; if $\beta_1=0$, it is the convex hull of $(\nu_g^{\min},0)$, $(\nu_g^{\max},0)$, $(0,\nu_c^{\min})$, $(0,\nu_c^{\max})$. In either case
\begin{equation}
\label{eq:polygon-ineq}
\frac{\mathrm{RQ}_{\mathrm{down}}(\mF)}{\nu_g^{\max}}+\frac{\mathrm{RQ}_{\mathrm{up}}(\mF)}{\nu_c^{\max}}\le 1 ,
\end{equation}
equivalently
\[
\mathrm{RQ}_{\mathrm{up}}(\mF)
\le
\frac{\nu_c^{\max}}{\nu_g^{\max}}
\bigl(\nu_g^{\max}-\mathrm{RQ}_{\mathrm{down}}(\mF)\bigr)
\qquad\text{and}\qquad
\mathrm{RQ}_{\mathrm{down}}(\mF)
\le
\frac{\nu_g^{\max}}{\nu_c^{\max}}
\bigl(\nu_c^{\max}-\mathrm{RQ}_{\mathrm{up}}(\mF)\bigr).
\]
Consequently, GD implies $(\mathrm{RQ}_{\mathrm{down}},\mathrm{RQ}_{\mathrm{up}})\to(\nu_g^{\max},0)$, CD implies $(\mathrm{RQ}_{\mathrm{down}},\mathrm{RQ}_{\mathrm{up}})\to(0,\nu_c^{\max})$, and the regimes HD, GD and CD are pairwise mutually exclusive.
\end{lemma}
\begin{proof}
The identity \eqref{eq:rq-masses} exhibits the pair as a convex combination of the points $(\nud{j},\nuu{j})$, each of which is $(\nu,0)$ with $\nu\in\Sigma_g$, $(0,\nu)$ with $\nu\in\Sigma_c$, or $(0,0)$, the last only if $\beta_1>0$, so are elements of $\mathcal P$. For every $j$ at most one of $\nud{j},\nuu{j}$ is nonzero, so $\nud{j}/\nu_g^{\max}+\nuu{j}/\nu_c^{\max}\le1$; averaging against the masses gives \eqref{eq:polygon-ineq}, and multiplying by $\nu_c^{\max}$ or by $\nu_g^{\max}$ gives the two equivalent forms. If the dynamics is GD then $\mathrm{RQ}_{\mathrm{down}}\to\nu_g^{\max}$, hence $\mathrm{RQ}_{\mathrm{up}}\to0$ by the first form; the CD case is symmetric, and the two limit points are distinct, so GD and CD exclude each other. Finally, GD forces $\mathrm{RQ}\to\nu_g^{\max}>0$ and CD forces $\mathrm{RQ}\to\nu_c^{\max}>0$, whereas HD requires $\mathrm{RQ}\to0$.
\end{proof}

\subsection{Why HD is over-smoothing, and GD and CD are not}
\label{app:whyhd}
\paragraphnoper{Why HD recovers over-smoothing on graphs}
By \cref{lem:hd}, HD means that the normalized feature mass concentrates on the harmonic space $\mathcal H_1$, the eigenspace of the smallest eigenvalue of $\mcL_1$: asymptotically the normalized divergence $\hat\mB_1\mF(t)/\|\mF(t)\|_F$ and curl $\hat\mB_2^\top\mF(t)/\|\mF(t)\|_F$ both vanish, and every spectral component above the lowest Hodge frequency is suppressed relative to $\|\mF(t)\|_F$. This is the direct analogue of low-frequency dominance on graphs (\cref{app:graphbg}), where the lowest eigenspace of the symmetric normalized Laplacian is spanned by the degree-scaled constant $\mD^{1/2}\vone$ and the limiting node features therefore carry no information beyond node degree. The difference lies in what the limit looks like: a harmonic edge signal is not determined by local degrees and may retain global circulation determined by the topology of the complex, so HD erases local divergence and curl structure while preserving the $\beta_1$ topological modes. When $\beta_1=0$ there is no harmonic space and HD cannot occur; \cref{app:nobetti} shows that the flow then concentrates on one of the four endpoint frequencies instead.

\paragraphnoper{Why GD and CD split the notion of over-sharpening} GD and CD mean that the features remain dominated by the most oscillatory mode of the gradient or of the curl family: divergence, respectively curl, persists relative to $\|\mF(t)\|_F$ at maximal frequency. There is an exact correspondence with the graph case: $\mcL_1^{\mathrm{down}}=\hat\mB_1^\top\hat\mB_1$ and $\mcL_0^{\mathrm{up}}=\hat\mB_1\hat\mB_1^\top$ share their nonzero spectra, so gradient dominance means that the normalized features align with the coboundary lift of the top graph frequency, that is, over-sharpening in the sense of \citet{digiovanni2023understandingconvolutiongraphsenergies} seen from edges rather than nodes. Curl dominance is its face-level counterpart and has no rank-$0$ analogue. \Cref{thm:trichotomy} shows that for the flow \eqref{eq:flow} only the three endpoint limits occur generically.

\section{Energy, attraction and repulsion}
\label{app:energy}

\noindent We now use the physical particle attraction and repulsion interpretation of the gradient flow of the parametric energy to explain why both smoothing and sharpening are possible.
Recall
\[
\mcL_1^{\mathrm{down}}=\hat \mB_1^\top\hat \mB_1,
\qquad
\mcL_1^{\mathrm{up}}=\hat \mB_2\hat \mB_2^\top,
\]
with $\hat \mB_1$ and $\hat \mB_2$ the normalized boundary operators. We interpret
each edge-feature vector $\vf_e\in\mathbb R^d$ as a particle in feature space.
Unlike on a graph, however, these particles interact through two different
higher-order structures: node stars and $2$-simplices.

\noindent Starting from the energy
\[
\mathcal E(\mF)
=
\underbrace{
-\frac12
\mathrm{Tr}\!\big(
\mF^\top \mF\,\mW_0
\big)
}_{\text{external term}}
\;
\underbrace{
-\frac{1}{2}
\mathrm{Tr}\!\big(
\mF^\top\mcL_1^{\mathrm{down}}\mF \mW_1
\big)
}_{\text{node term}}
\;
\underbrace{
-\frac{1}{2}
\mathrm{Tr}\!\big(
\mF^\top\mcL_1^{\mathrm{up}}\mF \mW_2
\big)
}_{\text{face term}} ,
\]
\noindent the external term is independent of the simplicial incidence structure. Defining
\[
\mW_{\mathrm{ext}}
:=
\mW_0,
\]
we have
\[
\mathrm{Tr}(\mF^\top \mF \mW_{\mathrm{ext}})
=
\sum_e
\langle \vf_e,\mW_{\mathrm{ext}}\vf_e\rangle,
\]
and therefore
\[
\mathcal E_{\mathrm{ext}}(\mF)
=
-\frac12
\sum_e
\langle \vf_e,\mW_{\mathrm{ext}}\vf_e\rangle.
\]
This is the analogue of the graph-independent external energy in~\citet{digiovanni2023understandingconvolutiongraphsenergies}.

\noindent We begin by looking at the node term, corresponding to divergence. Insert
$\mcL_1^{\mathrm{down}}=\hat \mB_1^\top\hat \mB_1$ and regroup:
\[
\mathrm{Tr}\!\big(
\mF^\top\hat \mB_1^\top\hat \mB_1\mF \mW_1
\big)
=
\mathrm{Tr}\!\big(
(\hat \mB_1\mF)^\top(\hat \mB_1\mF)\mW_1
\big).
\]
Define the normalized divergence field
\[
\mathrm{div}\,\mF
:=
\hat \mB_1\mF
\in\mathbb R^{N_0\times d},
\]
whose row at node $v$ is
\[
\vg_v
:=
(\hat \mB_1\mF)_{v,:}
=
\sum_e(\hat \mB_1)_{ve}\vf_e
\in\mathbb R^d.
\]
Only edges incident to $v$ contribute to this sum. Thus $\vg_v$ is the signed, normalization-weighted aggregate of the edge features in the star of $v$, or equivalently the normalized net in/outflow at that node, channel by channel. Applying the same trace identity gives
\[
\mathrm{Tr}\!\big(
\mF^\top\mcL_1^{\mathrm{down}}\mF \mW_1
\big)
=
\sum_v
\langle \vg_v,\mW_1\vg_v\rangle.
\]
Hence the node contribution is
\[
\mathcal E_{\mathrm{down}}(\mF)
=
-\frac{1}{2}
\sum_v
\langle \vg_v,\mW_1\vg_v\rangle.
\]

We now look at the face term, corresponding to curl. Similarly, using
$\mcL_1^{\mathrm{up}}=\hat \mB_2\hat \mB_2^\top$,
\[
\mathrm{Tr}\!\big(
\mF^\top\hat \mB_2\hat \mB_2^\top \mF \mW_2
\big)
=
\mathrm{Tr}\!\big(
(\hat \mB_2^\top \mF)^\top
(\hat \mB_2^\top \mF)
\mW_2
\big).
\]
Define the normalized curl field
\[
\mathrm{curl}\,\mF
:=
\hat \mB_2^\top \mF
\in\mathbb R^{N_2\times d},
\]
whose row at a $2$-simplex $\sigma\in X_2$ is
\[
\vh_\sigma
:=
(\hat \mB_2^\top \mF)_{\sigma,:}
=
\sum_e(\hat \mB_2)_{e\sigma}\vf_e
\in\mathbb R^d.
\]
The nonzero coefficients occur on the three oriented edges of $\sigma$, so
$\vh_\sigma$ is the signed, normalization-weighted circulation of the feature
signal around that face, channel by channel. Therefore
\[
\mathrm{Tr}\!\big(
\mF^\top\mcL_1^{\mathrm{up}}\mF \mW_2
\big)
=
\sum_{\sigma\in X_2}
\langle \vh_\sigma,\mW_2\vh_\sigma\rangle,
\]
and the face contribution is
\[
\mathcal E_{\mathrm{up}}(\mF)
=
-\frac{1}{2}
\sum_{\sigma\in X_2}
\langle \vh_\sigma,\mW_2\vh_\sigma\rangle.
\]

\paragraph{Decomposition by Hodge sector}
Write $\mF=\mF_H+\mF_G+\mF_C$ for the column-wise orthogonal projections of $\mF$ onto the harmonic, gradient and curl subspaces. Because the three subspaces are mutually orthogonal, $\mF_s^\top\mF_{s'}=0$ for $s\neq s'$, so the external term splits as $\mathcal E_{\mathrm{ext}}(\mF)=\mathcal E_{\mathrm{ext}}(\mF_H)+\mathcal E_{\mathrm{ext}}(\mF_G)+\mathcal E_{\mathrm{ext}}(\mF_C)$. Moreover $\hat\mB_1\mF=\hat\mB_1\mF_G$, since $\hat\mB_1$ vanishes on $\mathcal H_1$ and on $\operatorname{im}\hat\mB_2$, and likewise $\hat\mB_2^\top\mF=\hat\mB_2^\top\mF_C$. Hence
\[
\mathcal E(\mF)
=
\mathcal E_{\mathrm{ext}}(\mF_H)
+\bigl[\mathcal E_{\mathrm{ext}}(\mF_G)+\mathcal E_{\mathrm{down}}(\mF_G)\bigr]
+\bigl[\mathcal E_{\mathrm{ext}}(\mF_C)+\mathcal E_{\mathrm{up}}(\mF_C)\bigr]:
\]
the energy is a sum of three energies, one per Hodge sector; the node term sees only the gradient part and the face term sees only the curl part. This is the energetic counterpart of the decoupling of the flow into independent joint eigenmodes in \cref{sec:trichotomy}.

\paragraph{Attraction and repulsion}
We now split each channel-mixing matrix according to the signs of its
eigenvalues. As in ~\citet[Sec.~4.2.2]{digiovanni2023understandingconvolutiongraphsenergies},
for $s\in\{1,2\}$, symmetry allows us to write
\[
\mW_s
=
\mTheta_{s,+}^\top\mTheta_{s,+}
-
\mTheta_{s,-}^\top\mTheta_{s,-},
\]
so that, for any $\vz\in\mathbb R^d$,
\[
\langle \vz,\mW_s\vz\rangle
=
\|\mTheta_{s,+}\vz\|_2^2
-
\|\mTheta_{s,-}\vz\|_2^2.
\]
Substituting these decompositions into the energy gives
\[
\begin{aligned}
\mathcal E(\mF)
={}&
-\frac12
\sum_e
\langle \vf_e,\mW_{\mathrm{ext}}\vf_e\rangle
\\
&-
\frac{1}{2}
\sum_v
\left(
\|\mTheta_{1,+}\vg_v\|_2^2
-
\|\mTheta_{1,-}\vg_v\|_2^2
\right)
\\
&-
\frac{1}{2}
\sum_{\sigma\in X_2}
\left(
\|\mTheta_{2,+}\vh_\sigma\|_2^2
-
\|\mTheta_{2,-}\vh_\sigma\|_2^2
\right).
\end{aligned}
\]

\noindent The flow follows the direction of steepest descent of the total
energy. Whether a particular divergence or circulation component is attractive
or repulsive therefore depends on the sign of the corresponding channel eigenvalue. More precisely, let $\mu$ be an eigenvalue of
$\mW_1$. Along that channel direction, the node contribution has
coefficient
\[
-\frac12 \mu
\]
multiplying the squared divergence. Hence
\[
\mu<0
\quad\Longrightarrow\quad
\text{positive divergence penalty},
\]
whose gradient contribution suppresses that divergence component and therefore
induces a down-smoothing, or attractive, tendency. Conversely,
\[
\mu>0
\quad\Longrightarrow\quad
\text{negative divergence contribution},
\]
whose gradient contribution favors increasing that component and therefore
induces a down-sharpening, or repulsive, tendency.

\noindent The same interpretation holds for the upper interaction. If $\mu$
is an eigenvalue of $\mW_2$, then the corresponding squared
circulation appears with coefficient
\[
-\frac12 \mu.
\]
Thus $\mu<0$ gives an attractive, up-smoothing tendency, while
$\mu>0$ gives a repulsive, up-sharpening tendency. The lower and upper
families have independent channel-mixing matrices and can therefore induce
different behaviors simultaneously.

\noindent Here the external term acts independently on
each edge feature and does not measure either divergence or circulation. In
particular, on the harmonic subspace,
\[
\hat \mB_1\mF=0,
\qquad
\hat \mB_2^\top \mF=0,
\]
so both structural terms vanish. The harmonic component then evolves only
through the external channel mixing,
\[
\mP_{\mathrm{harm}}\mF(t)
=
\mP_{\mathrm{harm}}\mF(0)
\exp\!\left(
t\mW_0
\right),
\]
where $\mP_{\mathrm{harm}}$ is the orthogonal projector onto the harmonic subspace; \cref{app:AvsL} discusses the consequences for the two tyings of \cref{sec:instantiation}.

\noindent These interpretations concern the contribution of each signed
quadratic term to the gradient. They do not imply that the individual terms
are themselves monotone along the full coupled flow; only the total energy
$\mathcal E(\mF(t))$ is guaranteed to decrease. Likewise, an attractive or
repulsive tendency in one interaction family does not by itself determine the
asymptotic regime. The eventual HD, GD, or CD behavior depends on the comparison between the full spectral growth rates derived in \cref{sec:trichotomy}.

\paragraph{Differences from the graph case}
There are two important differences from the graph case in~\citet{digiovanni2023understandingconvolutiongraphsenergies}.
First, their structural forces are pairwise, acting between node-feature
particles across graph edges. Here the particles are edge features and the
interactions are multi-way. The node term couples all edges incident to a
common node through $\vg_v$, while the face term couples the three boundary
edges of a $2$-simplex through $\vh_\sigma$. Second, the lower and upper
interaction families have independent channel-mixing matrices, so attraction and repulsion can occur differently in the gradient and curl sectors.

\section{Proofs of the results in \texorpdfstring{Section~\ref{sec:theory}}{Section 3}}
\label{app:proofs}

\noindent We use the notation of \cref{app:rqlemmas}: the joint eigenbasis $\{\vu_j\}_{j=1}^{N_1}$ of $\mcL_1^{\mathrm{down}}$ and $\mcL_1^{\mathrm{up}}$, the channel coefficients $\vc_j=\mF^\top\vu_j$, and the mode masses \eqref{eq:modemass}. The vectorized generator $\gen$ introduced in the first proof below is used throughout the remaining appendices.

\subsection{Proof of \texorpdfstring{Theorem~\ref{thm:gradflow}}{Theorem \ref{thm:gradflow}}}
\begin{proof}
Let $\vf=\operatorname{vec}(\mF)\in\mathbb R^{N_1d}$ stack the columns of $\mF$. Using the trace--Kronecker identity $\operatorname{Tr}(\mX^\top\mA\mX\mB)=\operatorname{vec}(\mX)^\top(\mB^\top\otimes\mA)\operatorname{vec}(\mX)$ and the symmetry of $\mW_0,\mW_1,\mW_2$, the energy is the quadratic form
\begin{equation}
\label{eq:vecgen}
\mathcal E(\mF)=-\tfrac12\,\vf^\top\gen\,\vf,
\qquad
\gen:=\mW_0\otimes\mI+\mW_1\otimes\mcL_1^{\mathrm{down}}+\mW_2\otimes\mcL_1^{\mathrm{up}} .
\end{equation}
Each factor is symmetric and the Kronecker product of symmetric matrices is symmetric, so $\gen=\gen^\top$ and $\nabla_{\vf}\mathcal E=-\gen\vf$. The negative gradient flow $\dot\vf=-\nabla_{\vf}\mathcal E=\gen\vf$ is, by $(\mB^\top\otimes\mA)\operatorname{vec}(\mX)=\operatorname{vec}(\mA\mX\mB)$ applied to each term, exactly \eqref{eq:flow} in matrix form. Finally $\tfrac{d}{dt}\mathcal E(\mF(t))=\langle\nabla_{\vf}\mathcal E,\dot\vf\rangle=-\|\gen\vf\|_2^2\le0$.
\end{proof}
\noindent The proof uses only that each factor is symmetric, so the gradient-flow property holds for every affine (indeed every polynomial) propagation operator built from symmetric Laplacians, normalized or not. We call $\gen$ the \emph{vectorized generator} of the flow.

\subsection{Proof of \texorpdfstring{Proposition~\ref{prop:closedform}}{Proposition \ref{prop:closedform}}}
\label{app:normlimit}
\begin{proof}
For the Frobenius inner product,
$\langle\vu_i\vpsi^\top,\vu_j\vphi^\top\rangle
=\operatorname{Tr}(\vpsi\vu_i^\top\vu_j\vphi^\top)
=(\vu_i^\top\vu_j)(\vpsi^\top\vphi)$, which is $1$ if $(i,\vpsi)=(j,\vphi)$
and $0$ for distinct elements of $\{\vu_j\vpsi_{j,r}^\top\}$; hence these
$N_1d$ matrices are orthonormal, and the norm identity
$\|\mF(t)\|^2=\sum_{j,r}e^{2\mu_{j,r}t}a_{j,r}^2$ is the sum of
squared coefficients in \eqref{eq:closedform}. Factoring $e^{\mu_*t}$ out of
\eqref{eq:closedform} and out of the norm identity, every term with
$\mu_{j,r}<\mu_*$ carries a factor $e^{(\mu_{j,r}-\mu_*)t}\to0$, which gives
the first limit in \eqref{eq:normlimit}. For the second, left-multiplication
by $\mcL_1^{\mathrm{down}}$ (resp.\ $\mcL_1^{\mathrm{up}}$) maps
$\vu_j\vpsi_{j,r}^\top$ to $\nud{j}\,\vu_j\vpsi_{j,r}^\top$ (resp.\
$\nuu{j}\,\vu_j\vpsi_{j,r}^\top$) by shared basis, and the terms
are orthogonal, so
$\operatorname{Tr}(\mF^\top\mcL_1^{\mathrm{down}}\mF)
=\sum_{j,r}e^{2\mu_{j,r}t}a_{j,r}^2\,\nud{j}$ and likewise for
$\mcL_1^{\mathrm{up}}$; dividing by the norm identity and letting
$t\to\infty$ leaves the stated weighted averages over $T_*$. The excluded set $\{\sum_{(j,r)\in T_*}a_{j,r}^2=0\}$ is the orthogonal complement $E_*^\perp$ of the top eigenspace, a proper subspace since $T_*\neq\varnothing$. That the $\vu_j\vpsi_{j,r}^\top$ are eigenvectors of the map $\mF\mapsto\mF\mW_0+\mcL_1^{\mathrm{down}}\mF\mW_1+\mcL_1^{\mathrm{up}}\mF\mW_2$ with eigenvalues $\mu_{j,r}$ is the identity $\gen(\vpsi_{j,r}\otimes\vu_j)=\mu_{j,r}(\vpsi_{j,r}\otimes\vu_j)$ established next.
\end{proof}

\paragraph{Spectrum of the vectorized generator}
Since $\operatorname{vec}(\vu_j\vpsi^\top)=\vpsi\otimes\vu_j$ and
$(\mW\otimes\mA)(\vpsi\otimes\vu)=(\mW\vpsi)\otimes(\mA\vu)$,
shared orthonormality gives $\gen(\vpsi\otimes\vu_j)=(\mG_j\vpsi)\otimes\vu_j$. The subspaces $\mathbb R^d\otimes\operatorname{span}\{\vu_j\}$ are therefore invariant under $\gen$, mutually orthogonal, and span $\mathbb R^{N_1d}$, and $\gen$ is block diagonal on them with the symmetric blocks $\mG_j$. Its eigenpairs are exactly $(\mu_{j,r},\vpsi_{j,r}\otimes\vu_j)$, so
\begin{align}
\label{eq:specgen}
\operatorname{Spec}(\gen)=\bigcup_{j=1}^{N_1}\operatorname{Spec}(\mG_j),\\
\lambda_{\max}(\gen)=\max_j\lambda_{\max}(\mG_j)=\mu_*,
\lambda_{\min}(\gen)=\min_j\lambda_{\min}(\mG_j)=\mu_{\min}.
\end{align}
Because $\mG_j$ depends on $\vu_j$ only through its frequency pair, the index set $T_*$ is a union of full joint eigenspaces and does not depend on the chosen basis, and \cref{prop:closedform} says that $\vf(t)/\|\vf(t)\|_2$ converges to the normalized orthogonal projection of $\vf(0)$ onto the top eigenspace $\ker(\gen-\mu_*\mI)$. The extreme rates $\mu_*$ and $\mu_{\min}$ of \cref{sec:trichotomy} are thus the extreme eigenvalues of $\gen$, which is the form used in \cref{app:discrete}.

\subsection{Endpoint reduction}
\label{app:endpoint}
For a real parameter $\nu$, write
\begin{equation}
\label{eq:generators}
\mG^{\mathrm{harm}}:=\mW_0,
\qquad
\mG^{\mathrm{grad}}_{\nu}:=\mW_0+\nu\mW_1,
\qquad
\mG^{\mathrm{curl}}_{\nu}:=\mW_0+\nu\mW_2 ,
\end{equation}
so that the channel generator \eqref{eq:mode-ode} of a harmonic mode, a gradient mode of frequency $\nu$, or a curl mode of frequency $\nu$ is $\mG^{\mathrm{harm}}$, $\mG^{\mathrm{grad}}_\nu$ or $\mG^{\mathrm{curl}}_\nu$ respectively, and the endpoint rates \eqref{eq:rates} are $\rho_H=\lambda_{\max}(\mG^{\mathrm{harm}})$, $\rho_G=\lambda_{\max}(\mG^{\mathrm{grad}}_{\nu_g^{\max}})$ and $\rho_C=\lambda_{\max}(\mG^{\mathrm{curl}}_{\nu_c^{\max}})$. The affine families \eqref{eq:generators} are defined for every real $\nu$, not only for the frequencies that occur in the complex. The following lemma is the convexity step used in \cref{sec:trichotomy}.

\begin{lemma}[Endpoint reduction]
\label{lem:endpoint}
The map $\nu\mapsto\lambda_{\max}(\mG^{\mathrm{grad}}_\nu)$ is convex on
$\mathbb R$, and hence for $0\le\nu\le\nu_g^{\max}$,
\begin{equation}
\label{eq:endpoint-bound}
\lambda_{\max}\!\left(\mG^{\mathrm{grad}}_\nu\right)
\le
\Big(1-\frac{\nu}{\nu_g^{\max}}\Big)\rho_H
+\frac{\nu}{\nu_g^{\max}}\,\rho_G .
\end{equation}
The same holds for the curl family with $\rho_C$ in place of $\rho_G$. Symmetrically, $\nu\mapsto\lambda_{\min}(\mG^{\mathrm{grad}}_\nu)$ is concave, and for $0\le\nu\le\nu_g^{\max}$,
\begin{equation}
\label{eq:endpoint-bound-min}
\lambda_{\min}\!\left(\mG^{\mathrm{grad}}_\nu\right)
\ge
\Big(1-\frac{\nu}{\nu_g^{\max}}\Big)\lambda_{\min}(\mW_0)
+\frac{\nu}{\nu_g^{\max}}\,\lambda_{\min}\!\left(\mG^{\mathrm{grad}}_{\nu_g^{\max}}\right),
\end{equation}
again with the curl analogue. Consequently, if $\beta_1>0$, the top and bottom rates of the flow are
\[
\mu_*=\max\{\rho_H,\rho_G,\rho_C\},
\qquad
\mu_{\min}=\min\bigl\{\lambda_{\min}(\mW_0),\ \lambda_{\min}(\mG^{\mathrm{grad}}_{\nu_g^{\max}}),\ \lambda_{\min}(\mG^{\mathrm{curl}}_{\nu_c^{\max}})\bigr\},
\]
the second being \eqref{eq:mumin}.
\end{lemma}
\begin{proof}
Write $r_g(\nu):=\lambda_{\max}(\mG^{\mathrm{grad}}_\nu)$ and $r_c(\nu):=\lambda_{\max}(\mG^{\mathrm{curl}}_\nu)$. We use the continuous parameter to obtain bounds and then evaluate them at the frequencies present.

\smallskip
\noindent\emph{Convexity.} For a real symmetric $\mS$, $\lambda_{\max}(\mS)=\max_{\|\vpsi\|_2=1}\vpsi^\top\mS\vpsi$. For each fixed unit $\vpsi$, the function $q_\psi(\nu):=\vpsi^\top\mG^{\mathrm{grad}}_\nu\vpsi=\vpsi^\top\mW_0\vpsi+\nu\,\vpsi^\top\mW_1\vpsi$ is affine in $\nu$, and $r_g(\nu)=\max_{\|\vpsi\|_2=1}q_\psi(\nu)$ is a pointwise maximum of affine functions, hence convex: for $\theta\in[0,1]$,
\[
r_g\bigl((1-\theta)\nu_0+\theta\nu_1\bigr)
=\max_{\|\vpsi\|_2=1}\bigl[(1-\theta)q_\psi(\nu_0)+\theta q_\psi(\nu_1)\bigr]
\le(1-\theta)\,r_g(\nu_0)+\theta\,r_g(\nu_1).
\]
The maximizing $\vpsi$ may change with $\nu$; no common eigenbasis of the channel weights is required. The same argument applies to $r_c$.

\smallskip
\noindent\emph{Endpoint bound.} At $\nu=0$ both affine families reduce to $\mG^{\mathrm{harm}}$, so $r_g(0)=r_c(0)=\rho_H$, while $r_g(\nu_g^{\max})=\rho_G$ and $r_c(\nu_c^{\max})=\rho_C$. For $0\le\nu\le\nu_g^{\max}$, convexity with $\theta=\nu/\nu_g^{\max}$ gives \eqref{eq:endpoint-bound}; since its right-hand side is a convex combination, $r_g(\nu)\le\max\{\rho_H,\rho_G\}$. Likewise $r_c(\nu)\le\max\{\rho_H,\rho_C\}$ on $[0,\nu_c^{\max}]$. These bounds do not require a harmonic mode to exist.

\smallskip
\noindent\emph{Top rate.} Assume $\beta_1>0$. By \eqref{eq:specgen}, $\mu_*=\max_j\lambda_{\max}(\mG_j)$, where each $\mG_j$ equals $\mG^{\mathrm{harm}}$, $\mG^{\mathrm{grad}}_{\nud{j}}$ or $\mG^{\mathrm{curl}}_{\nuu{j}}$ according to the type of $\vu_j$; by the endpoint bound every block is at most $\max\{\rho_H,\rho_G,\rho_C\}$. Conversely each of the three rates is attained: $\beta_1>0$ provides a harmonic $\vu_j$ with $\mG_j=\mG^{\mathrm{harm}}$, and the finite nonempty families $\Sigma_g$ and $\Sigma_c$ contain their maxima, providing modes with $\mG_j=\mG^{\mathrm{grad}}_{\nu_g^{\max}}$ and $\mG_j=\mG^{\mathrm{curl}}_{\nu_c^{\max}}$. Hence $\mu_*=\max\{\rho_H,\rho_G,\rho_C\}$.

\smallskip
\noindent\emph{Bottom rate.} The same three steps apply to $\lambda_{\min}$ with every inequality reversed: $\lambda_{\min}(\mS)=\min_{\|\vpsi\|_2=1}\vpsi^\top\mS\vpsi$ is a pointwise minimum of the affine functions $q_\psi$, hence concave, which gives \eqref{eq:endpoint-bound-min} and its curl analogue; the right-hand side is a convex combination, so on each family $\lambda_{\min}(\mG_j)$ is at least the smaller of its two endpoint values; by \eqref{eq:specgen}, $\mu_{\min}=\lambda_{\min}(\gen)=\min_j\lambda_{\min}(\mG_j)$; and the three endpoint values are attained by the same three modes as above. Hence $\mu_{\min}=\min\{\lambda_{\min}(\mW_0),\lambda_{\min}(\mG^{\mathrm{grad}}_{\nu_g^{\max}}),\lambda_{\min}(\mG^{\mathrm{curl}}_{\nu_c^{\max}})\}$.
\end{proof}

\subsection{Proof of \texorpdfstring{Theorem~\ref{thm:trichotomy}}{Theorem \ref{thm:trichotomy}}}
\begin{proof}
By \cref{prop:closedform}, for every $\mF(0)$ outside the proper subspace $\{\sum_{(j,r)\in T_*}a_{j,r}^2=0\}$, which consists of the initial conditions with zero projection onto the top-rate eigenspace and has Lebesgue measure zero, the normalized features converge and the Rayleigh-quotient pair converges to the weighted average \eqref{eq:normlimit} of the frequency pairs $(\nud{j},\nuu{j})$ over $T_*=\{(j,r):\mu_{j,r}=\mu_*\}$. It therefore suffices to show that, under the strict-maximum hypothesis, every $j$ appearing in $T_*$ has the same frequency pair, namely $(0,0)$, $(\nu_g^{\max},0)$ or $(0,\nu_c^{\max})$ in the three cases; the limit is then that pair. Recall from \cref{lem:endpoint} that $\mu_*=\max\{\rho_H,\rho_G,\rho_C\}$, and that $\mu_{j,r}\le\lambda_{\max}(\mG_j)$ with $\mG_j$ determined by the Hodge type and frequency of $\vu_j$.

\smallskip
\noindent\emph{Case (i): $\rho_H>\max\{\rho_G,\rho_C\}$.}
For a gradient mode $\vu_j$ of frequency $0<\nu\le\nu_g^{\max}$, \eqref{eq:endpoint-bound} gives
\[
\lambda_{\max}(\mG_j)\le\Bigl(1-\frac{\nu}{\nu_g^{\max}}\Bigr)\rho_H+\frac{\nu}{\nu_g^{\max}}\rho_G<\rho_H,
\]
strictly because the coefficient of $\rho_G<\rho_H$ is positive; the same holds for curl modes with $\rho_C$. Hence $\mu_{j,r}=\mu_*=\rho_H$ forces $\vu_j$ to be harmonic, every pair in $T_*$ is $(0,0)$, and $(\mathrm{RQ}_{\mathrm{down}},\mathrm{RQ}_{\mathrm{up}})\to(0,0)$: the flow is HD.

\smallskip
\noindent\emph{Case (ii): $\rho_G>\max\{\rho_H,\rho_C\}$.}
Harmonic modes have $\lambda_{\max}(\mG_j)=\rho_H<\rho_G$, and curl modes have $\lambda_{\max}(\mG_j)\le\max\{\rho_H,\rho_C\}<\rho_G$ by the endpoint bound. For a gradient mode of frequency $\nu<\nu_g^{\max}$, \eqref{eq:endpoint-bound} gives $\lambda_{\max}(\mG_j)<\rho_G$, strictly because the coefficient of $\rho_H<\rho_G$ is positive. Hence $\mu_{j,r}=\mu_*$ forces $\vu_j$ to be a gradient mode of frequency exactly $\nu_g^{\max}$, every pair in $T_*$ is $(\nu_g^{\max},0)$, and $(\mathrm{RQ}_{\mathrm{down}},\mathrm{RQ}_{\mathrm{up}})\to(\nu_g^{\max},0)$: the flow is GD.

\smallskip
\noindent\emph{Case (iii): $\rho_C>\max\{\rho_H,\rho_G\}$} is case (ii) with the roles of the gradient and curl families exchanged, giving the limit $(0,\nu_c^{\max})$ and CD.

\smallskip
\noindent Neither a simple top eigenvalue nor a positive value of $\mu_*$ is needed: only differences of growth rates enter. If two rates tied at the maximum, $T_*$ could contain modes of different types and the limit would be the corresponding weighted combination of their frequency pairs.
\end{proof}

\section{Extensions of the spectral concentration trichotomy}
\label{app:extensions}

\subsection{The case \texorpdfstring{$\beta_1=0$}{beta1 = 0}: no harmonic anchor}
\label{app:nobetti}
\noindent \Cref{sec:trichotomy} assumes $\beta_1>0$ throughout, so that the harmonic sector supplies the rate $\rho_H$ at frequency zero. When $\beta_1=0$ there is no harmonic mode, $\lambda_{\min}(\mcL_1)=\min\{\nu_g^{\min},\nu_c^{\min}\}>0$, and $\mathrm{RQ}(\mF)\ge\lambda_{\min}(\mcL_1)>0$ for every $\mF\neq0$, so HD in the sense of \cref{prop:regimeprops} cannot occur; the natural replacement of the harmonic sector is the bottom eigenspace of $\mcL_1$ (\cref{rem:bottom}). The endpoint reduction of \cref{lem:endpoint} still applies within each family, but the relevant endpoints are now the smallest and largest frequencies actually present. Define
\[
\rho_G^-
:=
\lambda_{\max}\!\left(
\mG^{\mathrm{grad}}_{\nu_g^{\min}}
\right),
\qquad
\rho_C^-
:=
\lambda_{\max}\!\left(
\mG^{\mathrm{curl}}_{\nu_c^{\min}}
\right).
\]

\begin{proposition}[No harmonic anchor]
\label{prop:no-harmonic-anchor}
Suppose $\beta_1=0$ and \cref{ass:nondeg} holds. Then the top rate of the flow \eqref{eq:flow} is
\[
\mu_*=\max\{\rho_G^-,\rho_G,\rho_C^-,\rho_C\}.
\]
If the maximum is attained by exactly one of the four rates, then for almost every $\mF(0)$ the normalized features concentrate on the corresponding family and endpoint frequency, and
\[
(\mathrm{RQ}_{\mathrm{down}},\mathrm{RQ}_{\mathrm{up}})
\longrightarrow
\begin{cases}
(\nu_g^{\max},0), & \rho_G\text{ largest (GD)},\\
(0,\nu_c^{\max}), & \rho_C\text{ largest (CD)},\\
(\nu_g^{\min},0), & \rho_G^-\text{ largest},\\
(0,\nu_c^{\min}), & \rho_C^-\text{ largest}.
\end{cases}
\]
In the language of \cref{lem:gdcd}, the limit is the vertex of $\mathcal P$ whose endpoint rate is largest. A bottom-endpoint limit concentrates on the bottom eigenspace of $\mcL_1$, i.e.\ $\mathrm{RQ}\to\lambda_{\min}(\mcL_1)$, exactly when its frequency equals $\min\{\nu_g^{\min},\nu_c^{\min}\}$. If $\nu_g^{\min}\neq\nu_c^{\min}$ and the larger bottom frequency wins, the limiting frequency lies strictly between $\lambda_{\min}(\mcL_1)$ and $\lambda_{\max}(\mcL_1)$, and the flow is neither GD, nor CD, nor concentrated on the bottom eigenspace.
\end{proposition}

\begin{proof}
\emph{Top rate.} With $r_g,r_c$ as in the proof of \cref{lem:endpoint}, $\rho_G^-=r_g(\nu_g^{\min})$, $\rho_G=r_g(\nu_g^{\max})$, and similarly for the curl family. If $\nu_g^{\min}<\nu_g^{\max}$, convexity of $r_g$ gives, for $\nu\in[\nu_g^{\min},\nu_g^{\max}]$ and $\theta_g(\nu):=(\nu-\nu_g^{\min})/(\nu_g^{\max}-\nu_g^{\min})\in[0,1]$,
\begin{equation}
\label{eq:nobetti-bound}
r_g(\nu)\le\bigl(1-\theta_g(\nu)\bigr)\rho_G^-+\theta_g(\nu)\,\rho_G\le\max\{\rho_G^-,\rho_G\};
\end{equation}
if $\nu_g^{\min}=\nu_g^{\max}$ the family has a single frequency and $\rho_G^-=\rho_G$. Both endpoints belong to $\Sigma_g$, so $\max_{\nu\in\Sigma_g}r_g(\nu)=\max\{\rho_G^-,\rho_G\}$, and likewise $\max_{\nu\in\Sigma_c}r_c(\nu)=\max\{\rho_C^-,\rho_C\}$. Since $\beta_1=0$, every $\vu_j$ is a gradient or a curl mode, so by \eqref{eq:specgen}
$\mu_*=\max_j\lambda_{\max}(\mG_j)=\max\{\rho_G^-,\rho_G,\rho_C^-,\rho_C\}$. This proves the first assertion without any uniqueness assumption.

\smallskip
\noindent\emph{Sector identification.} Suppose $\rho_G^-$ is the unique maximum. Then $\rho_G^->\rho_G$ forces $\nu_g^{\min}<\nu_g^{\max}$, since equal frequencies have equal generators. For every gradient frequency $\nu>\nu_g^{\min}$ we have $\theta_g(\nu)>0$, so \eqref{eq:nobetti-bound} gives $r_g(\nu)<\rho_G^-$, and every curl mode has $\lambda_{\max}(\mG_j)\le\max\{\rho_C^-,\rho_C\}<\rho_G^-$. Hence $\mu_{j,r}=\mu_*$ forces $\vu_j$ to be a gradient mode of frequency exactly $\nu_g^{\min}$. If $\rho_G$ is the unique maximum the same argument applies with $1-\theta_g(\nu)>0$ for $\nu<\nu_g^{\max}$, and the two curl cases exchange the roles of the families. In each case all indices in $T_*$ share one frequency pair, so \cref{prop:closedform} gives, for almost every $\mF(0)$, concentration on that eigenspace and the stated limit of the Rayleigh-quotient pair. No simultaneous diagonalization of the channel weights is required.

\smallskip
\noindent\emph{Classification.} The first two limits are GD and CD by \cref{prop:regimeprops}. Since $\beta_1=0$, every joint eigenmode is a gradient or curl mode and is an eigenvector of $\mcL_1$ with eigenvalue equal to its frequency, so $\operatorname{Spec}(\mcL_1)=\Sigma_g\cup\Sigma_c$ and $\lambda_{\min}(\mcL_1)=\min\{\nu_g^{\min},\nu_c^{\min}\}$. In a bottom-gradient limit $\mathrm{RQ}\to\nu_g^{\min}$, which equals $\lambda_{\min}(\mcL_1)$ exactly when $\nu_g^{\min}\le\nu_c^{\min}$, and \cref{rem:bottom} then gives concentration on the bottom eigenspace; the bottom-curl case is symmetric. If instead $\nu_g^{\min}>\nu_c^{\min}$ and $\rho_G^-$ wins, then $\mathrm{RQ}\to\nu_g^{\min}>\lambda_{\min}(\mcL_1)$, while $\mathrm{RQ}_{\mathrm{down}}\to\nu_g^{\min}<\nu_g^{\max}$ and $\mathrm{RQ}_{\mathrm{up}}\to0<\nu_c^{\max}$, so the flow is neither GD nor CD, and its limiting frequency satisfies $\lambda_{\min}(\mcL_1)<\nu_g^{\min}<\nu_g^{\max}\le\lambda_{\max}(\mcL_1)$.
\end{proof}

\noindent \Cref{prop:no-harmonic-anchor} shows why concentration on the harmonic subspace alone is not a sufficient description of asymptotic behavior in simplicial complexes: when $\ker\mcL_1=\{0\}$ the dynamics can still concentrate on the lowest available Hodge frequency, and it can also settle at a frequency that is extreme within one Hodge family but intermediate in the full spectrum.

\subsection{Commuting channel weights}
\label{app:commuting}
\begin{corollary}[Commuting channel weights]
\label{cor:commuting}
Suppose $\mW_0$, $\mW_1$ and $\mW_2$ commute pairwise, and let $\{\vphi_r\}_{r=1}^d$ be a common orthonormal eigenbasis with eigenvalue triples $(\mu_r^{(0)},\mu_r^{(1)},\mu_r^{(2)})$. Then
\[
\rho_H=\max_r\mu_r^{(0)},
\qquad
\rho_G=\max_r\bigl(\mu_r^{(0)}+\nu_g^{\max}\mu_r^{(1)}\bigr),
\qquad
\rho_C=\max_r\bigl(\mu_r^{(0)}+\nu_c^{\max}\mu_r^{(2)}\bigr),
\]
where in each maximum the eigenvalues with index $r$ belong to the same eigenvector and are not maximized independently. For the adjacency-like tying of \cref{sec:instantiation} with commuting $\mW_{\mathrm{down}},\mW_{\mathrm{up}}$ and joint eigenvalue pairs $(\mu_r^{\mathrm{down}},\mu_r^{\mathrm{up}})$, this reads
\[
\rho_H=\max_r\bigl(\mu_r^{\mathrm{down}}+\mu_r^{\mathrm{up}}\bigr),
\qquad
\rho_G=\max_r\bigl((1-2\nu_g^{\max})\mu_r^{\mathrm{down}}+\mu_r^{\mathrm{up}}\bigr),\]
\[
\rho_C=\max_r\bigl(\mu_r^{\mathrm{down}}+(1-2\nu_c^{\max})\mu_r^{\mathrm{up}}\bigr),
\]
The trichotomy conditions thus reduce to explicit inequalities between the channel eigenvalues and the two extreme Hodge frequencies.
\end{corollary}
\begin{proof}
Commuting symmetric matrices are simultaneously diagonalizable by an orthonormal basis. In that basis $\mG^{\mathrm{grad}}_\nu\vphi_r=(\mu_r^{(0)}+\nu\mu_r^{(1)})\vphi_r$, so $\lambda_{\max}(\mG^{\mathrm{grad}}_\nu)=\max_r(\mu_r^{(0)}+\nu\mu_r^{(1)})$; evaluating at $\nu=0$ and $\nu=\nu_g^{\max}$ gives $\rho_H$ and $\rho_G$, and $\rho_C$ follows in the same way from $\mG^{\mathrm{curl}}_\nu$. The adjacency-like form substitutes $\mW_0=\mW_{\mathrm{down}}+\mW_{\mathrm{up}}$, $\mW_1=-2\mW_{\mathrm{down}}$, $\mW_2=-2\mW_{\mathrm{up}}$, which commute pairwise whenever $\mW_{\mathrm{down}}$ and $\mW_{\mathrm{up}}$ do.
\end{proof}

\subsection{Arbitrary Rank extension}
\label{app:rank_extension}
\begin{remark}[Extension to arbitrary ranks]
\label{rem:arbitrary-rank}
Every result in \cref{sec:theory} (\cref{prop:regimeprops,thm:gradflow,prop:closedform,lem:endpoint,thm:trichotomy}) together with \cref{prop:flip}, the finite-depth bound of \cref{sec:finite-depth} and \cref{prop:nonlinear}, extends verbatim to rank-$k$ features $\mF(t)\in\mathbb R^{N_k\times d}$ for any $k\ge1$ once $(\mcL_1^{\mathrm{down}},\mcL_1^{\mathrm{up}},\mcL_1,\beta_1)$ are replaced by $(\mcL_k^{\mathrm{down}},\mcL_k^{\mathrm{up}},\mcL_k,\beta_k)$ and the gradient and curl subspaces are replaced by the rank-$k$ coboundary $\operatorname{im}(\hat\mB_k^\top)$ and boundary $\operatorname{im}(\hat\mB_{k+1})$ subspaces. At rank~$k$, \cref{ass:nondeg} requires both $\operatorname{im}(\hat\mB_k^\top)$ and $\operatorname{im}(\hat\mB_{k+1})$ to be nontrivial (so the complex has at least one $k$-simplex and one $(k+1)$-simplex) with largest eigenvalues satisfying
\begin{equation}
\nu_g^{\max}>\lambda_{\min}(\mcL_k),
\qquad
\nu_c^{\max}>\lambda_{\min}(\mcL_k).
\end{equation}
Under these assumptions, the proofs in \cref{app:rqlemmas,app:proofs,app:discrete} carry over without change because they use only the symmetry, positive semidefiniteness, and orthogonality $\mcL_k^{\mathrm{down}}\mcL_k^{\mathrm{up}}\!=\!0$ of the lower and upper Hodge operators (\cref{app:r2l}, NH1). At a boundary rank where one of the two operators vanishes ($k\!=\!0$ with $\mcL_0^{\mathrm{down}}\!=\!0$, or $k\!=\!\dim X$ with $\mcL_k^{\mathrm{up}}\!=\!0$), the three-way trichotomy reduces to a two-way dichotomy between harmonic modes and the remaining highest-frequency family (\cref{app:graph}).
\end{remark}

\section{Discretization of the flow}
\label{app:discrete}

\noindent We return from the continuous flow to the discrete layer update. By \cref{app:normlimit}, after vectorization the flow reads $\dot\vf=\gen\vf$ with the symmetric generator $\gen$ of \eqref{eq:vecgen}, whose eigenpairs are $(\mu_{j,r},\vpsi_{j,r}\otimes\vu_j)$. The layer update \eqref{eq:simplicial-layer} with $\sigma=\mathrm{id}$ and shared symmetric weights is the forward Euler step with step size $\tau>0$,
\[
\vf_{m+1}
=
(\mI+\tau \gen)\vf_m,
\]
so a component along an eigenvector with $\gen\vpsi=\eta\vpsi$ evolves as $e^{\eta t}$ in continuous time and as $(1+\tau\eta)^m$ in discrete depth. The continuous flow is therefore governed by the largest eigenvalue of $\gen$, whereas the discrete iteration is governed by the eigenvalue for which $|1+\tau\eta|$ is largest. Let
\[
\eta_{\min}:=\lambda_{\min}(\gen)=\mu_{\min},
\qquad
\eta_{\max}:=\lambda_{\max}(\gen)=\mu_*,
\]
the smallest and largest rates of \cref{sec:trichotomy} (by \eqref{eq:specgen}), and assume $\eta_{\min}<\eta_{\max}$. Since
$\eta\mapsto|1+\tau\eta|$ is convex, its maximum over
$\mathrm{Spec}(\gen)\subseteq[\eta_{\min},\eta_{\max}]$ is attained at one
of these two endpoints. Comparing their moduli gives the following result.

\begin{proposition}[Step size condition]
\label{prop:flip}
For almost every initial condition $\mF_0$:
\begin{itemize}
    \item[(i)] If
    \[
    \tau(\eta_{\max}+\eta_{\min})>-2,
    \]
    then
    \[
    |1+\tau\eta_{\max}|
    >
    |1+\tau\eta_{\min}|,
    \]
    and in fact $1+\tau\eta_{\max}>0$. The discrete iteration is therefore
    asymptotically dominated by the top eigenspace of $\gen$, just as in
    continuous time. Hence it has the same HD, GD, or CD regime predicted
    by the continuous-time analysis.

    \item[(ii)] If
    \[
    \tau(\eta_{\max}+\eta_{\min})<-2,
    \]
    then
    \[
    |1+\tau\eta_{\min}|
    >
    |1+\tau\eta_{\max}|,
    \qquad
    1+\tau\eta_{\min}<0.
    \]
    The bottom eigenspace of $\gen$ instead dominates. Its direction therefore
    alternates sign from one layer to the next while its magnitude dominates
    the remaining components.
\end{itemize}
At equality,
\[
\tau(\eta_{\max}+\eta_{\min})=-2,
\]
the two endpoint multipliers have equal modulus and the limiting behavior
depends on the initial projections onto both eigenspaces.
\end{proposition}

\begin{proof}
Let $\vf_m=\operatorname{vec}(\mF_m)$ and write
\[
\mT_\tau:=\mI+\tau\gen,\qquad\vf_m=\mT_\tau^m\vf_0 .
\]
Throughout, $\eta_{\min}=\lambda_{\min}(\gen)<\lambda_{\max}(\gen)=\eta_{\max}$ and $\tau>0$. All top and bottom eigenspaces below are eigenspaces of the full feature generator $\gen$, not of the Hodge Laplacian. For each distinct eigenvalue $\eta$ of $\gen$, let $\mPi_\eta$ denote its orthogonal spectral projector, and set
\[
\mPi_+:=\mPi_{\eta_{\max}},\qquad\mPi_-:=\mPi_{\eta_{\min}},\qquad\vv_+:=\mPi_+\vf_0,\qquad\vv_-:=\mPi_-\vf_0 .
\]

\smallskip
\noindent\emph{Moduli of the Euler multipliers.} If $\gen\vpsi=\eta\vpsi$, then $\mT_\tau\vpsi=(1+\tau\eta)\vpsi$. Thus $\mT_\tau$ has the same eigenspaces as $\gen$, with eigenvalue $\alpha_\eta:=1+\tau\eta$ on the $\eta$-eigenspace, and for every $m\ge1$,
\[
\vf_m=\sum_{\eta\in\operatorname{Spec}(\gen)}\alpha_\eta^m\,\mPi_\eta\vf_0 .
\]
The magnitude of each component is multiplied by $|\alpha_\eta|^m$, so we must maximize the modulus, not the signed multiplier. Write $\alpha_+:=1+\tau\eta_{\max}$ and $\alpha_-:=1+\tau\eta_{\min}$. The squared modulus $h(\eta):=(1+\tau\eta)^2$ is strictly convex, since $h''(\eta)=2\tau^2>0$. For an interior value $\eta=(1-\theta)\eta_{\min}+\theta\eta_{\max}$ with $0<\theta<1$,
\[
h(\eta)<(1-\theta)h(\eta_{\min})+\theta h(\eta_{\max})\le\max\{h(\eta_{\min}),h(\eta_{\max})\}.
\]
Both endpoints belong to the spectrum of $\gen$. Therefore $\max_{\eta\in\operatorname{Spec}(\gen)}|\alpha_\eta|=\max\{|\alpha_-|,|\alpha_+|\}$, and no interior eigenvalue can attain this maximum. To compare the two endpoints, factor the difference of their squares:
\[
|\alpha_+|^2-|\alpha_-|^2=\tau(\eta_{\max}-\eta_{\min})\bigl[2+\tau(\eta_{\max}+\eta_{\min})\bigr].
\]
The first factor is strictly positive, so the sign of the difference is exactly the sign of $2+\tau(\eta_{\max}+\eta_{\min})$. This proves the modulus comparisons in parts (i) and (ii), as well as equality of the moduli at the stated threshold. For the signs of the winning multipliers, note that
\[
\alpha_+-\alpha_-=\tau(\eta_{\max}-\eta_{\min})>0,\qquad\alpha_++\alpha_-=2+\tau(\eta_{\max}+\eta_{\min}).
\]

\smallskip
\noindent\emph{Top-eigenspace dominance in part (i).} Suppose $\tau(\eta_{\max}+\eta_{\min})>-2$. Then $|\alpha_+|>|\alpha_-|$, and $\alpha_+=\tfrac12\bigl[(\alpha_++\alpha_-)+(\alpha_+-\alpha_-)\bigr]>0$. Every other eigenvalue $\eta<\eta_{\max}$ satisfies $|\alpha_\eta|<\alpha_+$. Factoring out the winning multiplier gives
\[
\vf_m=\alpha_+^m\bigl(\vv_++\vr_m\bigr),\qquad
\vr_m:=\sum_{\eta<\eta_{\max}}\Bigl(\frac{\alpha_\eta}{\alpha_+}\Bigr)^m\mPi_\eta\vf_0\to0 .
\]
If $\vv_+\neq0$, orthogonality of the spectral eigenspaces yields $\vf_m/\|\vf_m\|_2\to\vv_+/\|\vv_+\|_2$. This is the same normalized limit as in continuous time (\cref{prop:closedform}), since $e^{-\eta_{\max}t}e^{t\gen}\vf_0\to\vv_+$. Continuity of the quadratic forms defining the Rayleigh quotients then gives the same limiting Rayleigh quotients. Consequently, whenever the continuous-time analysis establishes an HD, GD, or CD regime, the discrete iteration has that regime as well.

\smallskip
\noindent\emph{Alternating bottom-eigenspace dominance in part (ii).} Suppose $\tau(\eta_{\max}+\eta_{\min})<-2$. Then $|\alpha_-|>|\alpha_+|$, and $\alpha_-=\tfrac12\bigl[(\alpha_++\alpha_-)-(\alpha_+-\alpha_-)\bigr]<0$. Set $R:=|\alpha_-|=-\alpha_->0$. Every other eigenvalue $\eta>\eta_{\min}$ has $|\alpha_\eta|<R$, so
\[
\vf_m=R^m\bigl((-1)^m\vv_-+\vr_m\bigr),\qquad
\vr_m:=\sum_{\eta>\eta_{\min}}\Bigl(\frac{\alpha_\eta}{R}\Bigr)^m\mPi_\eta\vf_0\to0 .
\]
If $\vv_-\neq0$, orthogonality gives $(-1)^m\vf_m/\|\vf_m\|_2\to\vv_-/\|\vv_-\|_2$. Thus the bottom eigenspace of $\gen$ dominates relative to all remaining components, while the normalized even and odd subsequences approach opposite directions. This proves part (ii), regardless of whether the unnormalized feature norm grows or decays.

\smallskip
\noindent\emph{Equality.} At $\tau(\eta_{\max}+\eta_{\min})=-2$, we have $\alpha_+=-\alpha_-=\tfrac\tau2(\eta_{\max}-\eta_{\min})=:R>0$, and all interior eigenvalues have multiplier modulus strictly smaller than $R$. Hence
\[
\vf_m=R^m\bigl(\vv_++(-1)^m\vv_-+\vr_m\bigr),\qquad\vr_m\to0,
\]
and, with $D:=\sqrt{\|\vv_+\|_2^2+\|\vv_-\|_2^2}>0$,
\[
\frac{\vf_{2k}}{\|\vf_{2k}\|_2}\longrightarrow\frac{\vv_++\vv_-}{D},\qquad
\frac{\vf_{2k+1}}{\|\vf_{2k+1}\|_2}\longrightarrow\frac{\vv_+-\vv_-}{D}.
\]
These formulas describe the dependence on both initial projections.

Finally, the nonzero-projection conditions used above hold for almost every initial condition: the excluded sets are $\ker\mPi_+$, $\ker\mPi_-$, and $\ker\mPi_+\cap\ker\mPi_-$ respectively, each a proper linear subspace of $\mathbb R^{N_1d}$ and therefore of Lebesgue measure zero. The step-size formulation follows by rearranging part (i): if $\eta_{\max}+\eta_{\min}<0$, its condition is equivalent to $0<\tau<2/\bigl(-(\eta_{\max}+\eta_{\min})\bigr)$; if $\eta_{\max}+\eta_{\min}\ge0$, it holds for every $\tau>0$.
\end{proof}

\noindent This is a consequence of discretization with no continuous-time counterpart. A very negative continuous rate decays rapidly under
$e^{\eta t}$, but if the Euler step is sufficiently large then
$1+\tau\eta<-1$, so the same mode grows in magnitude while alternating
sign. Therefore we impose a step-size restriction in the discretized flow model, just as Di Giovanni et
al.~\citep{digiovanni2023understandingconvolutiongraphsenergies}
do to exclude this behavior in the graph setting. Equivalently, if
$\eta_{\max}+\eta_{\min}<0$, preservation of the continuous-time
asymptotic regime requires
\[
\tau
<
\frac{2}{-(\eta_{\max}+\eta_{\min})},
\]
and if $\eta_{\max}+\eta_{\min}\geq0$ the condition in
\cref{prop:flip}(i) holds for every $\tau>0$. The proof uses only the symmetry of the generator, so the proposition holds verbatim for the general flow \eqref{eq:generalflow}.

\begin{proof}[Proof of the linear finite-depth bound]
Under the hypotheses of \cref{thm:trichotomy} and the step-size condition
of \cref{prop:flip}(i), assume $\mF_0\not\perp E_*$.
For $m\geq 1$, define
\[
s_m:=
\frac{
\sum_{(j,r)\notin T_*}a_{j,r}^{2}
\left|\frac{1+\tau\mu_{j,r}}{1+\tau\mu_*}\right|^{2m}
}{
\sum_{(j,r)\in T_*}a_{j,r}^{2}
}
\leq C\kappa^{2m}.
\]
By orthogonality, $s_m/(1+s_m)$ is the fraction of squared
feature mass outside $E_*$. Every leading mode has frequency
$q_\infty$, and all remaining frequencies differ from it by
at most $\nu_{\max}$. Hence
\[
|q(\mF_m)-q_\infty|
\leq \nu_{\max}\frac{s_m}{1+s_m}
\leq \nu_{\max}\frac{C\kappa^{2m}}{1+C\kappa^{2m}}.
\]
This completes the proof.
\end{proof}

\begin{proof}[Proof of Proposition~\ref{prop:nonlinear} for $\sigma=\tanh$]
Let $\vf_m=\operatorname{vec}(\mF_m)\in\mathbb R^{N_1d}$ and $\mT_\tau=\mI+\tau\gen$ as in the proof of \cref{prop:flip}. With $\mPi_+$ the orthogonal projector onto $\operatorname{vec}(E_*)=\ker(\gen-\mu_*\mI)$ and $\mPi_\perp:=\mI-\mPi_+$, the vectorized components satisfy
\begin{equation}
\vf_m^+:=\operatorname{vec}(\mF_m^+)=\mPi_+\vf_m,
\quad
\vf_m^\perp:=\operatorname{vec}(\mF_m^\perp)=\mPi_\perp\vf_m,
\quad
\|\mF_m^+\|=\|\vf_m^+\|_2,
\quad
\|\mF_m^\perp\|=\|\vf_m^\perp\|_2.
\end{equation}

Write the vectorized nonlinear update as
\begin{equation}
\vf_{m+1}=\tanh(\mT_\tau\vf_m)=\mT_\tau\vf_m+\mathbf{e}_m,
\qquad
\mathbf{e}_m:=\tanh(\mT_\tau\vf_m)-\mT_\tau\vf_m.
\end{equation}
Under the step-size condition of \cref{prop:flip}(i), $\alpha_+=1+\tau\mu_*=\|\mT_\tau\|_2>0$. Since $|\tanh(x)|\le|x|$ elementwise, we have
\begin{equation}
\|\vf_{m+1}\|_2\le\|\mT_\tau\vf_m\|_2\le\alpha_+\|\vf_m\|_2,
\qquad
\|\mT_\tau\vf_m\|_2\le\alpha_+^{m+1}\|\vf_0\|_2
\quad\text{for all }m\ge0.
\end{equation}
Using $|\tanh(x)-x|\le\tfrac13|x|^3$ and $\|\vz\|_6\le\|\vz\|_2$, we obtain
\begin{equation}
\|\mathbf{e}_m\|_2
\le\tfrac13\|\mT_\tau\vf_m\|_6^3
\le\tfrac13\|\mT_\tau\vf_m\|_2^3
\le\tfrac13\alpha_+^{3(m+1)}\|\vf_0\|_2^3.
\end{equation}
Because $\mT_\tau$ is symmetric, it commutes with $\mPi_+$ and $\mPi_\perp$, acts as $\alpha_+\mI$ on $\operatorname{im}\mPi_+$, and satisfies $\|\mPi_\perp\mT_\tau\vv\|_2\le\kappa\alpha_+\|\vv\|_2$ on $\operatorname{im}\mPi_\perp$. Projecting the update onto $\operatorname{im}\mPi_+$ and $\operatorname{im}\mPi_\perp$ gives
\begin{equation}
\vf_{m+1}^+=\alpha_+\vf_m^++\mPi_+\mathbf{e}_m,
\qquad
\|\vf_{m+1}^\perp\|_2\le\kappa\alpha_+\|\vf_m^\perp\|_2+\|\mPi_\perp\mathbf{e}_m\|_2.
\end{equation}
Unrolling both relations from $k=0$ to $m-1$, dividing by $\alpha_+^m$, and using
\begin{equation}
\max\{\|\mPi_+\mathbf{e}_k\|_2,\|\mPi_\perp\mathbf{e}_k\|_2\}
\le\|\mathbf{e}_k\|_2
\le\tfrac13\alpha_+^{3(k+1)}\|\vf_0\|_2^3,
\quad
\delta=\frac{\|\vf_0\|_2^3}{3\|\vf_0^+\|_2},
\quad
\|\vf_0^\perp\|_2=\sqrt{C}\,\|\vf_0^+\|_2,\nonumber
\end{equation}
yields
\begin{equation}
\alpha_+^{-m}\|\vf_m^+\|_2
\ge\|\vf_0^+\|_2-\tfrac13\|\vf_0\|_2^3\sum_{k=0}^{m-1}\alpha_+^{2(k+1)}
=\|\vf_0^+\|_2(1-\delta\Phi_m),
\end{equation}
and
\begin{equation}
\alpha_+^{-m}\|\vf_m^\perp\|_2
\le\kappa^m\|\vf_0^\perp\|_2+\tfrac13\|\vf_0\|_2^3\sum_{k=0}^{m-1}\kappa^{m-1-k}\alpha_+^{2(k+1)}
=\|\vf_0^+\|_2\bigl(\sqrt{C}\,\kappa^m+\delta\Psi_m\bigr).
\end{equation}
Whenever $\delta\Phi_m<1$, the lower bound implies $\|\mF_m^+\|=\|\vf_m^+\|_2>0$, so $s_m=\|\mF_m^\perp\|^2/\|\mF_m^+\|^2$ is well defined and satisfies the bound in~\eqref{eq:nonlinear-finite-depth}.

To bound $|q(\mF_m)-q_\infty|$, let $\mcL\in\{\mcL_1^{\mathrm{down}},\mcL_1^{\mathrm{up}}\}$ be the operator defining $q$. Under the strict-maximum hypothesis of \cref{thm:trichotomy}, every basis matrix $\vu_j\vpsi_{j,r}^\top$ spanning $E_*$ has the same eigenvalue $q_\infty\in\{0,\nu_{\max}\}$ under left-multiplication by $\mcL$, so $\mcL\mF_m^+=q_\infty\mF_m^+$. Since $\mcL$ is symmetric and $\langle\mF_m^+,\mF_m^\perp\rangle_F=0$, the cross terms vanish:
\begin{equation}
\mathrm{Tr}\bigl(\mF_m^\top(\mcL-q_\infty\mI)\mF_m\bigr)
=\mathrm{Tr}\bigl((\mF_m^\perp)^\top(\mcL-q_\infty\mI)\mF_m^\perp\bigr).
\end{equation}
Because $\operatorname{Spec}(\mcL)\subseteq[0,\nu_{\max}]$ and $q_\infty\in\{0,\nu_{\max}\}$, we have $\|\mcL-q_\infty\mI\|_2\le\nu_{\max}$, while $\|\mF_m\|^2=\|\mF_m^+\|^2+\|\mF_m^\perp\|^2$. Dividing by $\|\mF_m\|^2$ gives
\begin{equation}
|q(\mF_m)-q_\infty|
\le\frac{\nu_{\max}\|\mF_m^\perp\|^2}{\|\mF_m^+\|^2+\|\mF_m^\perp\|^2}
=\nu_{\max}\frac{s_m}{1+s_m},
\end{equation}
completing part~(i).

\textbf{For part~(ii)}, if $0<\alpha_+<1$, then
\begin{equation}
\Phi_m<\sum_{k=1}^\infty\alpha_+^{2k}=\frac{\alpha_+^2}{1-\alpha_+^2},
\end{equation}
so~$\mF_0$ satisfying $\delta\,\le\, {(1-\alpha_+^{p-1})}/{(2\alpha_+^{p-1})}$, ensures $\delta\Phi_m<\tfrac12$ for all $m\ge1$. Put $b:=\alpha_+^2<1$ and $\tilde\kappa:=\max\{\kappa,b\}<1$. Then
\begin{equation}
\Psi_m=
\begin{cases}
\displaystyle b\,\frac{\kappa^m-b^m}{\kappa-b}, & \kappa\ne b,\\[6pt]
m\,b^m, & \kappa=b,
\end{cases}
\end{equation}
so $\Psi_m\le m\tilde\kappa^m$ in all cases. Substituting into~\eqref{eq:nonlinear-finite-depth} gives~\eqref{eq:nonlinear-asymptotic}. When $\kappa\ne b$, $\Psi_m=\mathcal{O}(\tilde\kappa^m)$ and the factor $m^2$ in~\eqref{eq:nonlinear-asymptotic} can be omitted.
\end{proof}

\noindent\textbf{Depth range of the finite-depth bound.} Part~(i) of \cref{prop:nonlinear} holds only at depths with $\delta\Phi_m<1$. Write $b:=\alpha_+^{p-1}$. When $\alpha_+<1$, $\Phi_m<b/(1-b)$ for every $m$, so the condition holds at all depths for sufficiently small inputs, which is part~(ii). When $\alpha_+\ge1$, as for flows that amplify their top eigenspace, $\Phi_m$ grows without bound; for $\alpha_+>1$, summing the geometric series $\Phi_m=\sum_{k=1}^m b^k$ shows that $\delta\Phi_m<1$ exactly when
\begin{equation}
m<\frac{1}{\log b}\log\Bigl(1+\frac{b-1}{b\,\delta}\Bigr),
\end{equation}
a depth that shrinks as $\delta=M_\sigma\|\mF_0\|^{p-1}/\gamma_0$ grows. Beyond it, the nonlinear remainder may outweigh the linear spectral gap, and the bound no longer guarantees the linear endpoint.

\noindent\textbf{General activations.} The proof above is written for $\sigma=\tanh$, which satisfies~\eqref{eq:activation-bounds} with $p=3$ and $M_\sigma=\tfrac13$. For a general activation satisfying~\eqref{eq:activation-bounds}, the same argument applies with $\|\mathbf{e}_m\|_2\le M_\sigma\|\mT_\tau\vf_m\|_{2p}^p\le M_\sigma\alpha_+^{p(m+1)}\|\vf_0\|_2^p$, which yields $\delta$, $\Phi_m$, $\Psi_m$, $\tilde\kappa$ and $\mF_0$ satisfying $\delta\,\le\, {(1-\alpha_+^{p-1})}/{(2\alpha_+^{p-1})}$, as stated in \cref{prop:nonlinear}; for the identity, $\mathbf{e}_m=0$.

\paragraph{Intuition and implications} In the linear regime, the lower and upper Hodge Laplacians commute and the three Hodge subspaces decouple completely. Each joint eigenmode evolves independently, and the dominant eigenspace $E_*$ (with multiplier $\alpha_+ = 1 + \tau\mu_*$) outgrows all other modes at the geometric rate $\kappa < 1$. Introducing an elementwise activation $\sigma = \tanh$ breaks this invariance: applying $\tanh$ entrywise mixes coordinates in the simplex basis and leaks feature energy across the harmonic, gradient, and curl subspaces at every layer.

Our proposition shows that the linear trichotomy nonetheless governs the nonlinear dynamics whenever this cross-subspace leakage remains smaller than the linear spectral gap. The mechanism relies only on the two bounds~\eqref{eq:activation-bounds} on the activation. At each layer, the linear step suppresses off-target modes relative to $E_*$ by the factor $\kappa$, while the nonlinearity injects a perturbation controlled by $\delta = \|\mathbf{F}_0\|^2 / (3\gamma_0)$, which weighs the initial feature scale $\|\mathbf{F}_0\|^2$ against the initial alignment $\gamma_0 = \|\mathbf{F}_0^+\| / \|\mathbf{F}_0\|$.

Unrolling this over $m$ layers yields two distinct regimes depending on whether the dominant linear multiplier $\alpha_+$ is expansive or contractive:

\begin{enumerate}[leftmargin=*,topsep=0.01em,itemsep=0.01em]
    \item \textbf{Finite-depth regime (Part i, any $\alpha_+ > 0$):} At any finite depth $m$ where the cumulative nonlinear perturbation satisfies $\delta\Phi_m < 1$, the dominant projection $\mathbf{F}_m^+$ cannot be canceled by nonlinear leakage, and the Rayleigh quotient distance to the linear endpoint $q_\infty$ is explicitly bounded by the sum of the linear decay term $\sqrt{C}\kappa^m$ and the accumulated nonlinear error $\delta\Psi_m$. When $\alpha_+ \ge 1$, features can grow across layers and $\Phi_m$ eventually exceeds $1/\delta$ as $\tanh$ enters its saturated regime, where the bound becomes vacuous. At practical training depths (such as $m = 16$ in \cref{fig:drifters-tanh}), however, the bound is consistent with nonlinear networks closely tracking the linear HD, GD, and CD trajectories before saturation sets in.
    \item \textbf{Asymptotic regime (Part ii, contractive flows $0 < \alpha_+ < 1$):} When $\mu_* < 0$, the step-size condition places $\alpha_+$ in $(0,1)$, so the feature norm decays at rate $\alpha_+^m$ and the cubic nonlinear error decays even faster, at rate $\alpha_+^{3m}$. After rescaling by $\alpha_+^{-m}$ to track the normalized direction, the per-layer leakage decays geometrically as $\alpha_+^{2k}$, making the infinite series $\Phi_\infty$ convergent. For any initial feature with $\mathbf{F}_0^+ \ne \mathbf{0}$ whose norm satisfies the basin condition, nonlinear effects vanish asymptotically relative to the dominant linear mode, and the normalized representation converges to the exact linear HD, GD, or CD endpoint at rate $\mathcal{O}(m^2 \tilde{\kappa}^{2m})$.
\end{enumerate}

\section{Adjacency-like versus Laplacian propagation}
\label{app:AvsL}

\noindent The two tyings of \cref{sec:instantiation} differ in how they treat the harmonic sector. Recall that a harmonic
mode has
\[
\nu_d=\nu_u=0,
\]
and therefore evolves according to the feature-space generator
\[
\mG^{\mathrm{harm}}
=
\mW_0.
\]
Thus the identity weight $\mW_0$ determines whether the harmonic
component is part of the learned dynamics at all.

\noindent For pure Laplacian propagation,
\[
\mW_0=0,
\]
so
\[
\mG^{\mathrm{harm}}=0.
\]
Then,
\[
\mP_{\mathrm{harm}}\mF(t)
=
\mP_{\mathrm{harm}}\mF(0)
\]
and the harmonic component is preserved exactly and cannot be amplified,
suppressed, or mixed across feature channels by the learned weights. This is
the higher-order analogue of the observation that pure graph-Laplacian
propagation leaves the lowest-frequency component unchanged
\citep{digiovanni2023understandingconvolutiongraphsenergies}.

\noindent By contrast, for the adjacency-like choice
\[
\mW_0=\mW_{\mathrm{down}}+\mW_{\mathrm{up}},
\]
we have
\[
\mG^{\mathrm{harm}}
=
\mW_{\mathrm{down}}+\mW_{\mathrm{up}},
\]
and hence
\[
\mP_{\mathrm{harm}}\mF(t)
=
\mP_{\mathrm{harm}}\mF(0)
\exp\!\bigl(
t(\mW_{\mathrm{down}}+\mW_{\mathrm{up}})
\bigr).
\]
The harmonic component can therefore be amplified, suppressed, and mixed
across feature channels. In this way, retaining the identity term allows the learned dynamics to control the harmonic subspace of the features. Recalling the growth rates and regime conditions of \cref{sec:trichotomy}, under pure Laplacian propagation the harmonic
growth rate is fixed at
\[
\rho_H=0,
\]
while adjacency-like propagation makes
\[
\rho_H
=
\lambda_{\max}
\bigl(\mW_{\mathrm{down}}+\mW_{\mathrm{up}}\bigr)
\]
learnable. The identity term therefore changes not only the evolution within
the harmonic subspace, but also whether harmonic modes can dominate or be
suppressed relative to the gradient and curl sectors.

\begin{remark}[Structural bias of pure Laplacian propagation]
\label{rem:initbias}
Let $\beta_1>0$. Under pure Laplacian propagation, $\rho_H=0$, $\rho_G=2\nu_g^{\max}\bigl(-\lambda_{\min}(\mW_{\mathrm{down}})\bigr)$ and $\rho_C=2\nu_c^{\max}\bigl(-\lambda_{\min}(\mW_{\mathrm{up}})\bigr)$. Hence the flow is HD if and only if $\mW_{\mathrm{down}}\succ0$ and $\mW_{\mathrm{up}}\succ0$ (ties aside), and it is GD or CD whenever either matrix has a negative eigenvalue. A generic symmetric initialization is indefinite, so pure Laplacian propagation is GD or CD at initialization with probability close to one, and harmonic dominance is reachable only if training makes both channel matrices positive definite. Even then the harmonic component is merely preserved, while the other sectors decay by at most a factor $e^{-2\nu\|\mW\|_2t}$ at frequency $\nu$. The adjacency-like flow is free from this restriction: $\rho_H=\lambda_{\max}(\mW_{\mathrm{down}}+\mW_{\mathrm{up}})$ is learnable and generically nonzero, so all three regimes are available at initialization and the selected regime is determined by training.
\end{remark}

\section{Additional Experiments \& Details}
\label{app:experimental-details}

\subsection{Main Paper Experimental Details}
\label{app:main-experimental-details}

\subsubsection{Datasets and input signals}
\label{app:datasets}

\paragraph{Four disks}
We reconstruct the example of \citet{grande2024hodge} by sampling 400
points uniformly in each radius-2 disk centered at $(0,0)$, $(0,4.2)$,
$(4.2,0)$ and $(4.2,4.2)$ (seed~0), and constructing a GUDHI alpha
complex with $\alpha=0.45$. It has 1600 vertices, 4671 edges, 3071
triangles and $\beta_1=1$. All tasks use this same complex and the
normalized operators of \cref{app:r2l}.

The shared eight-channel edge input mixes five spectral bands: the
lowest three gradient modes, lowest four curl modes, the harmonic mode,
and highest six modes of each nonzero family. For each band $U_b$, we
draw a Gaussian mixing matrix $R_b$ with eight columns, normalize
$U_bR_b$ to equal Frobenius norm across bands, and rescale their sum
to unit RMS (feature seed 20260918). Each observation adds independent
Gaussian noise of standard deviation 0.1. Training uses two fresh
fields per epoch; validation and testing use four and eight fixed
fields (seeds 81101 and 82202).

\paragraph{Ocean drifters}
We use the released GitHub data and outer splits for the SCoNe
ocean-drifter benchmark of Roddenberry et al., cited in Section~4.3.
The fixed complex has 133 vertices, 320 edges, 186 triangles and
$\beta_1=2$, with 200 trajectory-prefix examples. Inputs are signed
net edge-traversal counts. The target is the next vertex among the
prefix endpoint's neighbors. No coordinate features or synthetic
noise are added. We use the ten supplied 180/20 train/test splits unchanged. Our only additional partition is an inner 144/36
fitting/validation split within each training set, grouping identical
input/endpoint pairs and using seed $7000+s$ for split $s$.

\subsubsection{Four-disk target labels}
\label{app:four-disk-labels}

For the harmonic, gradient and curl tasks, following the spectral
clustering construction of \citet{grande2024hodge}, we cluster edge rows of three embeddings: the harmonic eigenvector,
the six highest-frequency gradient eigenvectors, and the six
highest-frequency curl eigenvectors from our normalized Hodge
Laplacians. Upper-frequency selection is our extension of the original
construction.

Before clustering, each row $z$ is multiplied by
$\operatorname{sign}(\mathbf1^\top z)$. For a zero sum, the first
largest-magnitude entry sets the sign. Zero rows remain zero.
Rows are not normalized. We apply $k$-means with 30 initializations
and seed~0, using two harmonic classes and five classes for each of the gradient
and curl tasks. Label classes are ordered by centroid norm. We note that gray denotes a supervised class (the remainder of the edges after selecting out the clusters) that are included in the classification task.

The region task needs no clustering since each edge is labeled by its disk membership.

All tasks share 2802 training, 930 validation and 939 test edges
(seed~823). We conduct a transductive experiment. The full complex is available, but optimization only uses training-edge labels.

\subsubsection{Models and training}
\label{app:models-training}

Both experiments evolve only edge features using the Euler flows of
Section~4.1. Channel weights are independent, symmetric and shared
across depth, with no commutation constraint. We do not include
activation, bias or feature normalization in or between the propagation layers. A nonlinear readout layer is applied only at the end.

\paragraph{Four disks}
We use adjacency-like propagation with width $d=8$, depth 32 and
$\tau=0.25$, without a learned input lift. Weights are parameterized
as $W=(Q+Q^\top)/2$, with initial entries of $Q$ drawn from
$\mathcal N(0,0.01^2)$ and no spectral-norm bound. The final field is
divided by its whole-field RMS. Each edge is then classified by an
affine map of its 36 quadratic features $f_i^2$ and $\sqrt2f_if_j$
($i<j$), retaining information from every Hodge sector.

We train with Adam for 250 epochs, learning rate 0.02, weight decay
$10^{-4}$ and gradient-norm clipping at 5. Each epoch uses one batch
of two noisy fields. Cross-entropy is restricted to training edges,
with class weights $N_{\mathrm{train}}/(K n_k)$. We evaluate validation
every five epochs and restore the minimum-loss checkpoint.
We use seeds $0,1,2$. Test macro-F1 pools the eight test fields.
Figure~3 shows seed~0, with error overlays for the first test field.

\paragraph{Ocean drifters}
Both flows use $F_0=xa^\top$ with learned $a\in\mathbb R^{16}$,
width 16, depth 16 and $\tau=0.1$. Each symmetric channel matrix
is parameterized as $W=Q/\max\{1,\|Q\|_2\}$. The candidate readout
concatenates the orientation-aligned candidate-edge feature with
$(\widehat B_1F)[u,:]$ at candidate vertex $u$, followed by an affine
$32\to16$ layer, $\tanh$, and a linear $16\to1$ score. Invalid
neighbors are masked before the softmax. Both models have 832
parameters and use paired initialization seeds.

For each split and architecture, Adam searches learning rates
$\{0.001,0.003,0.01\}$ and explicit squared-parameter penalties
$\{0,10^{-4}\}$, using batches of at most 32 and gradient clipping at 1.
Selection minimizes validation negative log-likelihood, with at most
1000 epochs and patience 100 (selection seed $12000+s$).
We then refit the chosen configuration on all 180 training examples
for the selected epoch count with seeds $0,1,2$, giving 30 checkpoints
per architecture.

\paragraph{Implementation}
Four-disk training uses float32 and two CPU threads; ocean training
uses float64 and one thread per worker. Both use PyTorch~2.12.0,
NumPy~2.4.6 and Python~3.13.7. Ocean runs used an eight-logical-CPU
ARM64 host, with no GPU; summed fitting times, including selection,
were 625.0\,s for adjacency-like and 587.0\,s for Laplacian flow.

\subsubsection{Spectral diagnostics}
\label{app:additional-results}

For each frozen model and joint spatial frequency
$(\nu_j^{\mathrm{down}},\nu_j^{\mathrm{up}})$, we diagonalize
\[
G_j=(s-2\nu_j^{\mathrm{down}})W_{\mathrm{down}}
   +(s-2\nu_j^{\mathrm{up}})W_{\mathrm{up}},
\]
where $s=1$ for adjacency-like and $s=0$ for pure-Laplacian propagation.
We compare the three endpoint rates
and verify
$2+\tau(\eta_{\max}+\eta_{\min})>0$ over all modal eigenvalues.
The Euler step-size check passes for every reported checkpoint.

Frozen inference uses exact Euler multipliers $(1+\tau\eta)^m$,
with common rescaling for numerical stability, and records the
lower and upper Rayleigh quotients defined in the main text.
Four-disk curves use the first test field through depth 65,536;
ocean curves extend to 1,024. Figure~4 shows means and pointwise
interquartile ranges over test occurrences, grouped by the
generator-predicted regime: 600 adjacency-like, 520 Laplacian-gradient
and 80 Laplacian-curl occurrences. Figure~2 instead shows each
checkpoint's mean over its 20 test inputs. These are frozen-flow
diagnostics, not retraining at each depth. The predicted asymptotic
limit is conditional on initial excitation of the leading eigenspace.

\subsection{Additional Experiments}
\label{app:additional-experiments}

\subsubsection{Additional experiments on ocean drifters}
\label{app:additional-ocean-drifters}

\paragraph{Nonlinear flows}
Our analysis concerns linear flows ($\sigma=\mathrm{id}$ in \eqref{eq:simplicial-layer}). To test whether the regime selection on ocean drifters persists under a nonlinearity, we repeat the experiment of \cref{sec:drifters} with $\sigma=\tanh$. We choose $\tanh$ because simplicial networks should be equivariant to the arbitrary choice of simplex orientations. Reversing the orientation of an edge flips the sign of its feature, so each layer must commute with such sign flips. The linear part of \eqref{eq:simplicial-layer} preserves this property. Elementwise activations that are odd, as $\tanh$ is but, e.g., ReLU is not, also preserve it.

\begin{figure}[ht]
\centering
\includegraphics[width=0.85\linewidth]{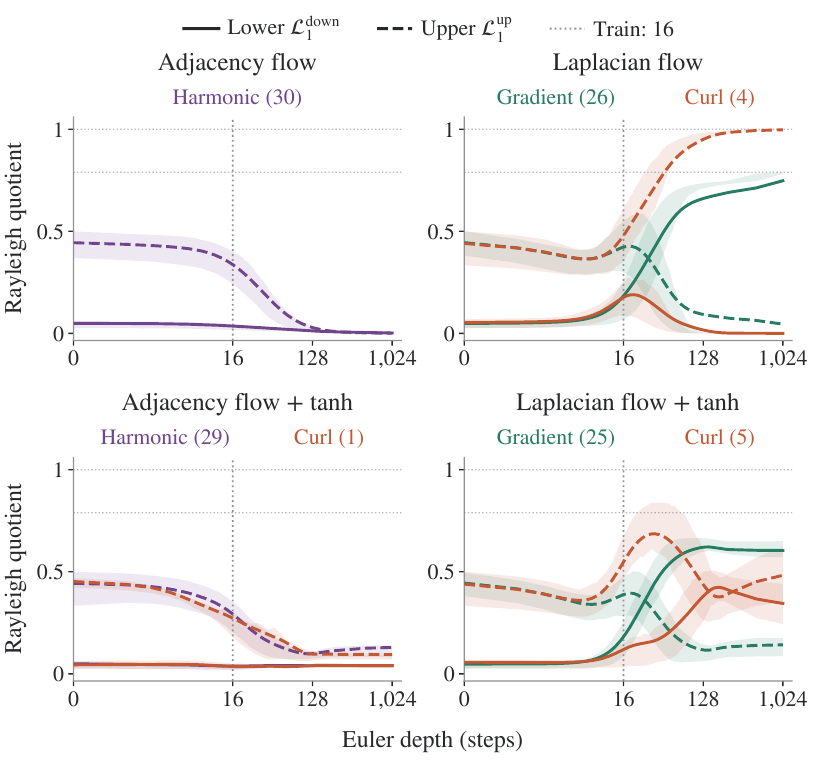}
\caption{\textbf{Frozen-flow spectral evolution on the ocean-drifter complex without and with a $\tanh$ nonlinearity}, for adjacency-like (left) and pure-Laplacian (right) flows. Top: linear flows, as in \cref{fig:drifters} (right). Bottom: $\sigma=\tanh$. Checkpoints are grouped by their predicted endpoint, with group sizes in parentheses. Lines indicate means and shading interquartile ranges across test occurrences. Vertical dotted lines mark training depth 16 and horizontal dotted lines the predicted limiting Rayleigh quotients.}
\label{fig:drifters-tanh}
\end{figure}

\Cref{fig:drifters-tanh} compares the linear flows (top row, as in \cref{fig:drifters}, right) with their $\tanh$ counterparts (bottom row). The regime predicted from the learned generators is nearly unchanged: 29 of the 30 adjacency-like checkpoints are predicted harmonic-dominant (one top-curl), and the pure-Laplacian checkpoints split into 25 top-gradient and 5 top-curl, against 26 and 4 in the linear case. The Rayleigh-quotient trajectories follow the same qualitative pattern, although with $\tanh$ they do not settle exactly at the linear endpoints.

\paragraph{Additional architectures}
\Cref{tab:drifters-architectures} compares our two flows with three published simplicial architectures on ocean drifters: SCoNe~\citep{roddenberry2021scone}, SCNN~\citep{yang2022scnn} of order 2 and SCCNN~\citep{yang2023sccnn} of order 1 (\cref{app:instantiations}), each in its published $\tanh$ form with 3 layers of width 16 and our normalized operators. Besides test accuracy, we track the Rayleigh-quotient pair through each network.

\begin{table}[ht]
\centering
\caption{\textbf{Additional architectures on ocean drifters.} Test accuracy is the mean $\pm$ standard deviation across the 10 released splits, each averaged over 3 seeds, under the selection protocol of \cref{app:models-training}. Each RQ cell is the pair $(\mathrm{RQ}_{\mathrm{down}},\mathrm{RQ}_{\mathrm{up}})$, averaged over test inputs and the 30 models (10 splits $\times$ 3 seeds); first, middle and last are layers 1, 2 and 3 for SCoNe, SCNN and SCCNN, and Euler steps 1, 8 and 16 for our flows. Every input starts at about $(0.049, 0.444)$. The regime endpoints are harmonic $(0,0)$, gradient $(0.789,0)$ and curl $(0,1.000)$. All three baselines use our normalized operators, under which the RQs are comparable across models.}
\label{tab:drifters-architectures}
\footnotesize
\begin{tabular*}{\linewidth}{@{\extracolsep{\fill}}l c ccc@{}}
\toprule
 & & \multicolumn{3}{c}{RQ pair (down, up)} \\
\cmidrule(l){3-5}
Model & Test accuracy & First & Middle & Last \\
\midrule
Adjacency-like flow (ours)        & $0.365 \pm 0.110$ & $(0.048, 0.436)$ & $(0.042, 0.398)$ & $(0.036, 0.337)$ \\
Pure-Laplacian flow (ours)        & $0.242 \pm 0.071$ & $(0.051, 0.419)$ & $(0.088, 0.371)$ & $(0.183, 0.430)$ \\
\midrule
SCoNe                             & $0.298 \pm 0.100$           & $(0.068, 0.480)$ & $(0.147, 0.459)$ & $(0.260, 0.359)$ \\
SCNN (order 2)                    & $0.327 \pm 0.075$ & $(0.085, 0.517)$ & $(0.197, 0.516)$ & $(0.287, 0.451)$ \\
SCCNN (order 1)                   & $0.343 \pm 0.089$ & $(0.060, 0.491)$ & $(0.157, 0.510)$ & $(0.296, 0.393)$ \\
\bottomrule
\end{tabular*}
\end{table}

\begin{figure}[ht]
\centering
\begin{subfigure}{\linewidth}
\centering
\includegraphics[width=\linewidth]{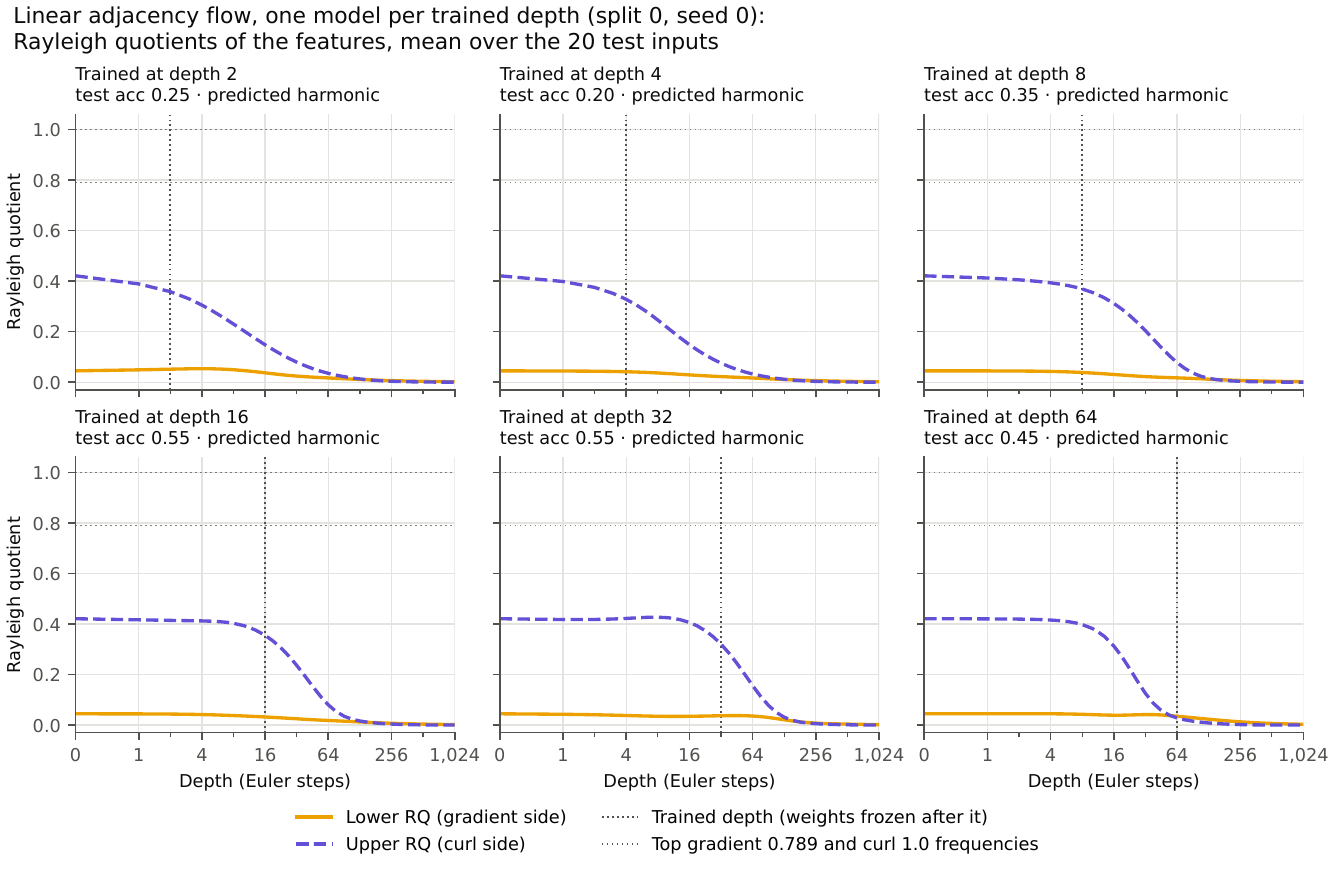}
\caption{Adjacency-like flow.}
\label{fig:drifters-depth-adj}
\end{subfigure}

\vspace{2mm}
\begin{subfigure}{\linewidth}
\centering
\includegraphics[width=\linewidth]{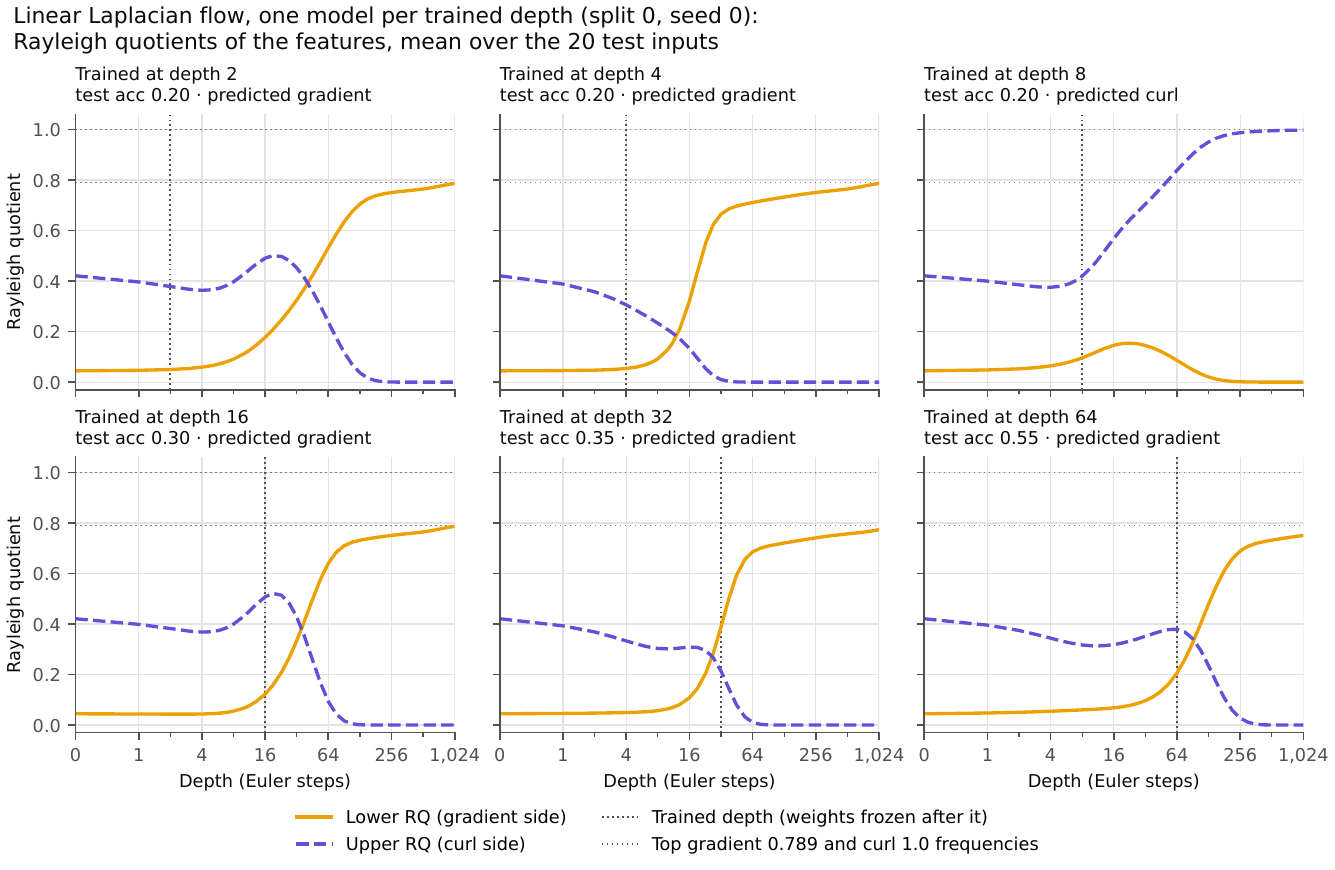}
\caption{Pure-Laplacian flow.}
\label{fig:drifters-depth-lap}
\end{subfigure}
\caption{\textbf{Training depth vs.\ learned Rayleigh quotients on ocean drifters} (split 0, seed 0). Each panel is one model trained at the indicated depth (vertical dotted line), after which its weights are frozen and propagation continues to depth 1,024. Curves are the lower and upper Rayleigh quotients averaged over the 20 test inputs. Horizontal dotted lines mark the top gradient (0.789) and curl (1.0) frequencies. Panel titles give test accuracy and the predicted regime.}
\label{fig:drifters-depth}
\end{figure}

\paragraph{Training depth vs.\ learned Rayleigh quotients}
\Cref{fig:drifters-depth} varies the training depth of the linear flows. For one split and seed (split 0, seed 0), we train one model per depth $L\in\{2,4,8,16,32,64\}$, freeze its weights after depth $L$, and continue propagation to depth 1,024, plotting the Rayleigh-quotient pair averaged over the 20 test inputs. Every adjacency-like model is predicted harmonic-dominant, and both Rayleigh quotients decay to zero at every training depth. The pure-Laplacian models are predicted top-gradient at every depth except $L=8$, which is predicted top-curl, and each approaches its predicted endpoint.

\subsubsection{Other datasets}
\label{app:other-datasets}

\begin{figure}[ht]
\centering
\begin{subfigure}{\linewidth}
\centering
\includegraphics[width=\linewidth]{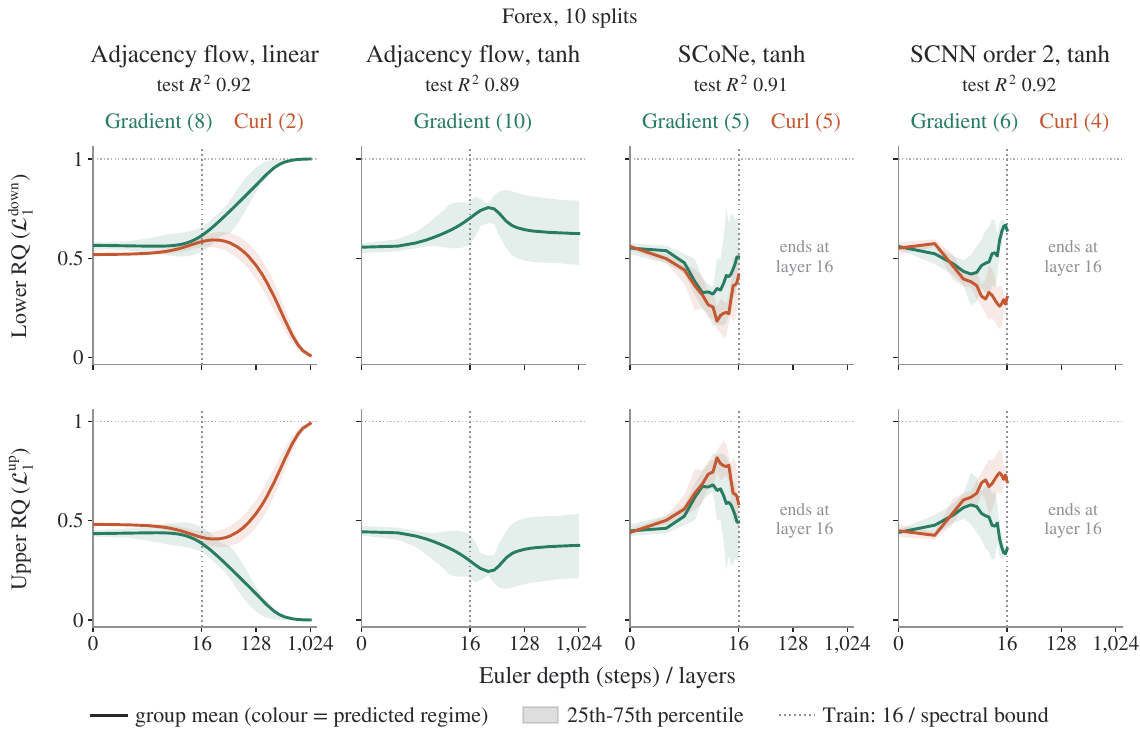}
\caption{Forex.}
\label{fig:other-datasets-forex}
\end{subfigure}

\vspace{2mm}
\begin{subfigure}{\linewidth}
\centering
\includegraphics[width=\linewidth]{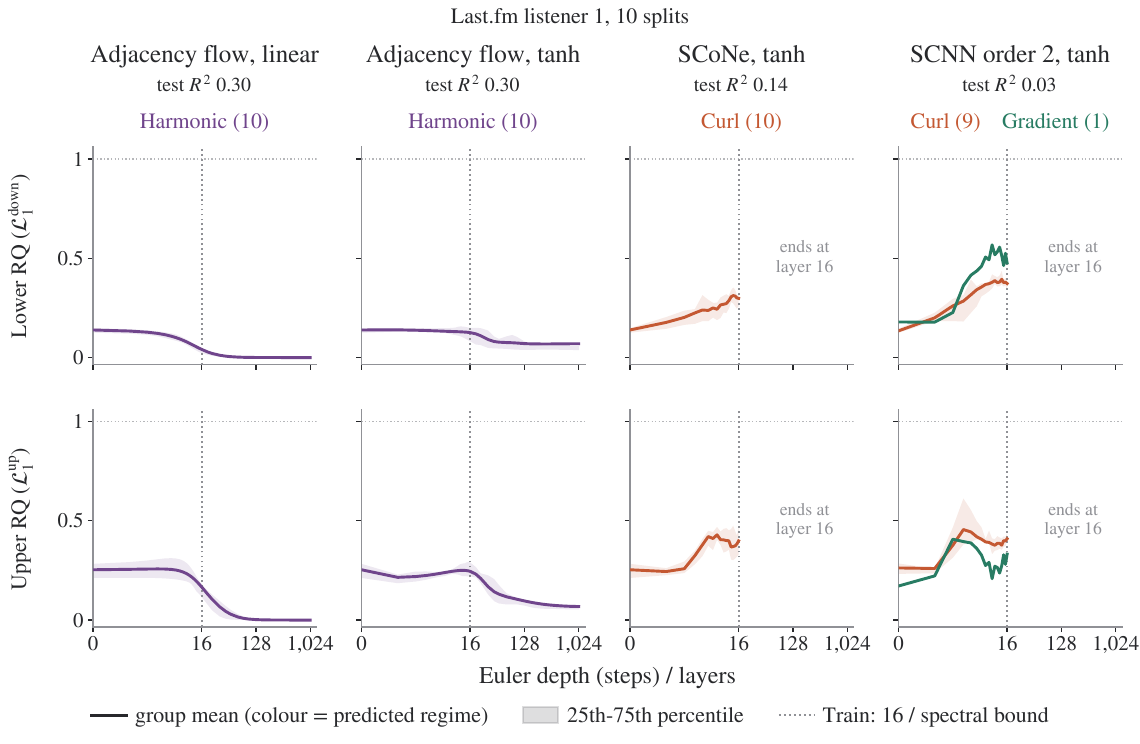}
\caption{Last.fm (listener 1).}
\label{fig:other-datasets-lastfm}
\end{subfigure}
\caption{\textbf{Lower (top row) and upper (bottom row) Rayleigh quotients on Forex and Last.fm}, over 10 splits. Column titles give the model and its test $R^2$. Checkpoints are grouped by their predicted regime (colour), with group sizes in parentheses; lines are group means and shading the 25th--75th percentiles. The vertical dotted line marks training depth 16, where SCoNe and SCNN, with 16 layers, end; horizontal dotted lines mark the spectral bounds.}
\label{fig:other-datasets}
\end{figure}

\Cref{fig:other-datasets} repeats the frozen-flow analysis on two further datasets, Forex and Last.fm (listener~1), each over 10 splits, for the adjacency-like flow (linear and with $\tanh$), SCoNe and SCNN of order 2 (both with $\tanh$).

\paragraph{Forex}
We use the exchange-rate snapshot released with the edge-flow
interpolation benchmark of \citet{jia2019graph}. We fill every 3-clique of the complete graph with a triangle.
The underlying complex then has 25 vertices, 300 edges, 2{,}300 triangles and
$\beta_1=0$. The flow on edge $(i,j)$ is the log mid rate of $i/j$,
averaged over both quoted directions of the exchange rate. The task is to interpolate
held-out edges of this single flow. We did not add additional features or
noise. We draw random edge splits into 50\% test, 10\%
validation and 40\% training.

\paragraph{Last.fm}
We use the Last.fm 1K dataset gathered by \citet{Celma2010lastfm}.
We keep the largest connected component of the dataset and within we fill every
3-clique. This gives 651 vertices, 1{,}527 edges, 500
triangles and $\beta_1=456$. We draw random edge splits into 50\% test, 10\%
validation and 40\% training.

\end{document}